\pdfoutput=1

\documentclass[11pt]{article}

\usepackage[preprint]{acl}
\usepackage{times}
\usepackage{latexsym}

\usepackage[T1]{fontenc}

\usepackage[utf8]{inputenc}

\usepackage{microtype}
\usepackage{graphicx} 
\usepackage{booktabs}
\usepackage{multirow}
\usepackage{colortbl}
\usepackage{inconsolata}
\usepackage{amsmath}
\usepackage{amsfonts}
\usepackage{bm}
\usepackage{subcaption}
\usepackage[ruled,vlined]{algorithm2e}
\usepackage{mdframed}
\usepackage{framed}
\usepackage{multicol}
\usepackage{bbm}

\title{Cracking Factual Knowledge: A Comprehensive Analysis of Degenerate Knowledge Neurons in Large Language Models}
\author{
    Yuheng Chen\textsuperscript{\rm 1,\rm 2 
    }
    Pengfei Cao\textsuperscript{\rm 1,\rm 2 
    },
    Yubo Chen\textsuperscript{\rm 1,\rm 2 
    },
    Yining Wang \textsuperscript{\rm 3},
    Shengping Liu \textsuperscript{\rm 3}
    Kang Liu\textsuperscript{\rm 1,\rm 2},
    Jun Zhao\textsuperscript{\rm 1,\rm 2}\\
  \textsuperscript{\rm 1}The Laboratory of Cognition and Decision Intelligence for Complex Systems \\
  \textsuperscript{\rm 2}Institute of Automation, Chinese Academy of Sciences, Beijing, China \\
  \textsuperscript{\rm 3}Unisound, Beijing, China \\
  \texttt{chenyuheng22@ia.ac.cn, \{pengfei.cao, yubo.chen\}@nlpr.ia.ac.cn,}\\
  \texttt{wangyining@unisound.com, 
  liushengping@unisound.com,
  \{kliu, jzhao\}@nlpr.ia.ac.cn}
}

\begin{document}
\maketitle
\begin{abstract}
Large language models (LLMs) store extensive factual knowledge, but the underlying mechanisms remain unclear. Previous research suggests that factual knowledge is stored within multi-layer perceptron weights, and some storage units exhibit degeneracy, referred to as \textit{Degenerate Knowledge Neurons} (DKNs).
Despite the novelty and unique properties of this concept, it has not been rigorously defined or systematically studied. We first consider the connection weight patterns of MLP neurons and define DKNs from both structural and functional aspects.  Based on this, we introduce the \textit{Neurological Topology Clustering} method, which allows the formation of DKNs in any numbers and structures, leading to a more accurate DKN acquisition.
Furthermore, inspired by cognitive science, we explore the relationship between DKNs and the robustness, evolvability, and complexity of LLMs. Our execution of 34 experiments under 6 settings demonstrates the connection between DKNs and these three properties. The code will be available soon.
\end{abstract}
\section{Introduction}
\begin{figure}[t]
\small
    \centering
    \includegraphics[width=\linewidth]{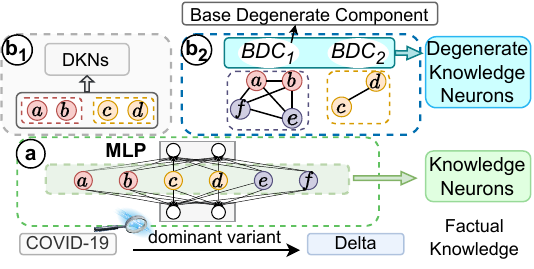} 
\caption{Explanation of KNs and DKNs. (a) illustrates the KNs corresponding to a fact, ($\text{b}_1$) and ($\text{b}_2$) are the preliminary definition of DKNs  and our complete definition of DKNs, respectively.}
    \label{intro-fig1-kn-dkn}
    \vspace{-10pt}
    
\end{figure}

Large pretrained language models (PLMs) are believed to  store extensive factual knowledge \cite{llama2-technical-report, openai2023gpt4}, yet the mechanisms of knowledge storage in PLMs remain largely unexplored. 
\citet{dai2022knowledge} propose that some multi-layer perceptron (MLP) neurons can store ``knowledge''.
As shown in the Figure \ref{intro-fig1-kn-dkn}(a), for the fact $\langle$\textit{COVID-19}, \textit{dominant variant}, \textit{Delta}$\rangle$, the corresponding knowledge storage units $a$ through $f$ are termed knowledge neurons (KNs).
\citet{chen2023journey} find that distinct KN pairs, such as \{$a$, $b$\} and \{$c$, $d$\} in Figure \ref{intro-fig1-kn-dkn}($\text{b}_1$), can store identical facts. They define the set of these pairs as degenerate knowledge neurons (DKNs) from a functional perspective.

While \citet{chen2023journey} have conducted some exploration on DKNs, their definition and acquisition method for DKNs still face two issues.
(1) \textit{Numerical Limitation}: They constrain each DKN's element to contain just two KNs, such as \{$a$, $b$\} in Figure \ref{intro-fig1-kn-dkn}($\text{b}_1$). However, factual knowledge may require more than two neurons for representation \cite{allen2023physics-3.2}. (2) \textit{Connectivity Oversight}: They only consider the neurons themselves, neglecting the connectivity weights between neurons. However, knowledge expression requires the interaction of multiple neurons \cite{zhu2023physics-3.1}, thus, it is necessary to consider the connectivity structure between neurons. 

To address these two issues,  we first provide a comprehensive definition of DKNs from two perspectives.
\textbf{Functionally}, some subsets of KNs can independently express the same fact, termed as Base Degenerate Components (BDCs), such as \textit{BDC-1} and \textit{BDC-2} in Figure \ref{intro-fig1-kn-dkn}($\text{b}_2$). In other words, they exhibit mutual degeneracy. The set of these BDCs is referred to as a DKN.
\textbf{Structurally}, as shown in Figure \ref{intro-fig1-kn-dkn}($\text{b}_2$), BDCs like \textit{BDC-1} and \textit{BDC-2} differ in KN number and connection tightness. 
To assess DKNs' structural traits, 
we define neuron distances based on connection weights and analyze the structural properties of neuron sets accordingly.

\begin{figure}[t]
    \centering
    \includegraphics[width=\linewidth]{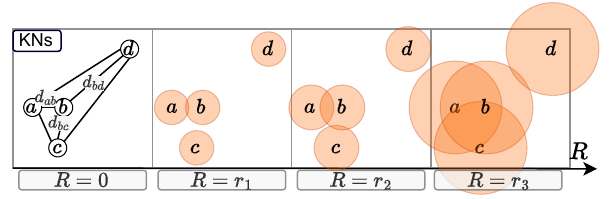} 
\caption{The clustering part of the Neurological Topology Clustering method. The $x$-axis ($R$) represents the increasing distance threshold starting from 0. Circles with radius $R$ are drawn around neurons, and intersecting circles indicate that the KNs are clustered together.}
    \vspace{-10pt}
    \label{intro-fig-tda}
\end{figure}

Based on the above definition, we introduce the \textit{Neurological Topology Clustering} \textbf{(NTC, \textsection \ref{section:method-NTC})} method to obtain DKNs, which includes clustering and filtering stages, enabling the formation of BDCs with an arbitrary number of neurons and arbitrary neuronal connection structures.
Figure \ref{intro-fig-tda} illustrates the clustering stage. Given four KNs $\{a,b,c,d\}$ with fixed connection weights and an increasing distance threshold $R$ starting from 0, we observe whether the KNs can cluster together as $R$ changes. At $R=0$, the KNs are isolated points. When $R=r_1>d_{ab}$, $\{a,b\}$ form a cluster; at $R=r_2>d_{bc}$, $\{a,b,c\}$ cluster together; and at $R=r_3>d_{bd}$, $\{a,b,c,d\}$ form a single cluster. Notably, a wide range of $R$ values maintains the $\{a,b,c\}$ cluster (from $r_2$ to $r_3$), indicating its stable existence. This stable cluster, suggesting a strong knowledge expression ability \cite{zhu2023physics-3.1}, is identified as a BDC. BDCs are then filtered to derive DKNs, as detailed in Section \ref{section:method-NTC}.

\begin{figure}[t]
    \centering
    \includegraphics[width=\linewidth]{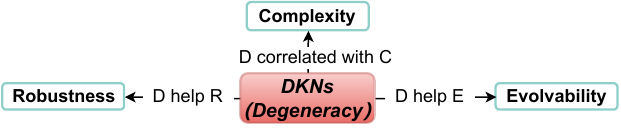} 
    \caption{The relationship between DKNs (with degeneracy property), robustness, evolvability and complexity in PLMs.}
    \label{intro-fig3-degenerate}
    \vspace{-10pt}
\end{figure}

Furthermore, inspired by cognitive science research on degeneracy \cite{whitacre2010degeneracy,edelman2001degeneracy,whitacre2010degeneracy2,mason2015degeneracy}, we investigate the relationship between DKNs and the robustness, evolvability, and complexity of PLMs.
Our findings are illustrated in Figure \ref{intro-fig3-degenerate} and elaborated upon in the following:


(1) \textbf{Robustness (\textsection \ref{section:Robustness})}: One aspect of PLMs' robustness is their ability to handle input interference \cite{fernandez2005modelrobust}.
Given a query subject to interference, we suppress (or enhance) the values or connection weights of DKNs and observe  resulting changes in the PLMs' prediction probability for the query.
Furthermore, we conduct a fact-checking experiment \cite{guo2022survey-fact-checking}, using DKNs to detect false facts. Experiments demonstrate that \textbf{DKNs can help PLMs cope with input interference}, indicating their contribution to robustness.

(2) \textbf{Evolvability (\textsection \ref{section:Evolvability})}: Evolvability is defined as the ability to adaptively evolve in new environments \cite{kirschner1998evolvability}. Inspired by this, we consider learning new knowledge as one aspect of PLMs' evolvability. We validate the significance of DKNs in PLMs' ability to learn new knowledge from two perspectives.
First, to prove that PLMs utilize DKNs to learn new knowledge, we directly fine-tune the PLMs and find that the regions of parameter changes highly overlap with DKNs. 
Second, we employ an efficient fine-tuning method,  freezing all MLP neurons except DKNs. We discover that PLMs can utilize DKNs to efficiently learn new knowledge while not forgetting old knowledge.
In summary, \textbf{DKNs enable PLMs to learn new knowledge more efficiently}, highlighting their crucial role in enhancing evolvability.

(3) \textbf{Complexity (\textsection \ref{section:Complexity})}: PLMs' complexity is positively correlated to the number of parameters \cite{mars2022word=degeneracy-complexity}.
In experiments related to the robustness and evolvability of PLMs,
we compare the performance of PLMs across different scales and find that \textbf{degeneracy is positively correlated with complexity.} 
In addition to these main experiments, we also conduct a supplementary  fact-checking experiment on complex texts. This additional  study proves that we can combine the large PLMs' complex text understanding ability with the DKNs' fact-checking capability to complete the task.

Our contributions can be summarized as follows:
\vspace{-15pt}
\begin{itemize}
\vspace{-5pt}
\item We provide a comprehensive definition of DKNs from both functional and structural aspects, pioneering the study of structures in PLMs' factual knowledge storage units.
\vspace{-5pt}
\item We introduce the Neurological Topology Clustering method, allowing the formation of DKNs in any numbers and structures, leading to a more accurate DKN acquisition.
\vspace{-5pt}
\item Through the lens of DKNs, we investigate the robustness, evolvability, and complexity of PLMs. By conducting 34 experiments
under 6 settings, we demonstrate the relationship between DKNs and these three properties. 
\end{itemize}

\vspace{-3mm}
\section{Datasets and Models}
\vspace{-1mm}
We utilize the TempLama dataset \cite{dhingra2022time-templama} to analyze DKNs. Each data instance includes a relation name, a date, a query, and an answer, such as $\langle$\textit{P37},  \textit{September 2021}, \textit{COVID-19}, \textit{dominant variant}, \underline{\hspace{0.3cm}}$\rangle$.
Except for timestamps, our dataset matches the Lama \cite{petroni2019language-lama-mlama-dataset, petroni2020how-lama-dataset} and mLama \cite{kassner2021multilingual-mlama} format used by \citet{dai2022knowledge} and \citet{chen2023journey}.
Under different experimental settings, we perform data augmentation or filtering operations on the dataset, resulting in six datasets that differ in content but are consistent in format.
Regarding model selection, we choose GPT-2 \cite{radford2019language-gpt2} and LLaMA2-7b \cite{llama2-technical-report}, both of which are based on the currently most popular auto-regressive architecture but have different parameter sizes, allowing us to test the generalization and scalability of our methods and conclusions.
\vspace{-1mm}
\section{Neurological Topology Clustering}
\label{section:method-NTC}
\vspace{-1mm}
\subsection{Definition of DKNs}
\paragraph{Formalization} Given a fact, we utilize the AMIG method \cite{chen2023journey} to obtain KNs, denoting them as $\mathcal{N} = \{n_1, n_2, \ldots, n_k\}$, where \(n_i\) is a KN.
For details of this method, see Appendix \ref{appendix: amig}. 
Let DKNs be denoted as \(\mathcal{D}\), containing \(s\) elements, \(\mathcal{D}= \{\mathcal{B}_1, \mathcal{B}_2, \ldots, \mathcal{B}_s\}\), where \(\mathcal{B}_j= \{n_{j1}, n_{j2}, \ldots, n_{\left|{\mathcal{B}_j}\right|} \}\) is named as the Base Degenerate Component (BDC). Thus, this fact ultimately corresponds to a set of DKNs:
\begin{equation}
    \small
\resizebox{\linewidth}{!}{$
\label{equ:DKNs}
    \mathcal{D} = \{\mathcal{B}_1, \ldots, \mathcal{B}_s\}=
    \{(n_{11}, \ldots, n_{\left|{\mathcal{B}_1}\right|}), \ldots, (n_{s1}, \ldots,n_{\left|{\mathcal{B}_s}\right|})\}
$}
\end{equation}
\vspace{-7mm}
\paragraph{Functional Definition}
Degeneracy requires that each BDC should independently express a fact. Let \({Prob}({\mathcal{B}})\) represent the PLMs' answer prediction probability when \(\mathcal{B}\) is activated, then the functional definition of DKNs is:
\begin{equation}
    \small
    {Prob}({\mathcal{D}}) \approx {Prob}({\mathcal{B}_i}), \forall i=1,2, \ldots, s
\label{dkn-define-1}
\end{equation}
\begin{equation}
\small
    {Prob}(\emptyset) \ll {Prob}({\mathcal{B}_i}), \forall i=1,2, \ldots, s
\label{dkn-define-2}
\end{equation}
where Equation \ref{dkn-define-1} indicates that activating any single BDC is sufficient to express the fact, and Equation \ref{dkn-define-2} suggests that if all BDCs are suppressed (i.e., activating the empty set $\emptyset$), the fact cannot be correctly expressed.
\begin{algorithm}[t]
\SetAlgoLined
\small
\caption{\small Neurological Topology Clustering}
\label{algorithm-ntc}
\KwIn{Knowledge neurons $\mathcal{N}$, Adjacent matrix $\mathcal{A}$, dynamic threshold $\tau_1$ and threshold $\tau_2$}
\KwOut{Degenerate knowledge neurons $\mathcal{D}$}

\textcolor{blue}{\tcp{Initialization}}
Initialize $\mathcal{D} = \emptyset$ and $R \gets 0$, where $R$ is the distance threshold for persistent homology.

\textcolor{blue}{\tcp{Persistent Homology (Clustering)}}
\While{$R$ increases}{
    Record all base degenerate components $\mathcal{B}_i$ and their corresponding persistence duration $R_{\text{p}}$\
}
\textcolor{blue}{\tcp{Filtering}}

\For{each $\mathcal{B}_i$}{
    \If{$R_{\text{p}}(\mathcal{B}_i) > \tau_1$ and $\text{Prob}(\mathcal{B}_i) > \tau_2$}{
        Add $\mathcal{B}_i$ to $\mathcal{D}$, where $\text{Prob}(\mathcal{B}_i)$ is the prediction probability when $\mathcal{B}_i$ is activated
    }
}
\Return{$\mathcal{D}$}
\end{algorithm}
\begin{table*}[h]
\vspace{-1mm}
\centering
\small
\resizebox{\textwidth}{!}
{
\begin{tabular}{c|ccccccc}
\toprule
& \multicolumn{7}{c}{\textbf{\cellcolor{gray!7}GPT-2}} 
\\
\cmidrule(lr){2-8}
\multirow{-2}{*}{\textbf{Method}} & \textbf{2} & \textbf{3} & \textbf{4} & \textbf{5} & \textbf{6} & \textbf{7} & 
\textbf{Average} \\
\midrule
DBSCAN 
& 12.0 $\rightarrow$ 26.0 
& 17.7 $\rightarrow$ 38.8 
& 23.4 $\rightarrow$ 33.8
& 32.9 $\rightarrow$ 53.4 
& 24.8 $\rightarrow$ 49.8 
& 26.4 $\rightarrow$ 46.6
& 23.90 $\rightarrow$ 34.72
\\
Hierarchical 
& 2.9 $\rightarrow$ 3.8 
& 15.5 $\rightarrow$ 7.3
& 3.7 $\rightarrow$ 12 
& 4.3 $\rightarrow$ 27 
& 25.0 $\rightarrow$ 92.0 
& ---
& 20.78 $\rightarrow$ 30.81
\\
K-Means 
& 27.4 $\rightarrow$ 38.3
& 17.8 $\rightarrow$ 29.5 
& 27.8 $\rightarrow$ 44.6 
& 32.4 $\rightarrow$ 58.7 
& 28.9 $\rightarrow$ 30.5 
& 21.8 $\rightarrow$ 30.1
&34.42 $\rightarrow$ 38.86

\\
AMIG & -0.8 $\rightarrow$  0.9  & --- &  ---  & --- &  ---  & --- 
& -0.8 $\rightarrow$  0.9  \\

\midrule
\textbf{NTC (Ours)}  
& \textbf{7.9 $\rightarrow$ 53} 
& \textbf{8.6 $\rightarrow$ 44}
& \textbf{12 $\rightarrow$ 92}
& \textbf{11 $\rightarrow$ 53}
& \textbf{3.1 $\rightarrow$ 32}
& \textbf{7.8 $\rightarrow$ 61}
& \textbf{9.32} $\rightarrow$ \textbf{55.60} \\
\bottomrule
\toprule
& \multicolumn{7}{c}{\cellcolor{gray!7}\textbf{LLaMA2}}
\\
\cmidrule(lr){2-8}
\multirow{-2}{*}{\textbf{Method}} 
& \textbf{2} & \textbf{3} & \textbf{8} & \textbf{11} & \textbf{14} & \textbf{17} & \textbf{Average} \\
\midrule
DBSCAN 
& 7.1 $\rightarrow$ 15.4
& 8.7 $\rightarrow$ 15.2
& 7.7 $\rightarrow$ 16.3 
& 7.2 $\rightarrow$ 25.3 
& --- 
& ---  
& 22.26 $\rightarrow$ 18.24 \\

Hierarchical 
& 27.6 $\rightarrow$ 44.3 
& 22.3 $\rightarrow$ 6.5
& --- 
&  ---  
& --- 
& --- 
& 27.50 $\rightarrow$ 44.04 \\

K-Means 
& 2.8 $\rightarrow$ 16.1
& 19.2 $\rightarrow$ 39.1
& 37.9 $\rightarrow$ 97.1 
& 50.7 $\rightarrow$ 111
& 36.3 $\rightarrow$ 50.0
& 4.9 $\rightarrow$ 17.9
& 20.08 $\rightarrow$ 39.24 \\

AMIG 
& -1.0 $\rightarrow$  2.0
& --- 
&  ---  
& --- 
&  ---  
& --- 
& -1.0 $\rightarrow$  2.0  \\
\midrule
 \textbf{NTC (Ours)}  
& \textbf{2.8 $\rightarrow$ 15.6}
& \textbf{4.3 $\rightarrow$ 19.1} 
& \textbf{7.8 $\rightarrow$ 50.1} 
& \textbf{4.6 $\rightarrow$ 25.7} 
& \textbf{13.6 $\rightarrow$ 37.7}
& \textbf{31.2 $\rightarrow$ 135}
& \textbf{14.11} $\rightarrow$ \textbf{28.78}\\
\bottomrule
\end{tabular}
}
\caption{Comparison of $\mathcal{D}$ obtained by different methods. The numbers above each column indicate the cardinality of the BDC sets. 
Each cell shows two values for the PLMs' prediction probability decrease ($\Delta Prob$ (\%)): the left is the average $\Delta Prob$ when partially suppressing BDCs (i.e., 1 to n-1 BDCs), and the right is the $\Delta Prob$ when fully suppressing all BDCs.  Lower left values, higher right values, and larger differences between them indicate better degeneracy.
The symbol ``---'' signifies that a specific method failed to generate a $\mathcal{D}$ with the specified cardinality.
}
\label{table-1-dkn-result}
\vspace{-2mm}
\end{table*}
\begin{figure*}[t]
    \centering
    \includegraphics[width=\linewidth]{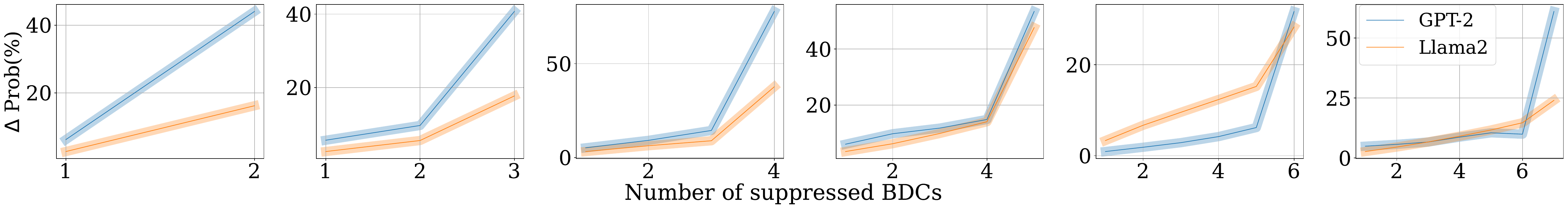} 
\small
\caption{The relationship between $\Delta Prob$ and the number of suppressed BDCs (using NTC). Table \ref{table-1-dkn-result} averages the results for suppressing 1 to n-1 BDCs, while this figure shows the changes as the number of suppressed BDCs varies from 1 to n, with the final point representing all BDCs suppressed. 
Lower $\Delta Prob$ for partial suppression  and higher $\Delta Prob$ for full suppression, i.e., a more prominent final turning point, indicate better degeneracy.
As the cardinality of $\mathcal{D}$ varies across PLMs and methods, Table \ref{table-1-dkn-result} and Figure \ref{fig-main} show representative results. Full results are in Appendix \ref{appendix:NTC} (Table \ref{tab:appendix:NTC}, Figures \ref{fig-appendix-main}, \ref{fig-appendix-DBSCAN}, \ref{fig-appendix-Hierarchical}, \ref{fig-appendix-K-Means-1}, \ref{fig-appendix-K-Means-2}).
}
    \label{fig-main}
    \vspace{-5mm}
\end{figure*}
\vspace{-1mm}
\paragraph{Structural Definition}
\citeauthor{zhu2023physics-3.1}\shortcite{zhu2023physics-3.1} argue that  tightly connected neurons  tend to store knowledge centrally.
Thus, we use the adjacency matrix $\mathcal{A}$ to evaluate the DKNs' connection structure.
For neurons \( A \) and \( B \) in layer \( l_A \) and layer \( l_B \) respectively, we calculate the distance \( d_{AB} \) as follows:
\begin{equation}
\small
\resizebox{\linewidth}{!}{$
\label{equal-distance}
    d_{AB} = 
    \begin{cases} 
      \left| {1}/{w_{AB}} \right| & \text{if } w_{AB} \neq 0 \text{ and } \left| l_A - l_B \right| = 1, \\
      \displaystyle\min_{\substack{P \in \text{Paths}(\mathcal{N})}}\sum_{\substack{(i,j) \in P}} d_{ij} & \text{if } \left| l_A - l_B \right| > 1 \text{ and a path exists}, \\
      \infty & \text{otherwise}.
    \end{cases}
$}
\end{equation}
where \( \text{Paths}(\mathcal{N})\) includes all paths from \(A\) to \(B\) through \(\mathcal{N}\).
Distance calculation varies in three scenarios. First, for neurons in adjacent layers, it is the reciprocal of the weight. Second, for neurons spanning multiple layers, it is the shortest distance determined by a dynamic programming algorithm.  Third, for neurons in the same layer, since information in PLMs transmits  between layers rather than within a layer \cite{meng2022locating}, we set the distance to $\infty$. 
Hence,  any $\mathcal{D}$ can correspond to an adjacency matrix $\mathcal{A}$, where \(\mathcal{A} \in \mathbb{R}^{k \times k}\), and \(k\) is the number of knowledge neurons contained in $\mathcal{D}$.
Based on $\mathcal{A}$, beyond traditional neuron value editing, modifying the connection weights offers another method to edit PLMs. 

\vspace{-1.5mm}
\subsection{The Acquisition of DKNs}
\paragraph{Persistent Homology}
In our method, we capture the persistent homology \cite{edelsbrunner2008persistent} of KN sets, which represents the duration of the set's existence and the tightness of the set's connections. Figure \ref{intro-fig-tda} informally illustrates this process. Formally, given two KNs, \(n_i\) and \(n_j\), which are both the centers of expanding circles. When the start radius (i.e., distance threshold) is 0, this corresponds to \(R_{\text{s}}=0\). Suppose they touch at
radius \(R{_\text{e}}=r_1\), which marks the end radius \(R=r_1\), and a new start radius \(R{_\text{s}}=r_1\).
At this point, \(n_i\) and \(n_j\) are clustered together, forming a BDC, \(\mathcal{B} = \{n_i, n_j\}\), corresponding to a persistence duration \(R{_\text{p}}=R{_\text{e}}-R{_\text{s}} = r_1\). 
For details on persistent homology, see Appendix \ref{appendix:persistent-homology}. 

\vspace{-1.5mm}
\paragraph{NTC Method}
Our method is shown in Figure \ref{intro-fig-tda} and Algorithm \ref{algorithm-ntc}. We design two steps, i.e., clustering and filtering steps, to obtain DKNs. During the clustering process, as \(R\) increases from 0 to infinity, we record all BDCs along with their corresponding \(R{_\text{p}}\). During the filtering process, we initially select BDCs with $R{_\text{p}}$ above \(\tau_1\). Then, among these, only BDCs with a ${Prob}({\mathcal{B}_i})$ greater than a threshold \(\tau_2\) are kept. Finally, these BDCs constitute  $\mathcal{D}$, as shown in the formula below.
\begin{equation}
\small
\mathcal{D} = \{\mathcal{B}_i | R{_\text{p}}(\mathcal{B}_i) > \tau_1 \text{ and } {Prob}({\mathcal{B}_i}) \geq \tau_2 \}
\label{euqal-dkn-finally}
\end{equation}
\subsection{Experiments of DKNs Acquisition} 
\paragraph{Experimental settings} Given a set $\mathcal{D} = \{\mathcal{B}_1, \mathcal{B}_2, \ldots, \mathcal{B}_n\}$, we traverse all subsets of $\mathcal{D}$, suppressing values or connection weights of all BDCs in each subset. We then calculate the drop in the answer probability before (b) and after (a) suppression: $\Delta Prob(\%) = \frac{Prob_b-Prob_a}{Prob_b}$.
We select four other methods as baselines, including K-Means \cite{ahmed2020kmeans}, DBSCAN \cite{ester1996density}, Hierarchical Clustering \cite{murtagh2012hierarchical} and AMIG \cite{chen2023journey}. 

\begin{figure*}[t]
    \centering
    \includegraphics[width=\linewidth]{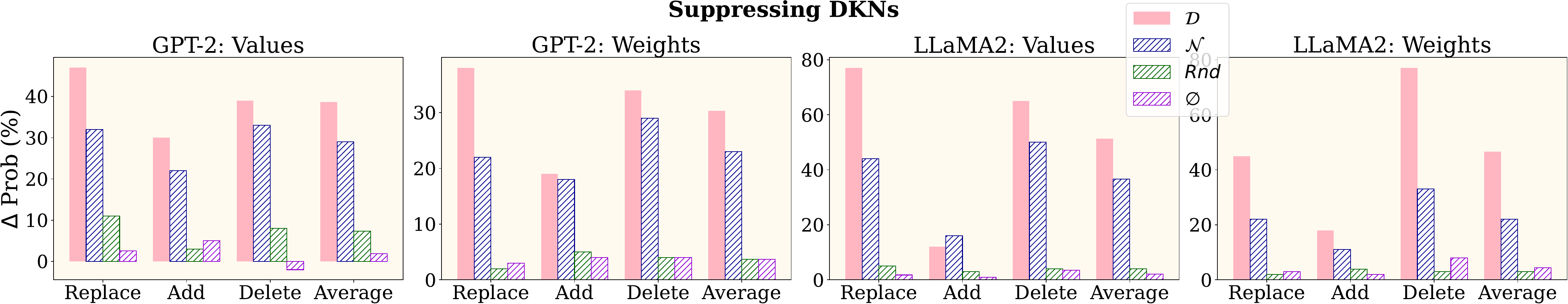} 
\small
\caption{Changes in the prediction probabilities of PLMs corresponding to the suppression of DKNs and other baselines. $\mathcal{D}$, $\mathcal{N}$, $Rnd$, and $\emptyset$ represent  DKNs,  KNs,  random neurons, and suppressing no neurons, respectively.}
    \label{fig-robust:suppress}
\vspace{-4mm}
\end{figure*}
\begin{figure*}[t]
    \centering
    \includegraphics[width=\linewidth]{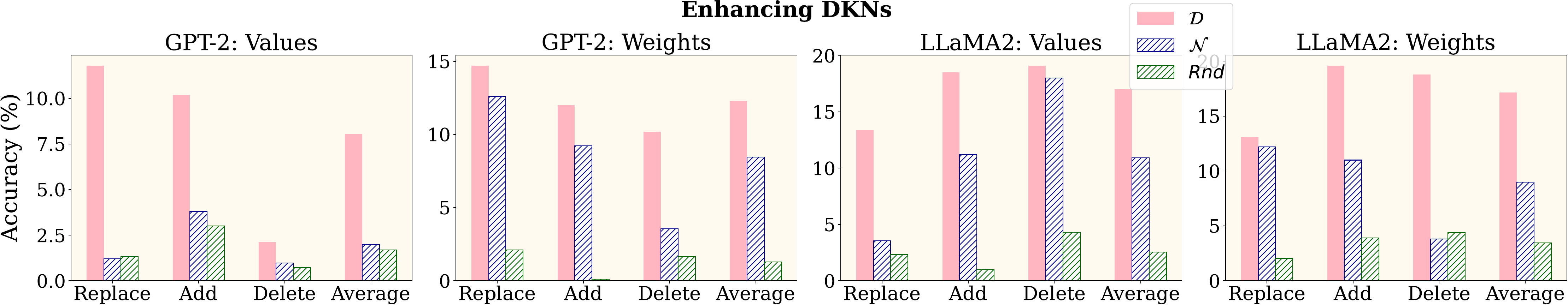} 
\small
\caption{Changes in the accuracy of PLMs corresponding to the enhancement of DKNs and other baselines. 
}
    \label{fig-robust:enhance}
\vspace{-3mm}
\end{figure*}
\paragraph{Findings}
(1)  DKNs identified by NTC exhibit strong degeneracy. As shown in Table \ref{table-1-dkn-result}, when suppressing a subset of BDCs, NTC yields a smaller $\Delta Prob$ compared to other baseline methods. Conversely, when suppressing all BDCs, NTC results in the largest $\Delta Prob$. The clear inflection point in Figure \ref{fig-main} further demonstrates that suppressing a subset of BDCs does not lead to a gradual increase in $\Delta Prob$. Instead, as long as one BDC exists, PLMs have a sufficient ability to express facts. In summary, the DKNs obtained by the NTC method most closely align with  the definitions of DKNs in Equations \ref{dkn-define-1} and \ref{dkn-define-2}, exhibiting the most desirable degeneracy properties.
(2) Suppressing DKNs may inadvertently enhance PLMs' knowledge expression. Negative values in Table \ref{table-1-dkn-result} indicate increased prediction probability, suggesting that suppressing some DKN subsets allows others to compensate and improve PLM performance.
\vspace{-1.5mm}
\section{The Impact of DKNs on Robustness}
\vspace{-1.5mm}
\label{section:Robustness}
\subsection{Query-Perturbation}
\begin{figure}[t]
\centering
    \includegraphics[width=\linewidth]{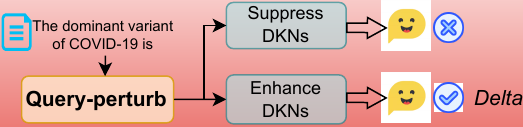} 
\small
\caption{Query perturbation and the impact of suppressing or enhancing DKNs on PLMs' performance.}
    \label{fig-robust}
    \vspace{-4mm}
\end{figure}

\paragraph{Explanation}
Robustness to input errors is one facet of PLMs' overall robustness. This is critical in practical scenarios where PLMs often encounter  user input errors like spelling mistakes or character omissions \cite{chen2010user-input-error}. 
To simulate this scenario, we apply random disturbances to the inputs.
Given an input sequence \(Q = \{q_1, q_2, \ldots, q_n\}\), we generate its perturbed counterpart \(Q^*\):
\begin{equation}
\small
\resizebox{\linewidth}{!}{
  $Q^* = 
  \begin{cases} 
      \{q_1, \ldots, q_{i-1}, \text{[replace]}, q_{i+1}, \ldots, q_n\} & \text{if replace}, \\
      \{q_1, \ldots, q_{i-1}, \text{[add]}, q_i, \ldots, q_n\} & \text{if add}, \\
      \{q_1, \ldots, q_{i-1}, q_{i+1}, \ldots, q_n\} & \text{if delete}. 
  \end{cases}$}
\end{equation}
where ``[replace]'' and ``[add]'' are special characters. To investigate the role of DKNs, we suppress and enhance them separately (Figure \ref{fig-robust}).
\paragraph{Suppressing DKNs}
We apply two different  operations to suppress DKNs:
(1) zeroing neuron values, and (2) nullifying neuron connection weights. 
For comparison, we select three baselines:  suppressing knowledge neurons, suppressing random neurons (the same number as DKNs),  and suppressing no neurons. For the PLMs subjected to suppression operations, we calculate their prediction probabilities for both $Q$ and $Q^*$, and compute the change in probability: $\Delta Prob (\%) = \frac{Prob(Q) - Prob(Q^*)}{Prob(Q)}$.
Figure \ref{fig-robust:suppress} presents our results.

\paragraph{Enhancing DKNs} Since robustness in PLMs is always limited, they may answer some \( Q^* \) queries incorrectly.  We record these queries as \(Q^*_{err}\), and apply two operations to enhance DKNs: (1) doubling neuron values, and (2) doubling neuron connection weights. Then, we reassess the enhanced PLMs' accuracy on \(Q^*_{err}\), denoted as \(Acc_{\text{err}}\).
Unlike the previous metric, $\Delta Prob$ may not affect PLMs' outputs, despite its suitability for evaluating DKNs' effect across all queries. Therefore, to measure enhanced DKNs' impact on PLM performance, we choose \(Acc_{\text{err}}\), as it requires a significant change in $Prob$ to alter the output. 
Similarly, we adopt two baselines: enhancing knowledge neurons and enhancing random neurons. Notably, enhancing no neurons directly corresponds to $Q^*_{err}$, with an accuracy of 0.
Figure \ref{fig-robust:enhance} presents our results.

\begin{table*}[h]
\centering
\small
\begin{tabular}{c|ccc|ccc|ccc|ccc}
\toprule
\multirow{2}{*}{\textbf{Method}} & \multicolumn{3}{c|}{\textbf{GPT-2 (Relation)}} & \multicolumn{3}{c|}{\textbf{LLaMA2 (Relation)}} & \multicolumn{3}{c|}{\textbf{GPT-2 (Golden)}} & \multicolumn{3}{c}{\textbf{LLaMA2 (Golden)}} \\ 
& \textbf{P} & \textbf{R} & \textbf{F1} & \textbf{P} & \textbf{R} & \textbf{F1} & \textbf{P} & \textbf{R} & \textbf{F1} & \textbf{P} & \textbf{R} & \textbf{F1} \\
\midrule
KNs 
& 0.489 & 0.511 & 0.500 
& 0.481 & 0.455 & 0.468 
& 0.600 & 0.636 & 0.618 
& 0.568 & 0.741 & 0.643 \\ 
PLMs 
& 0.015 & 0.015 & 0.015 
& 0.035 & 0.036 & 0.036
& 0.015 & 0.015 & 0.015
& 0.035 & 0.036 & 0.036 \\
\midrule
\textbf{DKNs} 
& \textbf{0.497} & \textbf{0.520} & \textbf{0.508} 
& \textbf{0.505} & \textbf{0.500} & \textbf{0.502}  
& \textbf{0.549} & \textbf{0.848} & \textbf{0.667} 
& \textbf{0.530} & \textbf{0.955} & \textbf{0.682}
\\ 
\bottomrule
\end{tabular}
\caption{Results of the fact-checking experiment. 
``Relation'' represents DKNs or KNs based on relation, and 
``Golden'' represents DKNs or KNs obtained using true answers. 
``Golden'' better reflects DKNs' actual capability, as ``Relation'' may yield mismatches between DKNs (or KNs) and specific queries due to aggregation and filtering.}
\label{tab3-fact_checking_experiment}
\vspace{-3mm}
\end{table*}
\paragraph{Findings}
(1) As depicted in Figure \ref{fig-robust:suppress}, compared to the baselines, suppressing DKNs leads to a significant decrease in PLMs' prediction probabilities for $Q^*$, indicating that suppressing DKNs impairs the robustness of PLMs.
(2) Figure \ref{fig-robust:enhance} shows that, compared to the baselines, enhancing DKNs enables PLMs to provide correct answers to queries that were previously answered incorrectly due to interference, demonstrating that enhancing DKNs improves the robustness of PLMs.
In summary, we can conclude that \textbf{DKNs can help PLMs cope with input interference}, indicating their contribution to robustness.

\subsection{Fact-Checking}
\paragraph{Explanation}Since factual errors can also be seen as a form of perturbation, we naturally think of conducting fact-checking experiments based on DKNs. Unlike knowledge localization where the true value \(y^*\) is known, fact-checking does not allow us to know \(y^*\) in advance.
To address this, we divide the factual knowledge by relation. For queries \(Q^r\) corresponding to a relation, we split them into two parts: \(Q^{r1}\) for DKNs acquisition and \(Q^{r2}\) for testing.
For a given query \( Q^{r1}_i \), the corresponding set of DKNs is denoted as \( D^{r1}_i \). 
We aggregate them and select those that appear more than \(\tau_3\) times, denoting as the relation-based DKNs $D^r$:
\begin{equation}
\small
D^r = \left\{ n_i \mid n_i \in \bigcup_{D^{r1}_i} \text{ and } \sum_{j=1}^{|\bigcup_{D^{r1}_i}|} \mathbbm{1}{\{n_i = n_j\}} > \tau_3 \right\}
\vspace{-1mm}
\end{equation}
where $\sum_{j=1}^{|\bigcup_{D^{r1}_i}|} \mathbbm{1}{\{n_i = n_j\}}$ is the number of occurrences of \(n_i\).
Then, for a query \( Q^{r2}_i \), we determine the factual correctness by computing the average attribution score of all neurons in \( \mathcal{D}^r \):
\begin{equation}
\label{equal:fact-checking}
\vspace{-1.5mm}
\small
{FC}(Q^{r2}) = 
\begin{cases} 
    \text{True} & \text{if } {\displaystyle\sum_{\substack{n \in \mathcal{D}^r}} Score(n)}\,/\,{|\mathcal{D}^r|} > \tau_4, \\
    \text{False} & \text{otherwise}.
\end{cases}
\vspace{-1.5mm}
\end{equation}
where \( Score(n) \) denotes the activation score of each neuron \( n \) in \( \mathcal{D}^{r} \) (see Appendix \ref{appendix: amig} for details), $\tau_4$ is
a threshold.  
\paragraph{Experimental settings}
First, we construct a dataset $Q^{r2}_f$. For the queries in $Q^{r2}_i$, we replace their correct answers with other answers from the same relation (i.e., incorrect answer), and then perform fact-checking on $Q^{r2}_f$.
Then, we employ two baseline methods:
(1) Fact-checking based on KNs, and 
(2) Direct fact-checking with PLMs providing True or False answers. 
Besides using relation-based \( \mathcal{D}^{r} \), we also utilize \( \mathcal{D}^{r}_i \) obtained from true values (``Golden''). 
For evaluation metrics, we choose Precision (P), Recall (R), and F1-Score. Table \ref{tab3-fact_checking_experiment} presents the overall results.
\paragraph{Findings}
(1) DKNs have the strongest fact-checking ability. Comparing the P, R, and F1 values of KNs and PLMs, it can be confirmed that DKNs achieve better results under various settings.
(2)  The ``Golden'' results in Table \ref{tab3-fact_checking_experiment} outperform the results corresponding to ``Relation''. Thus, the more precise the location of DKNs, the better the fact-checking performance.  
(3) The fact-checking ability of PLMs is weak. PLMs can correctly answer the query yet fail to identify the fact's errors due to insufficient fact familiarity.   Since true values are used in obtaining \(D^r\), the fact-checking ability for such facts is stronger.
\begin{figure}[t]
\centering
    \includegraphics[width=\linewidth]{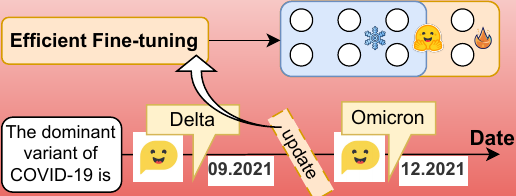} 
\small
\caption{Efficient fine-tuning of PLMs using DKNs to update knowledge across different timestamps.}
    \label{fig-evolve}
    \vspace{-2mm}
\end{figure}
\section{The Impact of DKNs on Evolvability}
\label{section:Evolvability}
In real-world scenarios, it is meaningful for PLMs to continuously learn or update new knowledge without forgetting old knowledge. Therefore, we investigate one aspect of PLMs' evolvability, namely the ability of PLMs to learn new knowledge. As shown in Figure \ref{fig-evolve}, given a timestamped query,  \textit{``In date \_\_, the dominant variant of COVID-19 is \_\_''}, a PLM may update its answer from ``Delta'' to ``Omicron'' across different timestamps, while still retaining the knowledge of the ``Delta'' variant.
\subsection{Overlap of DKNs and Parameter Changes} 
\paragraph{Experimental settings}We first directly fine-tuning the PLMs and then record the positions of neurons where significant parameter changes occur, denoted as \( \Delta N \):
\begin{figure*}[t]
\centering
\includegraphics[width=\linewidth]{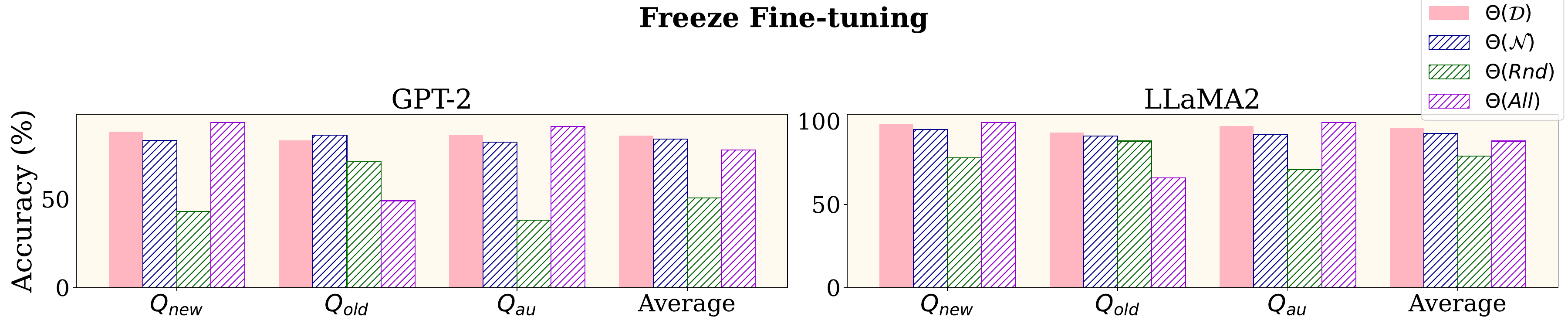} 
\small
\caption{Results of the fine-tuning experiment. $\Theta$ denotes the areas unfrozen in PLMs during fine-tuning, while \(Q_\text{new}\), \(Q_\text{old}\), and \(Q_\text{au}\) symbolize new, old, and augmented queries, respectively. A method is considered valuable for practical applications only if it achieves high and balanced accuracy across all datasets.}
    \label{fig-Fine-tuning:acc}
    \vspace{-4mm}
\end{figure*}
\begin{equation}
\label{equal:change para}
\small
\Delta N = \{ n \,|\, \Delta P(n) > \tau_{\Delta N} \} 
\end{equation}
\begin{equation}
\label{equation:overlap-change}
\small
\resizebox{\linewidth}{!}{$
\Delta P(n) = \sqrt{\left( \frac{\| w_2^{\text{fc}}(n) - w_1^{\text{fc}}(n) \|}{\| w_1^{\text{fc}}(n) \|} \right)^2 + \left( \frac{\| w_2^{\text{proj}}(n) - w_1^{\text{proj}}(n) \|}{\| w_1^{\text{proj}}(n) \|} \right)^2}
$}
\end{equation}where \( \tau_{\Delta N} \) is a dynamic threshold, \( \Delta P(n) \) indicates parameter change. \( w_1^{\text{fc}}(n) \) and \( w_1^{\text{proj}}(n) \) signify the feed-forward and projection weights of neuron \( n \) before fine-tuning, respectively, while \( w_2^{\text{fc}}(n) \) and \( w_2^{\text{proj}}(n) \) are their post-fine-tuning counterparts.
Then, we identify the corresponding $\mathcal{D}$ through Algorithm \ref{algorithm-ntc}, and calculate the overlap between \( \mathcal{D} \) and \( \Delta N \):
\begin{equation}
\small
\label{equal:overlap}
   O(\mathcal{D}, \Delta N) = |\mathcal{D} \cap \Delta N| \,/\, |\mathcal{D}|
\end{equation}
For comparison, we choose the KNs ($\mathcal{N}$) and randomly chosen neurons ($Rnd$) as baselines.

\begin{figure}[h]
\vspace{-2mm}
\centering
\begin{subfigure}{\linewidth}
\centering
\includegraphics[width=\linewidth]{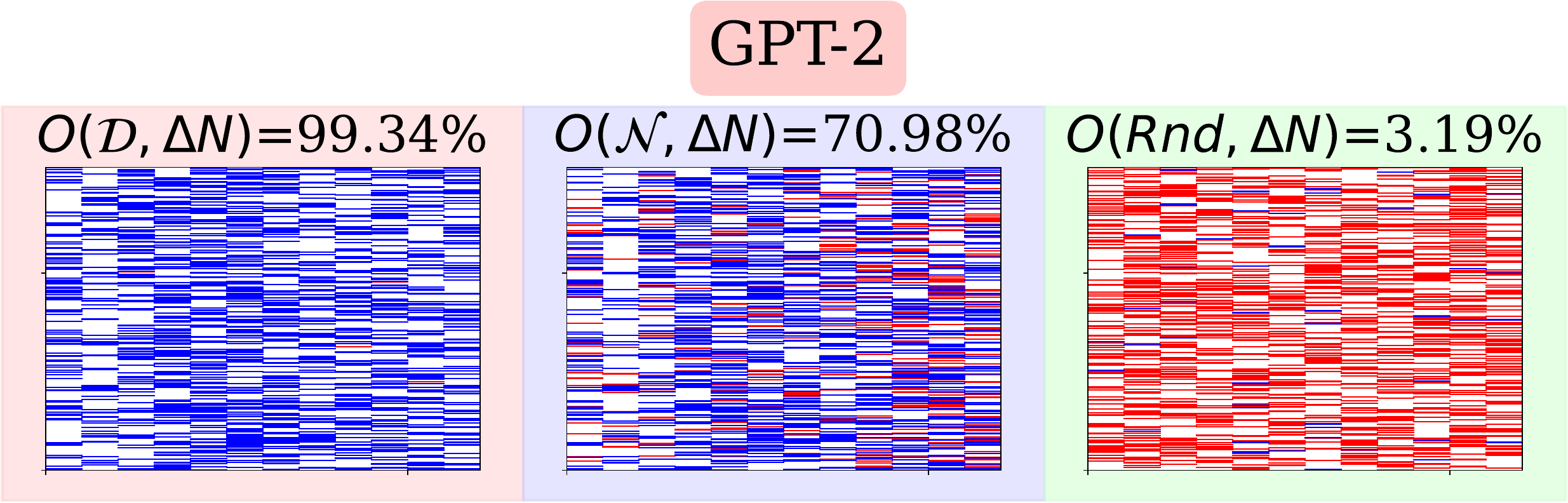}
\label{fig-overlap-dkn-gpt-2}
\vspace{-10pt}
\end{subfigure}
\begin{subfigure}{\linewidth}
\centering
\includegraphics[width=\linewidth]{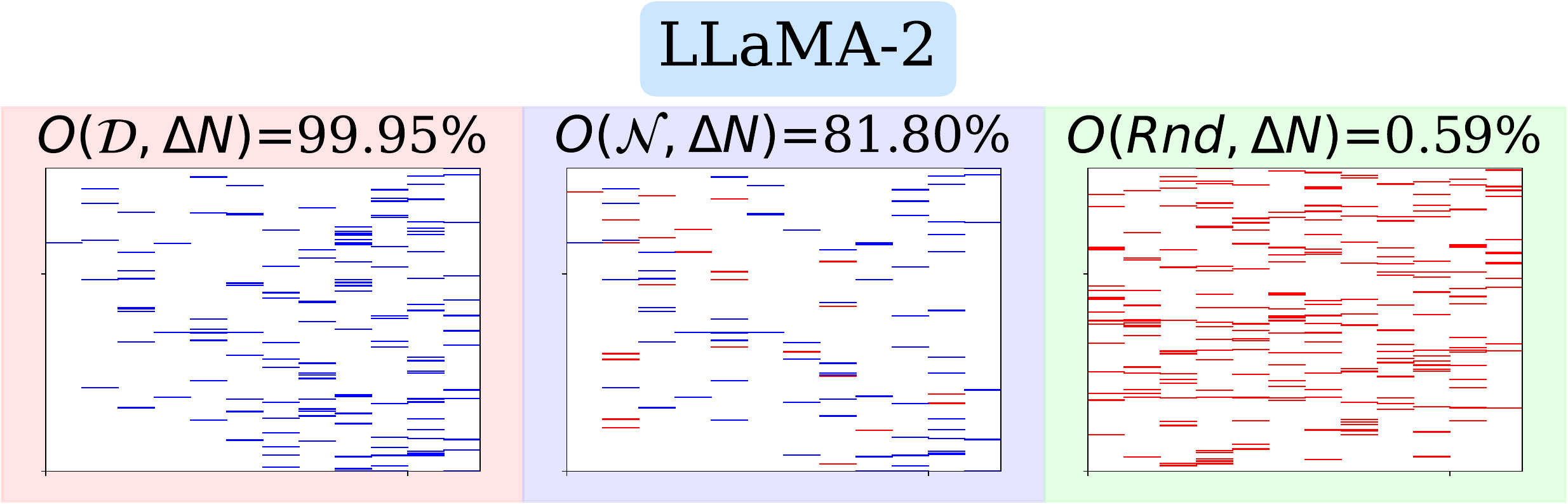}
\label{fig-overlap-dkn-llama}
\vspace{-10pt}
\end{subfigure}
\caption{Results of the parameter changes experiment. Blue represents the overlap between neurons and the area of parameter changes, while red indicates the non-overlapping portions.  ``$O$'' is the degree of overlap, with higher values suggest that PLMs  rely more on these neurons  for learning new knowledge.}
\vspace{-2mm}
\label{fig-overlap-dkn}
\end{figure}

\paragraph{Findings}
The PLMs indeed utilize DKNs to learn new knowledge. Figure \ref{fig-overlap-dkn} reveals \(O(\mathcal{D}, \Delta N)\) is highest, indicating that the parameter change area has the largest overlap degree with DKNs. Notably, LLaMA2 has more irrelevant neurons due to its larger number of parameters, but this does not conflict with its high \(O(\mathcal{D}, \Delta N)\).

\subsection{DKNs Guide PLMs to Learn Knowledge} 
\paragraph{Experimental settings} 
Since PLMs primarily utilize DKNs during fine-tuning, we naturally consider leveraging them for efficient fine-tuning to update knowledge (Figure \ref{fig-evolve}).
Given a dataset and corresponding \(\mathcal{D}\), we freeze the parameters outside of \(\mathcal{D}\) during fine-tuning.
Then we evaluate the role of DKNs in learning new knowledge by conducting the experiments with three distinct datasets.
(1) \(Q_\text{new}\): Comprising queries introducing new knowledge, aimed at evaluating PLMs' grasp of it.
(2) \(Q_\text{old}\): Comprising knowledge previously mastered by PLMs, with no overlap with \(Q_\text{new}\), used to assess if PLMs have forgotten old knowledge.
(3) \(Q_\text{au}\): Comprising augmented data, which rephrases \(Q_\text{new}\) into semantically identical but differently expressed queries. We use it to prove that PLMs actually learn knowledge, not just superficial semantic association.
Besides unfreezing DKNs ($\Theta(\mathcal{D})$), we select three baseline methods: unfreezing KNs (\(\Theta(\mathcal{N})\)), unfreezing a random set of neurons equal in number to the DKNs (\(\Theta(Rnd)\)), and unfreezing all neurons, i.e., direct fine-tuning (\(\Theta(All)\)).

\paragraph{Findings}
(1) \textbf{DKNs enable PLMs to learn new knowledge more efficiently while preserving old knowledge}. Figure \ref{fig-Fine-tuning:acc} shows that while direct fine-tuning can yield high accuracy on \(Q_\text{new}\), it may cause a markedly reduced accuracy on \(Q_\text{old}\). However, unfreezing DKNs not only maintains high accuracy on \(Q_\text{new}\) but also significantly improves performance on \(Q_\text{old}\) compared to direct fine-tuning.
This approach offers a new potential solution to the problem of catastrophic forgetting and makes the fine-tuning process more efficient.
(2) Comparing the accuracy of different methods, we find that DKNs possess the strongest ability to guide PLMs in learning new knowledge compared to other neurons.
(3) The close accuracy rates between $Q_\text{au}$ and \(Q_\text{new}\) indicate that PLMs can learn genuine knowledge rather than just superficial semantic information.

\section{The Impact of DKNs on Complexity}
\label{section:Complexity}
\subsection{Explanation of Complexity}
\paragraph{Hypotheses}In auto-regressive models, complexity, denoted by \( \mathcal{C}(M) \), is determined by the number of model parameters \cite{hu2021model-complexity}. Given two PLMs, \( M_1 \) and \( M_2 \). Let \( \mathcal{D}^*(M) \) be the PLMs' degeneracy,  \( \mathcal{T} \) be a downstream task using DKNs. The performance of \( M \) on \( \mathcal{T} \) is indicated by \( \mathcal{P}(M, \mathcal{T}) \), and \( {Param}(M) \) is the number of parameters in \( M \). We propose two hypotheses:
\begin{equation}
\small
\resizebox{\linewidth}{!}{$
\begin{aligned}
\textbf{Hypothesis 1} \quad & {Param}(M_1) < {Param}(M_2) \Rightarrow {\mathcal{C}}(M_1) < {\mathcal{C}}(M_2) \\
\textbf{Hypothesis 2} \quad & \mathcal{P}(M_1, \mathcal{T}) < \mathcal{P}(M_2, \mathcal{T}) \Rightarrow \mathcal{D}^*(M_1) < \mathcal{D}^*(M_2)
\end{aligned}
$}
\label{eq:complexity_theorem}
\end{equation}
\paragraph{Findings} 
(1) When DKNs are suppressed, LLaMA2 experiences a more significant drop in prediction probability compared to GPT-2. Conversely, enhancing DKNs leads to a more notable increase in LLaMA2's accuracy, as illustrated in Figures \ref{fig-robust:suppress} and \ref{fig-robust:enhance}.
(2) LLaMA2 surpasses GPT-2 in fact-checking abilities when utilizing DKNs, as indicated by Table \ref{tab3-fact_checking_experiment}.
(3) Compared to GPT-2, LLaMA2 has a stronger ability to learn new knowledge, retain old knowledge, and acquire genuine knowledge, as detailed in Figure \ref{fig-Fine-tuning:acc}.
The aforementioned phenomena indicate that \(\mathcal{P}(\text{GPT-2}, \mathcal{T}) < \mathcal{P}(\text{LLaMA2}, \mathcal{T})\), and given that \({Param}(\text{GPT-2}) < {Param}(\text{LLaMA2})\), according to Hypotheses \ref{eq:complexity_theorem}, we can conclude:
\begin{equation}
    \small
    {\mathcal{C}}(\text{GPT-2})<{\mathcal{C}}(\text{LLaMA2}) \Rightarrow \mathcal{D}^*(\text{GPT-2})<\mathcal{D}^*(\text{LLaMA2})
\end{equation}
In summary, \textbf{degeneracy is positively correlated with complexity}. 
\subsection{Fact-Checking within Complex Texts}
\paragraph{Experimental settings}
We investigate the potential of DKNs when integrated with PLMs' other abilities, through an experiment that melds the model's ability to comprehend complex texts with DKNs' fact-checking ability within such texts.
Given a triplet-form query \( Q \), first, we rewrite it  into a complex text \( Q^c \).
To perform fact-checking on \( Q^c \), the PLMs must possess both reasoning ability and fact-checking ability. 
Then, we provide a prompt to PLMs, requiring them to first identify both the query and answer in \( Q^c \), and then perform fact-checking according to Equation \ref{equal:fact-checking}.
\paragraph{Findings}(1) PLMs with the ability to understand complex texts can employ DKNs for fact-checking within complex texts. Table \ref{tab5-fact_checking_complex} shows LLaMA2's ability to utilize DKNs for fact-checking. However, GPT-2  only echoes prompts without yielding meaningful results, thus its exclusion from the table. 
\begin{table}[h]
\centering
\small
\begin{tabular}{c|ccc|ccc}
\toprule
\multirow{2}{*}{\textbf{Method}} 
& \multicolumn{3}{c|}{\textbf{LLaMA2 (Relation)}} 
& \multicolumn{3}{c}{\textbf{LLaMA2 (Golden)}} \\ 
& \textbf{P} & \textbf{R} & \textbf{F1} 
& \textbf{P} & \textbf{R} & \textbf{F1}\\
 \midrule
KNs & 0.59 & 0.48 & 0.53 
& 0.64 & 0.43 & 0.51  \\ 
PLMs & 0.02 & 0.02 & 0.02 
& 0.02 & 0.02 & 0.02 \\ 
\midrule
\textbf{DKNs} 
& \textbf{0.54} & \textbf{0.58} & \textbf{0.56} 
& \textbf{0.52} & \textbf{0.78} & \textbf{0.62} 
 \\
\bottomrule
\end{tabular}
\caption{Results of the fact-checking in complex texts. 
}
\label{tab5-fact_checking_complex}
\vspace{-3mm}
\end{table}
(2) When DKNs' locations are imprecise, pre-understanding complex texts can enhance fact-checking ability. The results of ``Relation'' in Table \ref{tab5-fact_checking_complex} surpass  those in Table \ref{tab3-fact_checking_experiment} (F1: 0.56 to 0.502),  
but the results of ``Golden''  decrease (F1: 0.62 to 
0.682). This indicates that relation-based DKNs lack precision, but complex text understanding increases PLMs' sensitivity to specific facts.

\section{Related Work}
\citet{petroni-etal-2019-language} argue  that numerous factual {knowledge} exists within PLMs and suggest using ``fill-in-the-blank'' cloze tasks determine if the models  have grasped specific facts. Meanwhile,
\citet{geva2021key-value} suggest that MLP neurons within transformer models  operate as key-value memories.
Building on this, \citet{dai2022knowledge} uncover that some MLP neurons are capable of storing factual knowledge, termed as {knowledge neurons} (KNs).
\citet{lundstrom2022rigorous} confirm the reliability of their knowledge localization method, while subsequent knowledge editing experiments by \citet{meng2022locating} and \citet{meng2023massediting} reinforce that MLP neurons indeed store factual knowledge.  Moreover, \citet{geva2023dissecting} investigate KN dynamics, and \citet{chen2023journey} discover multiple distinct KN sets storing identical facts, termed degenerate knowledge neurons (DKNs). Additionally, researchers have discovered various other KNs with unique properties \cite{Wang2022FindingSN,tang2024languagespecific}. 
In summary, despite limitations \cite{niu2024what,bricken2023towardsClaudeAI,chen2024knowledge}, the KN-based analysis approach remains valuable for further research. 

\section{Conclusion}
This paper delves into the mechanisms of factual knowledge storage in PLMs. First, we provide a comprehensive definition of DKNs that covers both structural and functional aspects, pioneering the study of the internal structures of PLMs. Based on this, we introduce the neurological topology clustering method for more precise DKN acquisition. Finally, through the lens of DKNs, we investigate the robustness, evolvability, and complexity of PLMs. Extensive experiments demonstrate the critical role of DKNs in PLMs.

\section*{Limitations}
First, limited by computational resources, our study involves only the GPT-2 and LLaMA2-7b models. To further validate the scalability of our methods and conclusions, it is imperative to conduct studies on larger models. Second, our research is confined to factual knowledge, and whether similar findings apply to other types of knowledge remains to be investigated.
Finally, the generalizability of our findings across different languages and cultural contexts remains an open question. Our study utilizes datasets in English, limiting our ability to assess the performance and applicability of our methods across non-English languages and datasets that encompass a broader range of cultural knowledge.
\section*{Ethics Statement}
Our research aims to deepen the understanding and enhance the functionality of PLMs by investigating the role of DKNs. While our research seeks to enhance the understanding and functionality of PLMs by exploring the role of DKNs, we are acutely aware of the potential for misuse of these findings. The increased capabilities of PLMs, driven by insights into DKNs, could potentially be exploited for generating misleading information, manipulating public opinion, or other malicious purposes. It is not our intention to facilitate such activities. Instead, our goal is to contribute to the scientific community's knowledge base, enabling the development of more robust, accurate, and ethically aligned language technologies..

We emphasize the responsibility of using these insights ethically, to avoid the exploitation of enhanced PLM capabilities for harmful purposes. To this end, we advocate for transparency in research, collaboration across disciplines for ethical oversight, and the development of regulatory frameworks to ensure that advancements in PLM technology contribute positively to society.
\bibliography{custom}

\appendix

\begin{table*}[h]
\centering
\small
\resizebox{\textwidth}{!}
{
\begin{tabular}{c|ccccccc}
\toprule
& \multicolumn{7}{c}{GPT-2} \\
\multirow{-2}{*}{Method} & 1 & 2 & 3 & 4 & 5 & 6 & 7 \\
\midrule
DBSCAN
& --- & 12 $\rightarrow$ 26 & 17.7 $\rightarrow$ 38.8 & 23.4 $\rightarrow$ 33.8 & 32.9 $\rightarrow$ 53.4 & 24.8 $\rightarrow$ 49.8 & 26.4 $\rightarrow$ 46.6 \\
Hierarchical
& --- & 2.9 $\rightarrow$ 3.8 & 15.5 $\rightarrow$ 7.3 & 3.7 $\rightarrow$ 12 & 4.3 $\rightarrow$ 27 & 25 $\rightarrow$ 92 & --- \\
K-means
& --- & 27.4 $\rightarrow$ 38.3 & 17.8 $\rightarrow$ 29.5 & 27.8 $\rightarrow$ 44.6 & 32.4 $\rightarrow$ 58.7 & 28.9 $\rightarrow$ 30.5 & 21.8 $\rightarrow$ 30.1 \\
AMIG
& --- & -0.79 $\rightarrow$  0.89 & --- & --- & --- & --- & --- \\
\midrule
 \textbf{NTC (Ours)}
& --- & \textbf{7.9 $\rightarrow$ 53} &  \textbf{8.6 $\rightarrow$ 44} &  \textbf{12 $\rightarrow$ 92} &  \textbf{11 $\rightarrow$ 53} &  \textbf{3.1 $\rightarrow$ 32} &  \textbf{7.8 $\rightarrow$ 61} \\
\bottomrule
\toprule
& \multicolumn{7}{c}{GPT-2} \\
\multirow{-2}{*}{Method} & 8 & 9 & 10 & 11 & 12 & 13 & 14 \\
\midrule
DBSCAN
& 9.1 $\rightarrow$ 12.5 & 30.4 $\rightarrow$ 32.9 & 39.8 $\rightarrow$ 19.3 & 11 $\rightarrow$ 22.9 & --- & --- & --- \\
Hierarchical
& --- & --- & --- & --- & --- & --- & --- \\
K-means
& 23.3 $\rightarrow$ 33.6 & 25.1 $\rightarrow$ 58.4 & 18.5 $\rightarrow$ 47.3 & 31 $\rightarrow$ 45 & 10.3 $\rightarrow$ 32.1 & 40.6 $\rightarrow$ 32.7 & 13 $\rightarrow$ 14.8 \\
AMIG
& --- & --- & --- & --- & --- & --- & --- \\

\midrule
 \textbf{NTC (Ours)}
& \textbf{0.6 $\rightarrow$ 1.3} & --- & --- & --- & --- & --- & --- \\
\bottomrule
\toprule
& \multicolumn{6}{c}{GPT-2} \\
\multirow{-2}{*}{Method} & 15 & 16 & 17 & 18 & 19 &  Overall \\
\midrule
DBSCAN
& --- & --- & --- & --- & --- & 23.90 $\rightarrow$ 34.72 \\
Hierarchical
& --- & --- & 20.8 $\rightarrow$ -0.13 & --- & --- & 20.77 $\rightarrow$ 25.05 \\
K-means
& 48.8 $\rightarrow$ 118.1 & 29.2 $\rightarrow$ 74.7 & 62.7 $\rightarrow$ -1 & --- & 37 $\rightarrow$ 26 & 34.42 $\rightarrow$ 38.86 \\
AMIG
& --- & --- & --- & --- & --- & -0.79 $\rightarrow$ 0.89 \\
\midrule
 \textbf{NTC (Ours)}
& --- & --- & --- & --- & --- & \textbf{9.32} $\rightarrow$ \textbf{55.60} \\
\bottomrule
\toprule& \multicolumn{6}{c}{Llama2} \\
\multirow{-2}{*}{Method} & 1 & 2 & 3 & 4 & 5 & 6 \\
\midrule
DBSCAN
& --- & 7.1 $\rightarrow$ 15.4 & 8.7 $\rightarrow$ 15.2 & 11.2 $\rightarrow$ 21.0 & 8.9 $\rightarrow$ 14.6 & 17.6 $\rightarrow$ 34.2 \\
Hierarchical
& --- & 27.6 $\rightarrow$ 44.3 & 22.3 $\rightarrow$ 6.5 & --- & --- & --- \\
K-means
& --- & 2.8 $\rightarrow$ 16.1 & 19.2 $\rightarrow$ 39.1 & 22.2 $\rightarrow$ 45.4 & 25.8 $\rightarrow$ 41.4 & 14.2 $\rightarrow$ 34.2 \\ 
AMIG
& --- & -0.99 $\rightarrow$ 1.99 & --- & --- & --- & ---\\
\midrule
 \textbf{NTC (Ours)}
& --- & \textbf{2.8 $\rightarrow$ 15.6} & \textbf{4.3 $\rightarrow$ 19.1} & \textbf{6.4 $\rightarrow$ 37.4} & \textbf{8.4 $\rightarrow$ 49.9} & \textbf{9.8 $\rightarrow$ 29.6}  \\
\bottomrule
\toprule
& \multicolumn{6}{c}{Llama2} \\
\multirow{-2}{*}{Method} & 7 & 8 & 9 & 10 & 11 & 12 \\
\midrule
DBSCAN
& 7.7 $\rightarrow$ 18.0 & 7.7 $\rightarrow$ 16.3 & 12.4 $\rightarrow$ 28.6 & 12.2 $\rightarrow$ 32.1 & 7.2 $\rightarrow$ 25.3 & 53.6 $\rightarrow$ 125.6 \\
Hierarchical
 & --- & --- & --- & --- & --- & ---\\
K-means
& 28.6 $\rightarrow$ 50.8 & 37.9 $\rightarrow$ 97.1 & 29.4 $\rightarrow$ 74.9 & 47.9 $\rightarrow$ 87.8 & 50.7 $\rightarrow$ 110.5 & 13.5 $\rightarrow$ 23.2\\
AMIG
 & --- & --- & --- & --- & --- & --- \\
\midrule
 \textbf{NTC (Ours)}
& \textbf{9.1 $\rightarrow$ 27.6} & \textbf{7.8 $\rightarrow$ 50.1} & \textbf{7.6 $\rightarrow$ 21.5} & \textbf{9.1 $\rightarrow$ 31.5} & \textbf{4.6 $\rightarrow$ 25.7} & \textbf{15.2 $\rightarrow$ 125.5} \\
\bottomrule
\toprule
& \multicolumn{6}{c}{Llama2} \\
\multirow{-2}{*}{Method}  & 13 & 14 & 15 & 16 & 17 &  Overall \\
\midrule
DBSCAN
& 92.5 $\rightarrow$ 29.4 & --- & --- & 0.2 $\rightarrow$ 0 & --- & 22.26 $\rightarrow$ 18.24 \\
Hierarchical
 & 9.7 $\rightarrow$ 9.1 & 36.3 $\rightarrow$ 20.2 & --- & --- & --- & 27.50 $\rightarrow$ 44.04 \\
K-means
 & 19.7 $\rightarrow$ 19.2 & 36.3 $\rightarrow$ 50.0 & 68.8 $\rightarrow$ 375.0 & 3.3 $\rightarrow$ 11.1 & 4.9 $\rightarrow$ 17.9 & 20.08 $\rightarrow$ 39.24 \\
AMIG
 & --- & --- & --- & --- & --- & -0.99 $\rightarrow$ 1.99 \\
\midrule
 \textbf{NTC (Ours)}
& \textbf{15.8 $\rightarrow$ 183.7} & \textbf{13.6 $\rightarrow$ 37.7} & \textbf{3.1 $\rightarrow$ 17.2} & \textbf{1.0 $\rightarrow$ 0.5} & \textbf{31.2 $\rightarrow$ 134.9} & \textbf{14.11} $\rightarrow$ \textbf{28.78} \\
\bottomrule
\end{tabular}
}
\caption{The full results of Comparison of the degeneracy properties of DKN using different methods.  ``---'' denotes methods that do not yield $\mathcal{D}$ of a specific length.}
\label{tab:appendix:NTC}
\end{table*}
\section{Experimental Hyperparameters}\label{appendix:Experimental hyperparameters}
\subsection{Hardware spcification and environment.} 
We ran our experiments on the machine equipped with the following specifications:
\begin{itemize}
\item CPU: Intel(R) Xeon(R) CPU E5-2680 v4 @ 2.40GHz, Total CPUs: 56
\item GPU: NVIDIA GeForce RTX 3090, 24576 MiB (10 units)
\item Software: 
  \begin{itemize}
  \item Python Version: 3.10.10
  \item PyTorch Version: 2.0.0+cu117
  \end{itemize}
\end{itemize}
\begin{figure*}[t]
    \centering
    \includegraphics[width=\linewidth, trim={0cm 0cm 0cm 0cm}, clip]{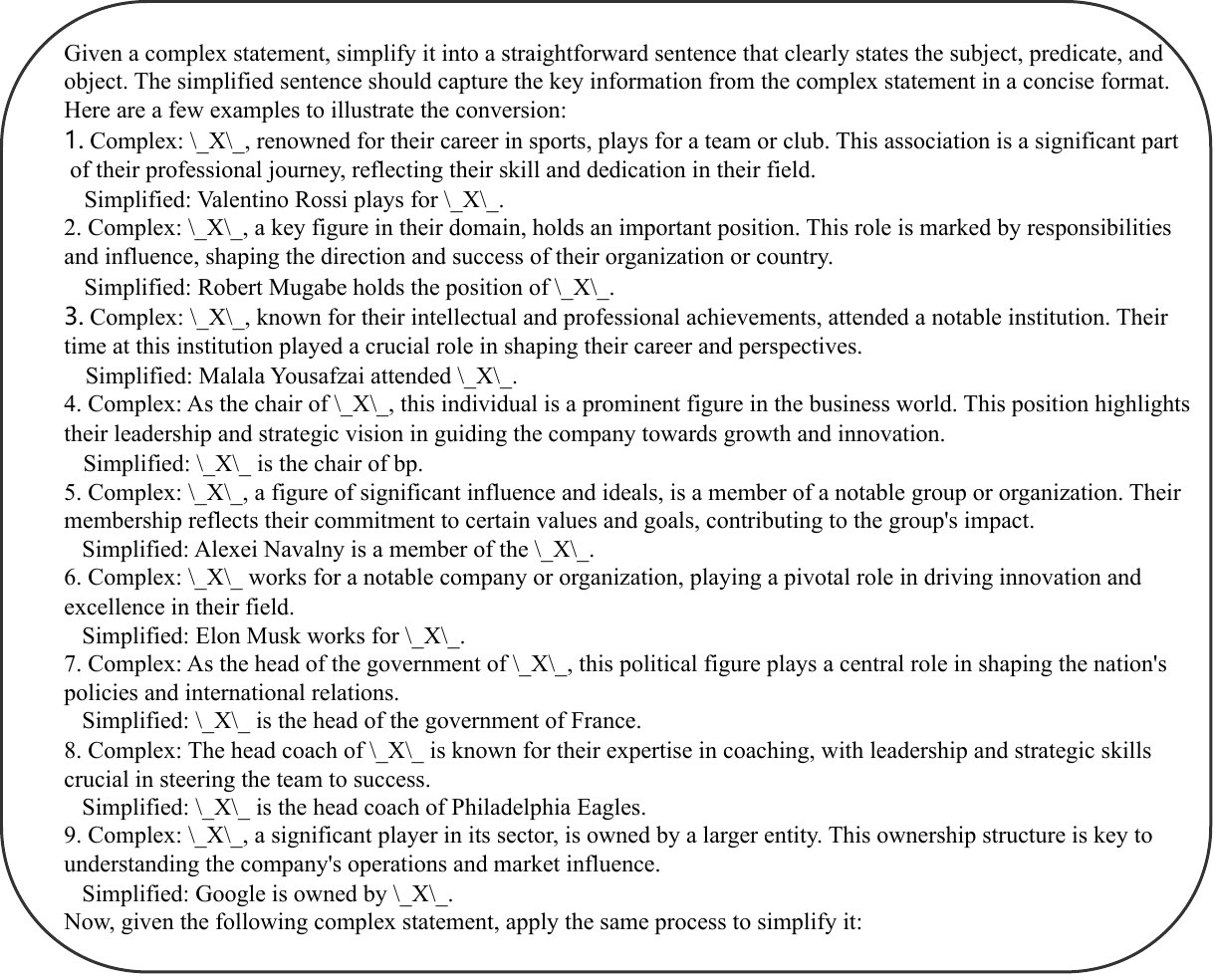} 
    \caption{The prompt of fact-checking experiments on complex texts.}
    \label{fig:appendix-prompt}
\end{figure*}
\subsection{Experimental Hyperparameters of Neurological Topology Clustering}
\label{appendix-ntc-hyper}
In Section \ref{section:method-NTC}, where we obtain degenerate knowledge neurons, the primary hyperparameters are \(\tau_1\) and \(\tau_2\). First, \(\tau_1\) is a dynamic threshold, set as \begin{equation}
    \tau_1 = 0.5 \times \max \left( R_{\text{persist}}(\mathcal{B}_1), 
    \ldots, R_{\text{persist}}(\mathcal{B}_n) \right)
\end{equation}
Then, \(\tau_2\) is a fixed value.
\begin{equation}
\tau_2 = 0.3
\end{equation} In this experiment, some data led to excessively large changes in predictive probability, indicating that the PLMs had not originally mastered this factual knowledge, resulting in a very low initial predictive probability. To study the storage mechanism of factual knowledge, it's essential to investigate facts already grasped by the model. Therefore, we set a threshold to exclude data that caused extreme changes in predictive probability. If the change in predictive probability \(\Delta Prob\) satisfies:
\begin{equation}
    \Delta Prob > 900
\end{equation}
then that data is excluded.
\subsection{Experimental Hyperparameters of Degneracy and Robustness}
In Section \ref{section:Robustness}, the first experiment under Query-Perturbation, namely the Suppressing DKN experiment, similar to \ref{appendix-ntc-hyper}, excludes data that satisfies the condition:
\begin{equation}
     Prob_\text{sup} > 900
\end{equation}
In the Fact-Checking experiment, \(\tau_3\) is similar to \(\tau_1\) as a dynamic threshold. We first count the total number of neurons, then set
\begin{equation}
    \tau_3 = 0.7 \times N_{\text{total}}
\end{equation}
where \(N_{\text{total}}\) represents the total number of neurons.

\subsection{Experimental Hyperparameters of Degeneracy and Evolvability}
In Section \ref{section:Evolvability}, the experimental hyperparameter \(\tau_{\Delta N}\) for the Overlap of DKN and Parameter Changes experiment is set as a dynamic threshold. The process involves calculating the maximum value of \(\Delta P(n)\) according to Equation \ref{equation:overlap-change}. Once this value is determined, \(\tau_{\Delta N}\) is set differently based on the model in use.
For the GPT-2 model, the threshold \(\tau_{\Delta N}\) is calculated as:
\begin{equation}
\small
\tau_{\Delta N} = 0.04 \times \max(\Delta P(n_1), \Delta P(n_2), \ldots,  \Delta P(n_k))
\end{equation}
In contrast, for the Llama2 model, the calculation of the threshold \(\tau_{\Delta N}\) is slightly adjusted:
\begin{equation}
\small
\tau_{\Delta N} = 0.05 \times \max(\Delta P(n_1), \Delta P(n_2), \ldots,  \Delta P(n_k))
\end{equation}
This distinction in the calculation of \(\tau_{\Delta N}\) reflects the specific characteristics and performance considerations of each model.

\subsection{Experimental Hyperparameters of Degneracy and  Complexity}
In Section \ref{section:Complexity}, during our fact-checking experiments on complex texts in Section \ref{section:Complexity}, we provide PLMs with a prompt requiring them to first extract factual knowledge in the form of triples and then perform fact-checking using DKNs. Our prompt is as follows:
\subsection{Fact-Checking of Complex Texts in Section \ref{section:Complexity}}
During our fact-checking experiments on complex texts in Section \ref{section:Complexity}, we prompt PLMs to simplify complex statements into straightforward sentences. Our prompt is shown in Figure \ref{fig:appendix-prompt}.
\section{Knowldege Localization}
\label{appendix: amig}
This section introduces the method we use to acquire knowledge neurons. We employ the approach proposed by \citeauthor{chen2023journey}\shortcite{chen2023journey}, which we will detail below.

Given a query \(q\), we can define the probability of the correct answer predicted by a PLMs as follows:
\begin{equation}
\label{eq:1}
    \operatorname{F}(\hat{w}^{(l)}_{j}) = p(y^* | q, w^{(l)}_{j}=\hat{w}^{(l)}_{j})
\end{equation}
Here, \(y^*\) represents the correct answer, \(w^{(l)}_{j}\) denotes the \(j\)-th neuron in the \(l\)-th layer, and \(\hat{w}^{(l)}_{j}\) is the specific value assigned to \(w^{(l)}_{j}\). To calculate the attribution score for each neuron, we employ the technique of integrated gradients.

To compute the attribution score of a neuron \(w^{(l)}_{j}\), we consider the following formulation:
\begin{equation}
\resizebox{\linewidth}{!}{$
    \label{eqution:attribute}
     \operatorname{Attr}({w}^{(l)}_{j}) = (\overline{w}^{(l)}_{j} - {w'}^{(l)}_{j}) \int_{0}^{1} \frac{\partial \operatorname{F}({w'}^{(l)}_{j} + \alpha(\overline{w}^{(l)}_{j} - {w'}^{(l)}_{j}))}{\partial {w}^{(l)}_{j}}  \, d\alpha
$}
\end{equation}Here, \(\overline{w}^{(l)}_{j}\) represents the actual value of \(w^{(l)}_{j}\),
\(w'^{(l)}_{j}\) serves as the baseline vector for \(w^{(l)}_{j}\). The term \(\frac{\partial \operatorname{F}(w'^{(l)}_{j} + \alpha(w^{(l)}_{j} - w'^{(l)}_{j}))}{\partial w^{(l)}_{j}}\) computes the gradient with respect to \(w^{(l)}_{j}\).  

Next, we aim to obtain ${w'}^{(l)}_{j}$. Starting from the sentence $q$, we acquire a baseline sentence and then encode this sentence as a vector.

Let the baseline sentence corresponding to $q_i$ be $q'_i$, and $q'_i$ consists of $m$ words, maintaining a length consistent with $q$, denoted as $q'_i=(q'_{i1} \ldots q'_{ik} \ldots q'_{im})$. Since we are using auto-regressive models, according to \citeauthor{chen2023journey}\shortcite{chen2023journey}' method, $q'_{ik}=\langle \text{eos}\rangle$, where $\langle \text{eos}\rangle$ represents ``end of sequence'' in auto-regressive models.

The attribution score \(Attr_i(w_j^{(l)})\) for each neuron, given the input \(q_i\), can be determined using Equation \eqref{eqution:attribute}. For the computation of the integral, the Riemann approximation method is employed:
\begin{equation}
    \resizebox{\linewidth}{!}{$
    {Attr_i(w_j^l)} \approx \frac{\overline{w}^{(l)}_{j}}{N} \sum_{k=1}^{N} \frac{ \partial F({w'}^{(l)}_{j} + \frac{k}{N} \times (\overline{w}^{(l)}_{j} - {w'}^{(l)}_{j})}{\partial {w}^{(l)}_{j}}
    $}
\end{equation}where $N$ is the number of approximation steps.

Then, the attribution scores for each word \(q_i\) are aggregated and subsequently normalized:
\begin{equation}
    Attr(w_j^l) = \frac{\sum_{i=1}^{m} Attr_i(w_j^l)}{\sum_{j=1}^{n} \sum_{i=1}^{m} Attr_i(w_j^l)},
\end{equation}


Let \( \mathcal{N} \) be the set of neurons classified as knowledge neurons based on their attribution scores exceeding a predetermined threshold \( \tau \), for a given input \( q \). This can be formally defined as:

\begin{equation}
\mathcal{N} = \left\{ w_j^{(l)} \,\middle|\, Attr(w_j^{(l)}) > \tau \right\}
\end{equation}where \(l\) encompassing all layers and \(j\) including all neurons within each layer.

\section{Persistent homology} \label{appendix:persistent-homology}
Persistent homology is a method for computing topological features of a space at different spatial resolutions. More persistent features are detected over a wide range of spatial scales and are deemed more likely to represent true features of the underlying space rather than artifacts of sampling, noise, or particular choice of parameters \cite{carlsson2009topology-appendix}.

To find the persistent homology of a space, the space must first be represented as a simplicial complex. A distance function on the underlying space corresponds to a filtration of the simplicial complex, that is a nested sequence of increasing subsets. One common method of doing this is via taking the sublevel filtration of the distance to a point cloud, or equivalently, the offset filtration on the point cloud and taking its nerve in order to get the simplicial filtration known as Čech filtration \cite{kerber2013approximate-appendix}. A similar construction uses a nested sequence of Vietoris–Rips complexes known as the Vietoris–Rips filtration \cite{dey2019simba-appendix}.

\subsection{Definition}
In persistent homology, formally, we consider a real-valued function defined on a simplicial complex, denoted as \( f:K \rightarrow \mathbb{R} \). This function is required to be non-decreasing on increasing sequences of faces, meaning that for any two faces \( \sigma \) and \( \tau \) in \( K \), if \( \sigma \) is a face of \( \tau \), then \( f(\sigma) \leq f(\tau) \).

For every real number \( a \), the sublevel set \( K_a = f^{-1}((-\infty, a]) \) forms a subcomplex of \( K \). The values of \( f \) on the simplices in \( K \) create an ordering of these sublevel complexes, which leads to a filtration:
\begin{equation}
\emptyset = K_0 \subseteq K_1 \subseteq \cdots \subseteq K_n = K
\end{equation} 
Within this filtration, for \( 0 \leq i \leq j \leq n \), the inclusion \( K_i \hookrightarrow K_j \) induces a homomorphism on the simplicial homology groups for each dimension \( p \), noted as \( f_p^{i,j}:H_p(K_i) \rightarrow H_p(K_j) \). The \( p^\text{th} \) persistent homology groups are the images of these homomorphisms, and the \( p^\text{th} \) persistent Betti numbers \( \beta_p^{i,j} \) are defined as the ranks of these groups \cite{edelsbrunner2022computational-appendix}.
Persistent Betti numbers for \( p=0 \) coincide with the size function, an earlier concept related to persistent homology \cite{verri1993use-appendix}.

The concept extends further to any filtered complex over a field \(F\). Such a complex can be transformed into its canonical form, which is a direct sum of filtered complexes of two types: one-dimensional complexes with trivial differential (expressed as \( d(e_{t_i})=0 \)) and two-dimensional complexes with trivial homology (expressed as \( d(e_{s_j+r_j})=e_{r_j} \)) \cite{barannikov1994framed-appendix}.

A persistence module over a partially ordered set \( P \) consists of a collection of vector spaces \( U_t \), indexed by \( P \), along with linear maps \( u_t^s: U_s \to U_t \) for \( s \leq t \). This module can be viewed as a functor from \( P \) to the category of vector spaces or \( R \)-modules.
Persistence modules over a field \( F \) indexed by \( \mathbb{N} \) can be expressed as:
\begin{equation}
\small
    U \simeq \bigoplus_i x^{t_i} \cdot F[x] \oplus \left(\bigoplus_j x^{r_j} \cdot (F[x]/(x^{s_j}\cdot F[x]))\right)
\end{equation} 
Here, multiplication by \( x \) represents a forward step in the persistence module. The free parts correspond to homology generators that appear at a certain filtration level and persist indefinitely, whereas torsion parts correspond to those that appear at a filtration level and last for a finite number of steps \cite{barannikov1994framed-appendix, zomorodian2004computing}.

This framework allows the unique representation of the persistent homology of a filtered simplicial complex using either a persistence barcode or a persistence diagram. In the barcode, each persistent generator is represented by a line segment starting and ending at specific filtration levels, while in the diagram, each generator is represented as a point with coordinates indicating its birth and death times. Barannikov's canonical form offers an equivalent representation.
\subsection{Stability}

The stability of persistent homology is a key attribute, particularly in its application to data analysis, as it ensures robustness against small perturbations or noise in the data \cite{cohen2005stability}. This stability is quantitatively defined in terms of the \textbf{bottleneck distance}, a metric for comparing persistence diagrams.

The bottleneck distance between two persistence diagrams \(X\) and \(Y\) is defined as:
\begin{equation}
W_\infty(X,Y) := \inf_{\varphi: X \to Y} \sup_{x \in X} \Vert x-\varphi(x) \Vert_\infty
\end{equation}where the infimum is taken over all bijections \( \varphi \) from \( X \) to \( Y \). This metric essentially measures the greatest distance between matched points (or generators) in two persistence diagrams, considering the optimal matching.

A fundamental result in the theory of persistent homology is that small changes in the input data (such as a filtration of a space) result in small changes in the corresponding persistence diagram, as measured by the bottleneck distance. This is formalized by considering a space \( X \), homeomorphic to a simplicial complex, with a filtration determined by the sublevel sets of a continuous tame function \( f:X \rightarrow \mathbb{R} \). The map \( D \) that takes the function \( f \) to the persistence diagram of its \( k \)th homology is 1-Lipschitz with respect to the supremum norm on functions and the bottleneck distance on persistence diagrams. Formally, this is expressed as \cite{cohen2005stability}:
\begin{equation}
W_\infty(D(f),D(g)) \leq \lVert f-g \rVert_\infty
\end{equation} 
This Lipschitz condition implies that a small change in the function \( f \), as measured by the supremum norm, will not cause a disproportionately large change in the persistence diagram. Consequently, persistent homology is particularly useful in applications where data may be subject to noise or small variations, as the essential topological features (captured by the persistence diagrams) are not overly sensitive to such perturbations.
\subsection{Computation}

There are various software packages for computing persistence intervals of a finite filtration \cite{otter2017roadmap}. The principal algorithm is based on the bringing of the filtered complex to its canonical form by upper-triangular matrices \cite{barannikov1994framed-appendix}.

\section{Complete Experimental Results of Neurological Topology Clustering} \label{appendix:NTC}
In Section \ref{section:method-NTC}, it is mentioned that the lengths of DKNs vary. For ease of presentation, we have only shown cases with a larger amount of data. Here, we provide the complete table, as shown in Table \ref{tab:appendix:NTC}, which is generated under the same settings as used for Table \ref{table-1-dkn-result}. 

\begin{figure}[h]
\centering
\begin{subfigure}{.5\linewidth}
\centering
\includegraphics[width=\linewidth]{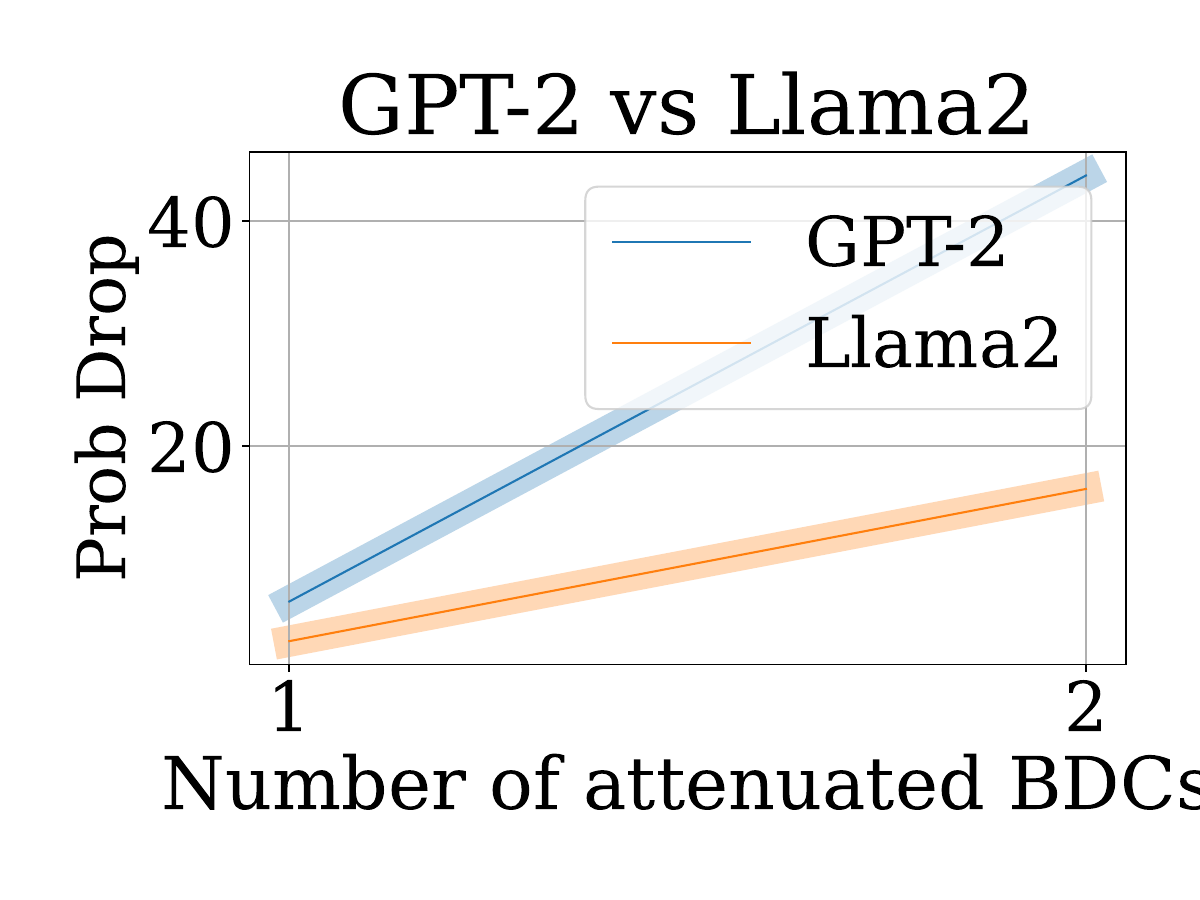}
\end{subfigure}%
\begin{subfigure}{.5\linewidth}
\centering
\includegraphics[width=\linewidth]{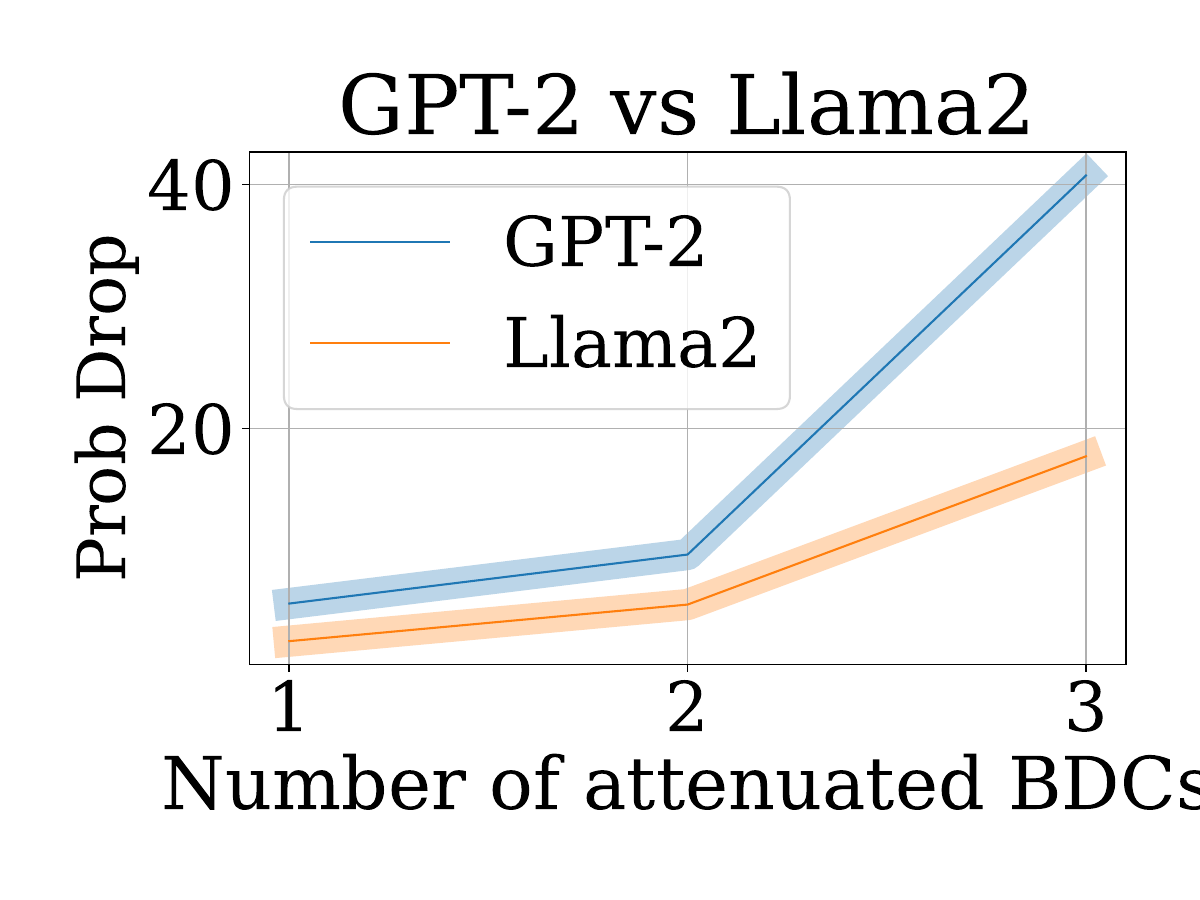}
\end{subfigure}

\begin{subfigure}{.5\linewidth}
\centering
\includegraphics[width=\linewidth]{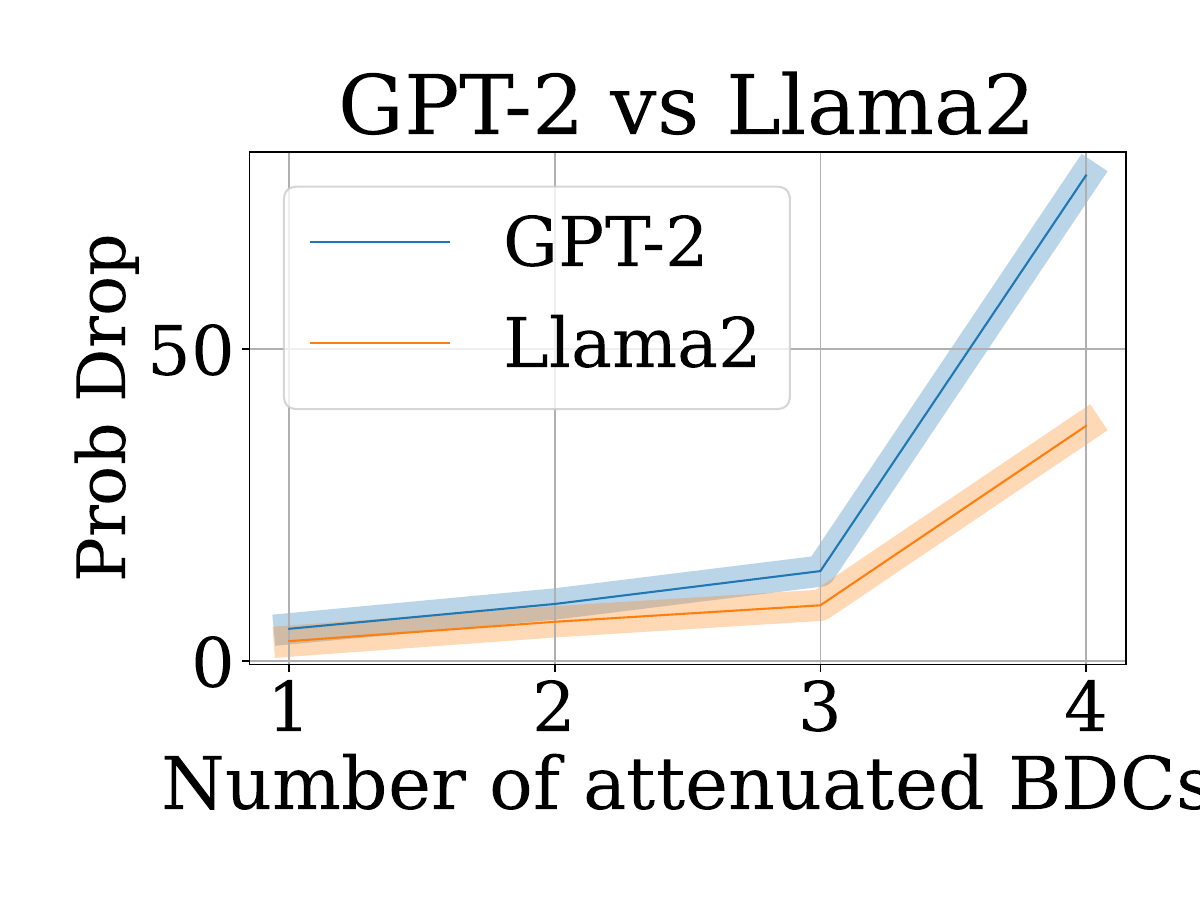}
\end{subfigure}%
\begin{subfigure}{.5\linewidth}
\centering
\includegraphics[width=\linewidth]{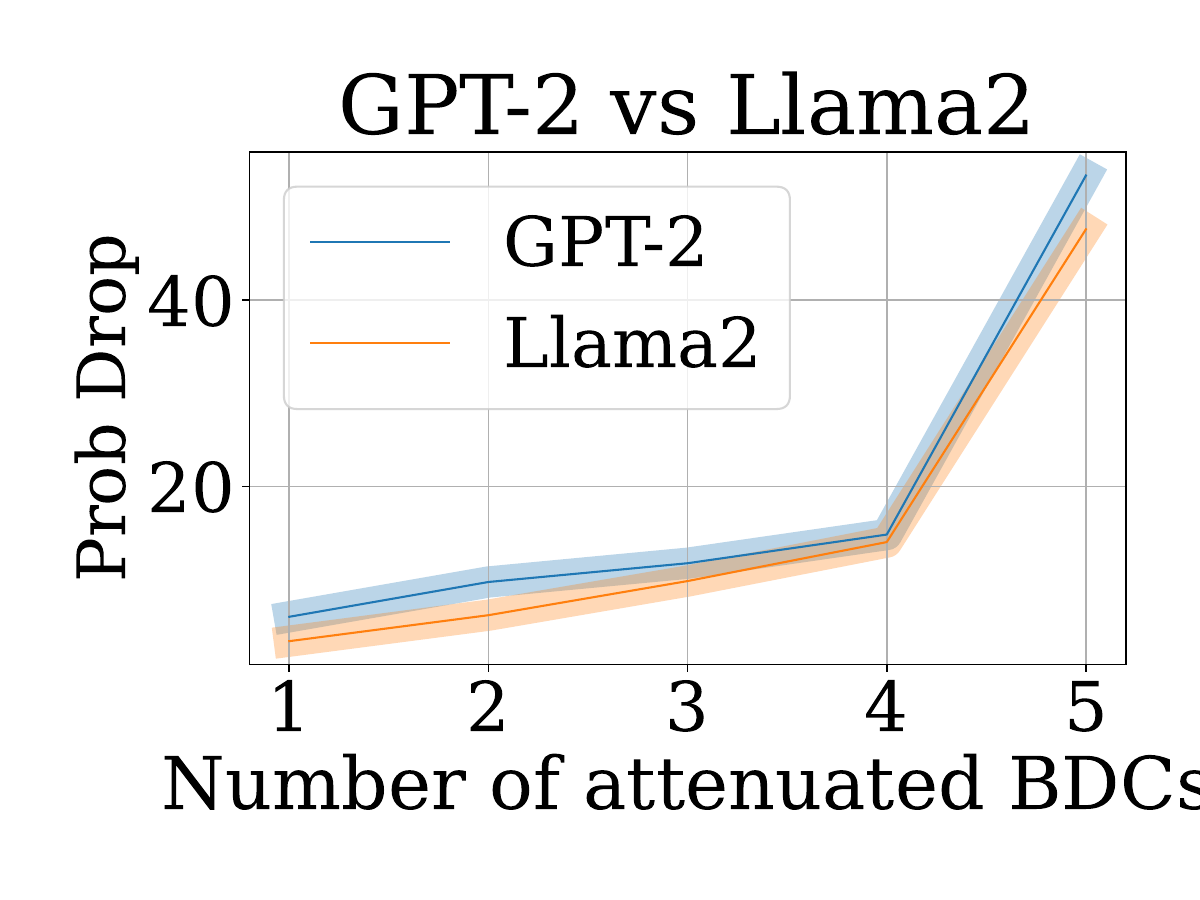}
\end{subfigure}

\begin{subfigure}{.5\linewidth}
\centering
\includegraphics[width=\linewidth]{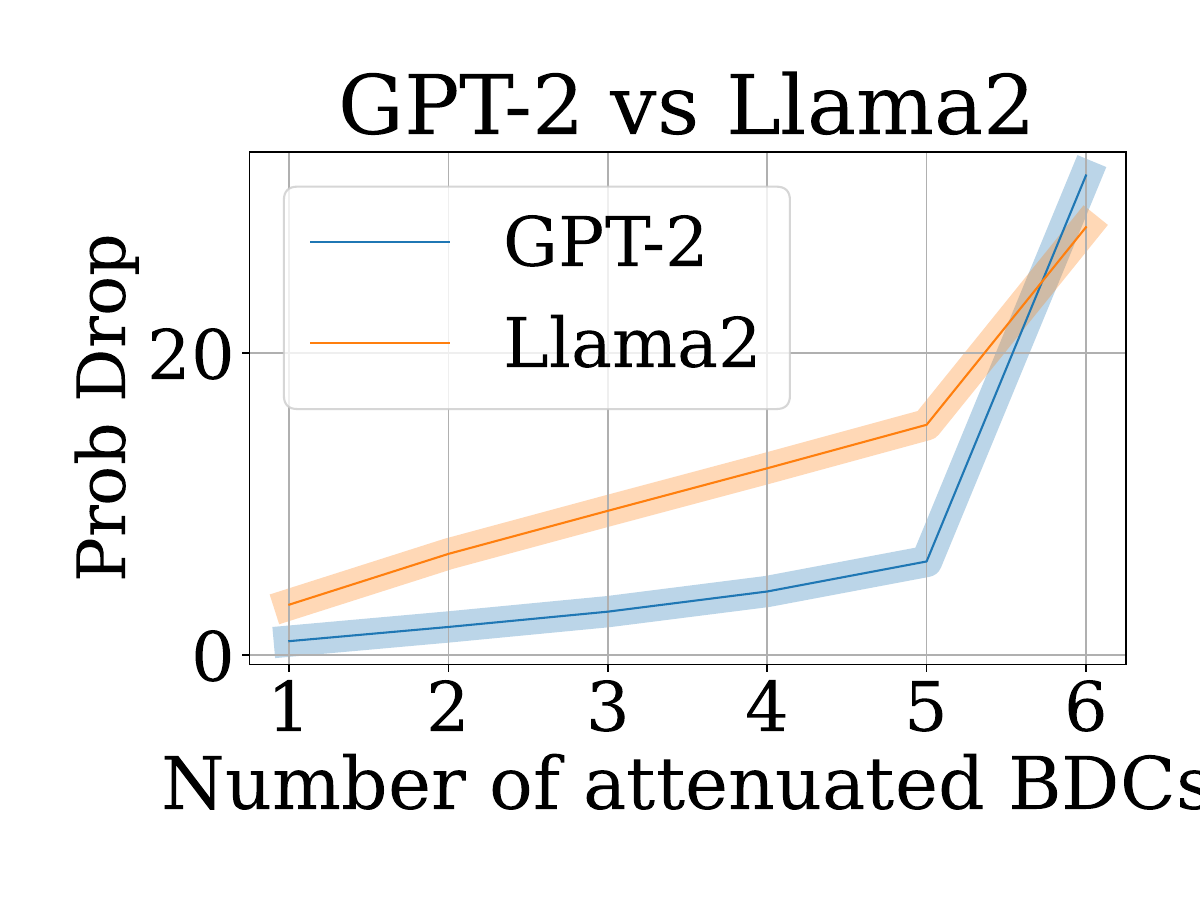}
\end{subfigure}%
\begin{subfigure}{.5\linewidth}
\centering
\includegraphics[width=\linewidth]{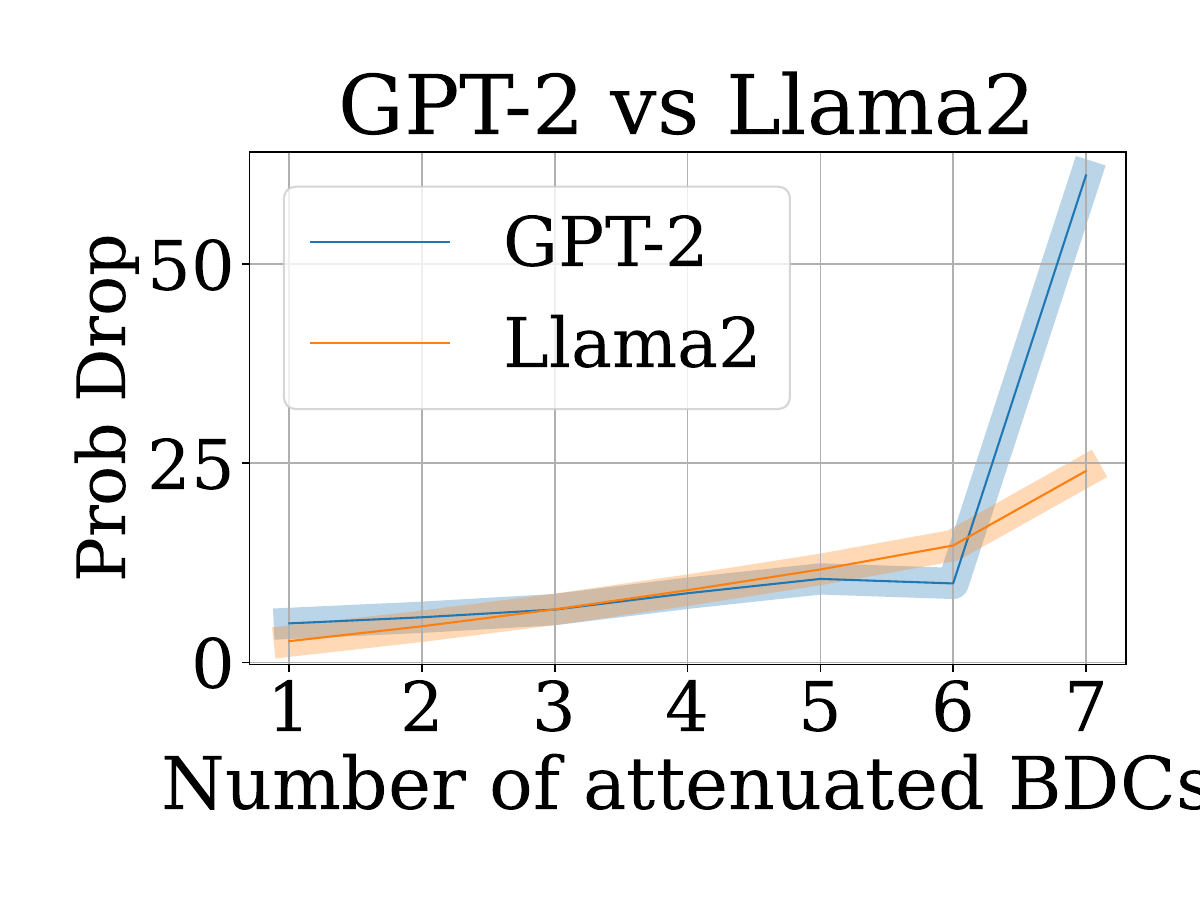}
\end{subfigure}

\begin{subfigure}{.5\linewidth}
\centering
\includegraphics[width=\linewidth]{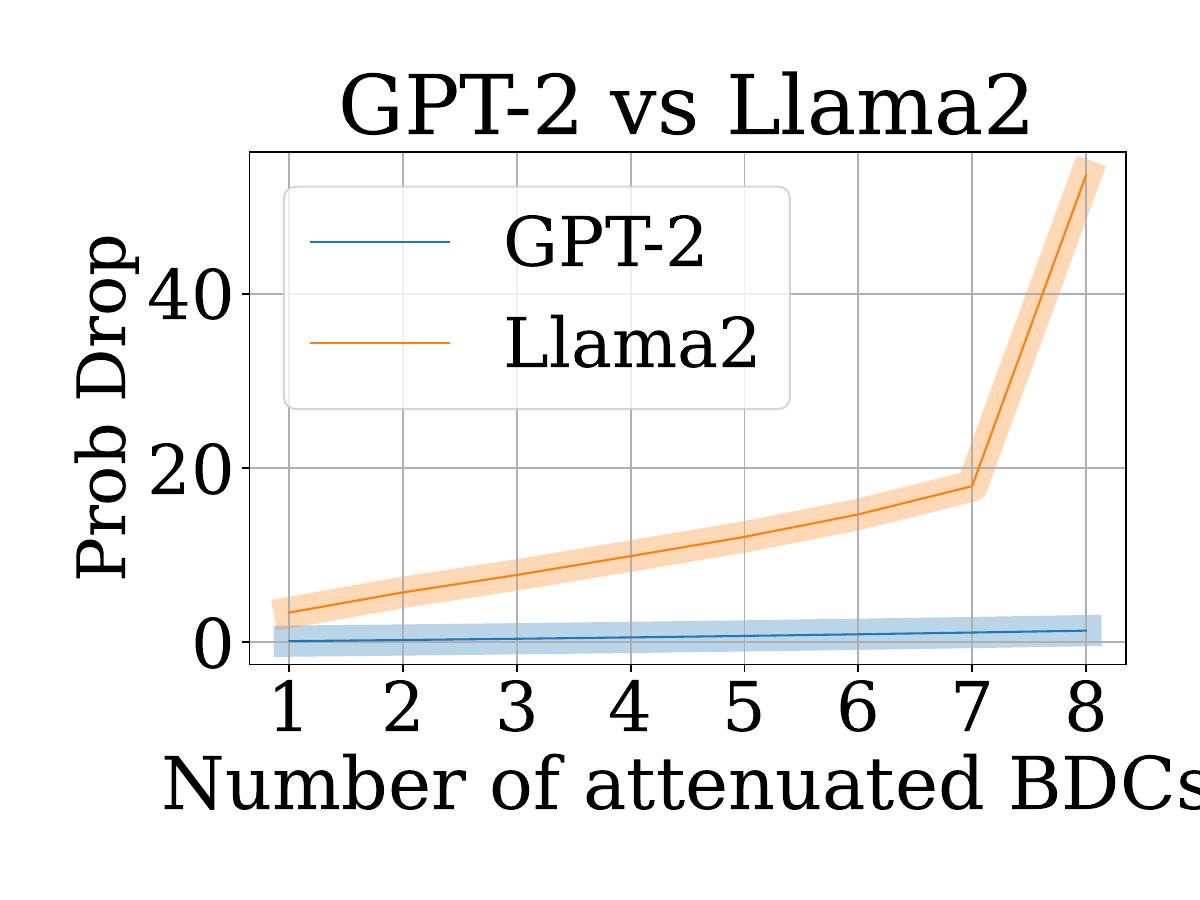}
\end{subfigure}%
\begin{subfigure}{.5\linewidth}
\centering
\includegraphics[width=\linewidth]{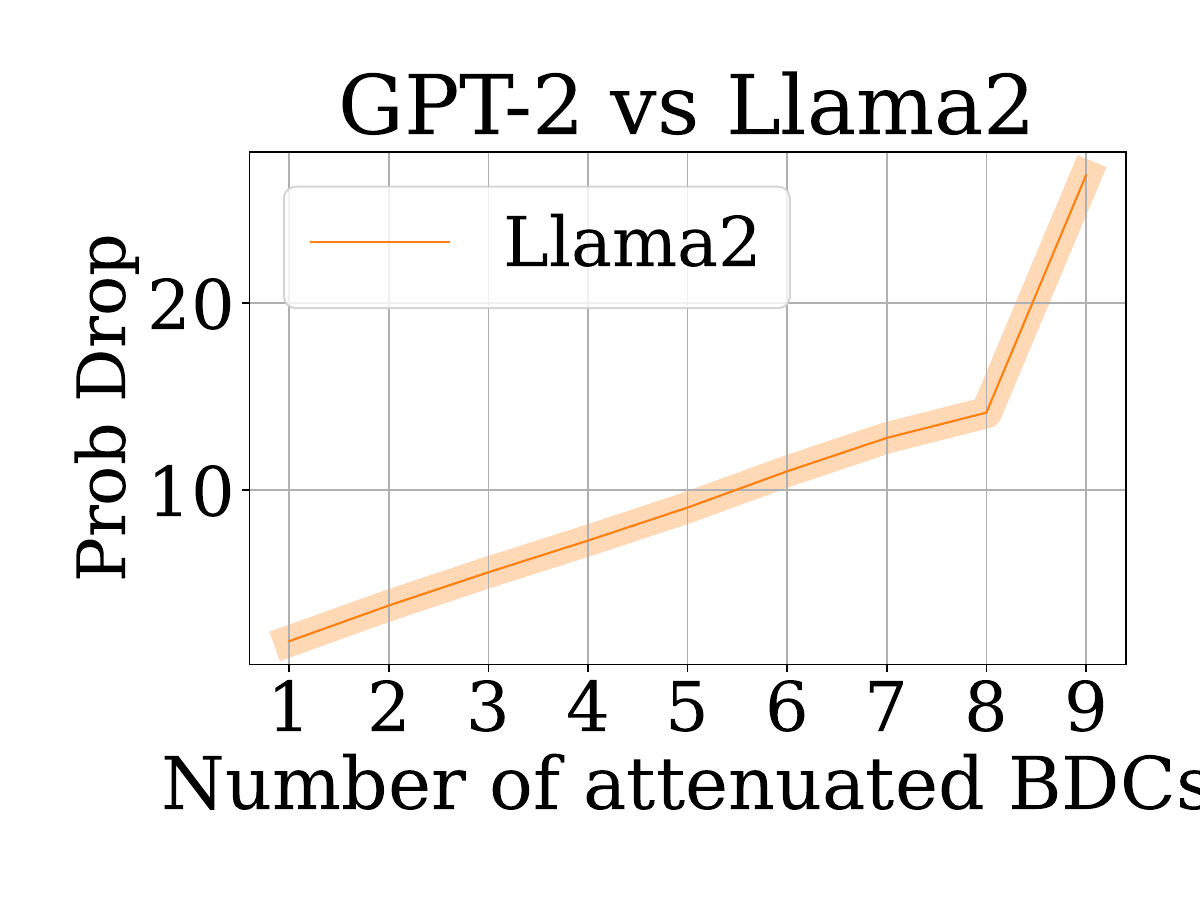}
\end{subfigure}

\begin{subfigure}{.5\linewidth}
\centering
\includegraphics[width=\linewidth]{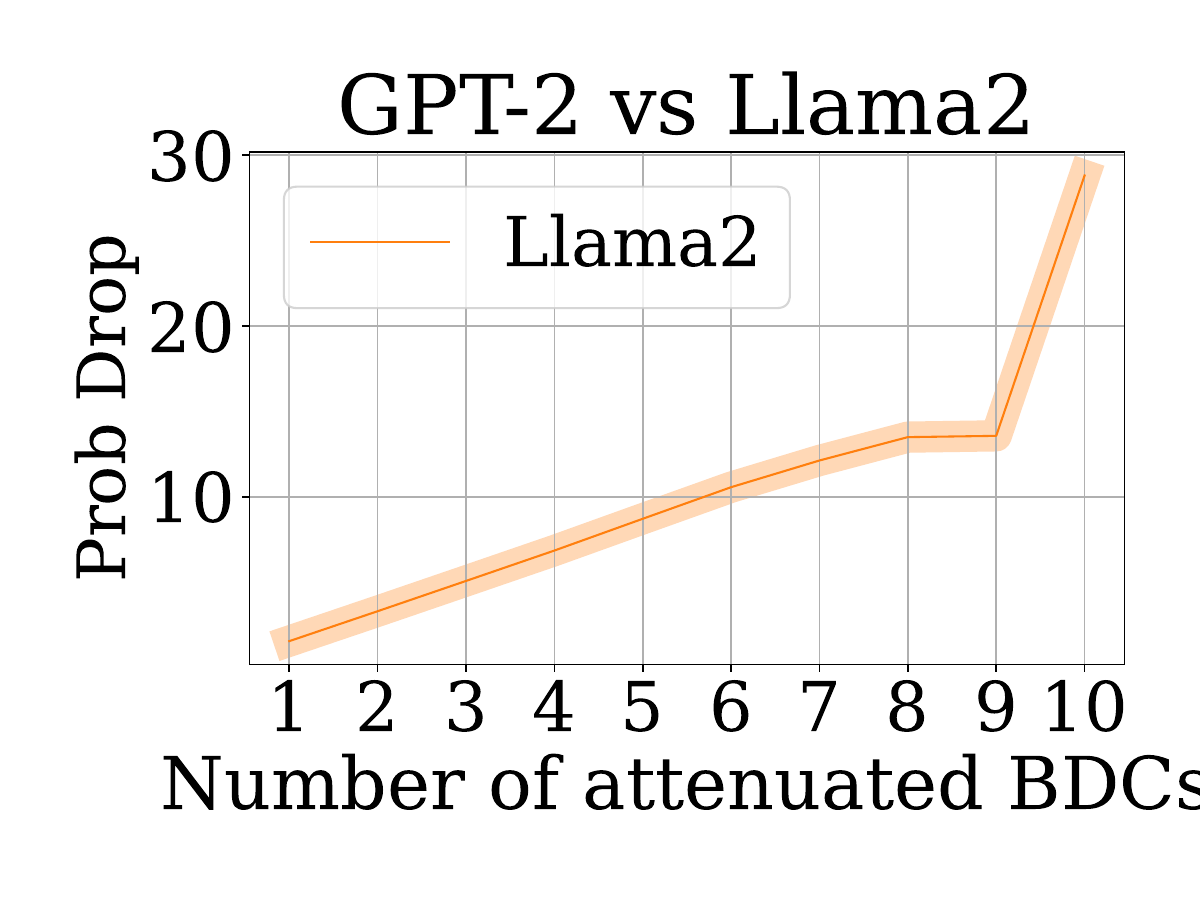}
\end{subfigure}%
\begin{subfigure}{.5\linewidth}
\centering
\includegraphics[width=\linewidth]{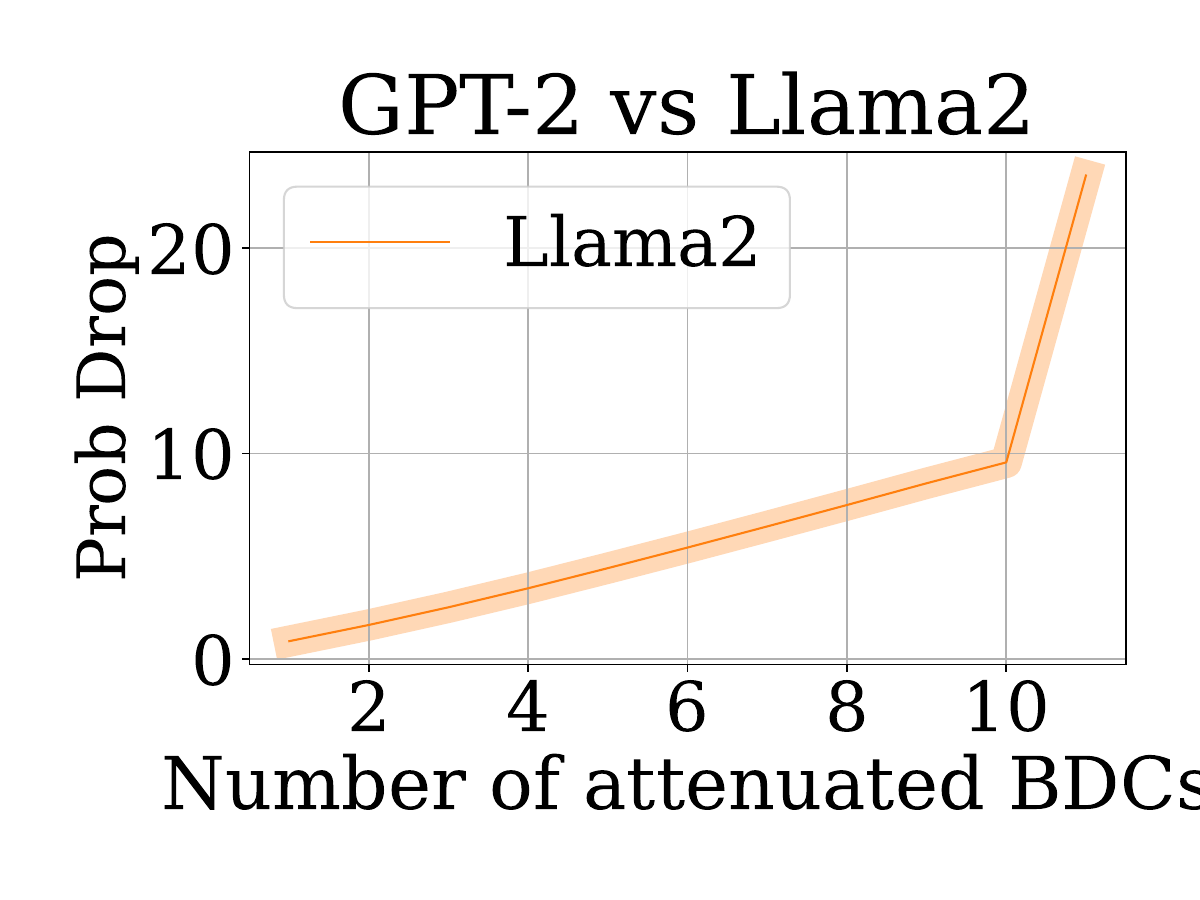}
\end{subfigure}

\begin{subfigure}{.5\linewidth}
\centering
\includegraphics[width=\linewidth]{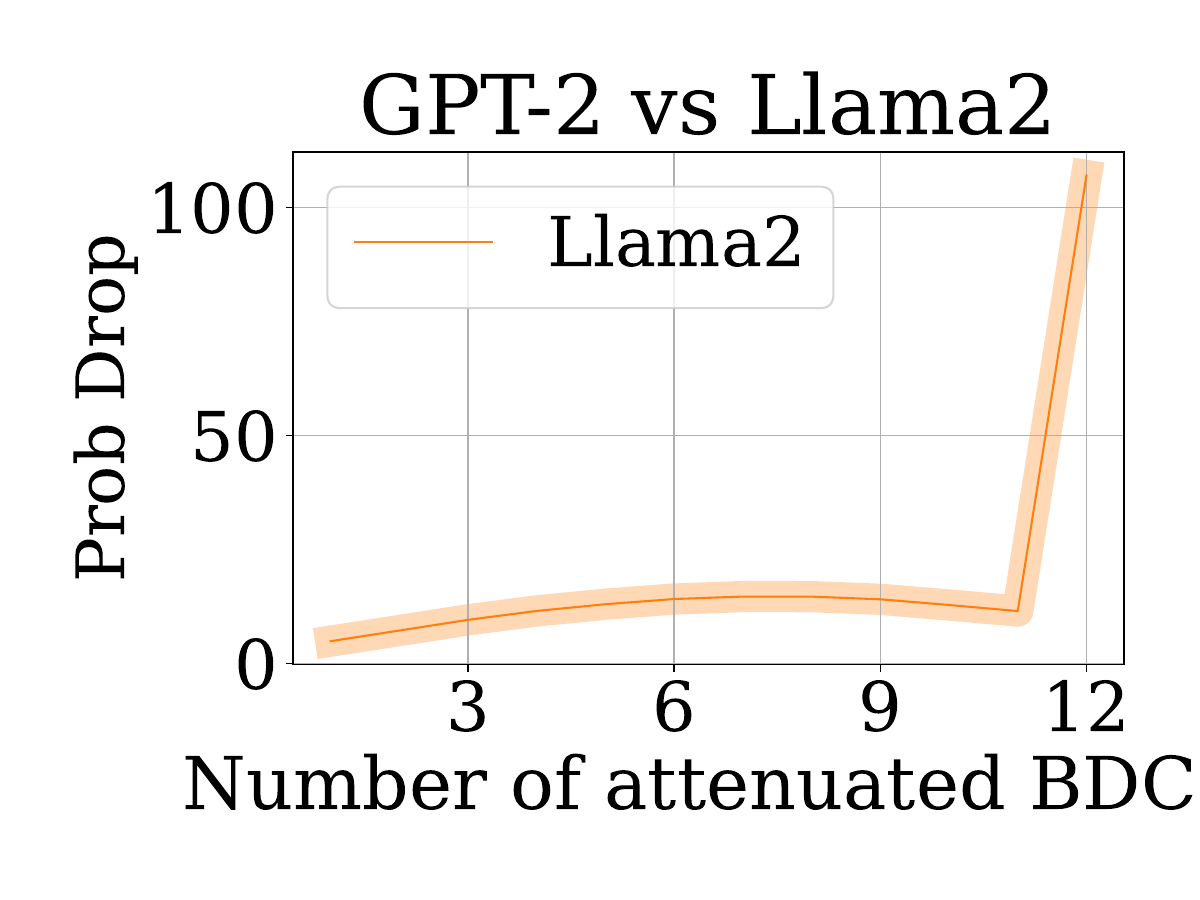}
\end{subfigure}%
\begin{subfigure}{.5\linewidth}
\centering
\includegraphics[width=\linewidth]{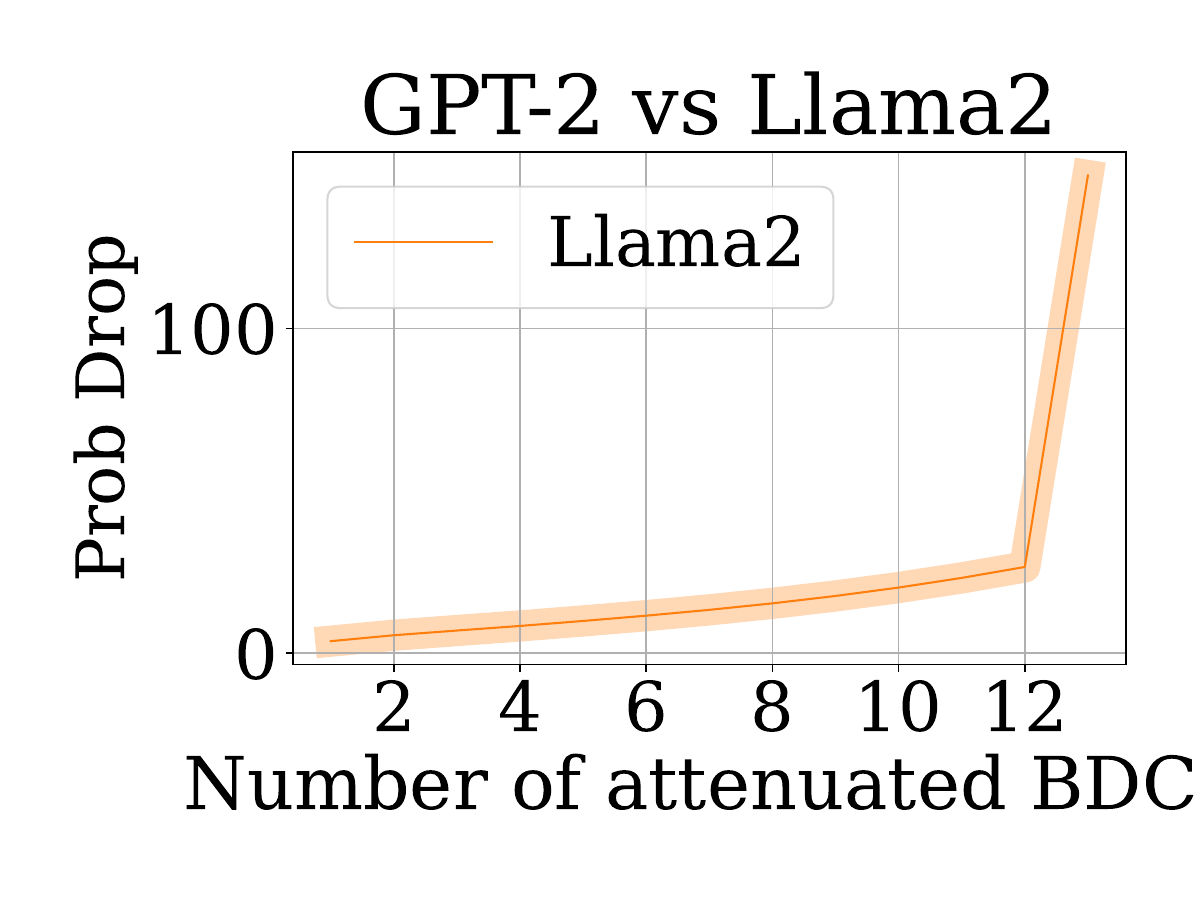}
\end{subfigure}

\begin{subfigure}{.5\linewidth}
\centering
\includegraphics[width=\linewidth]{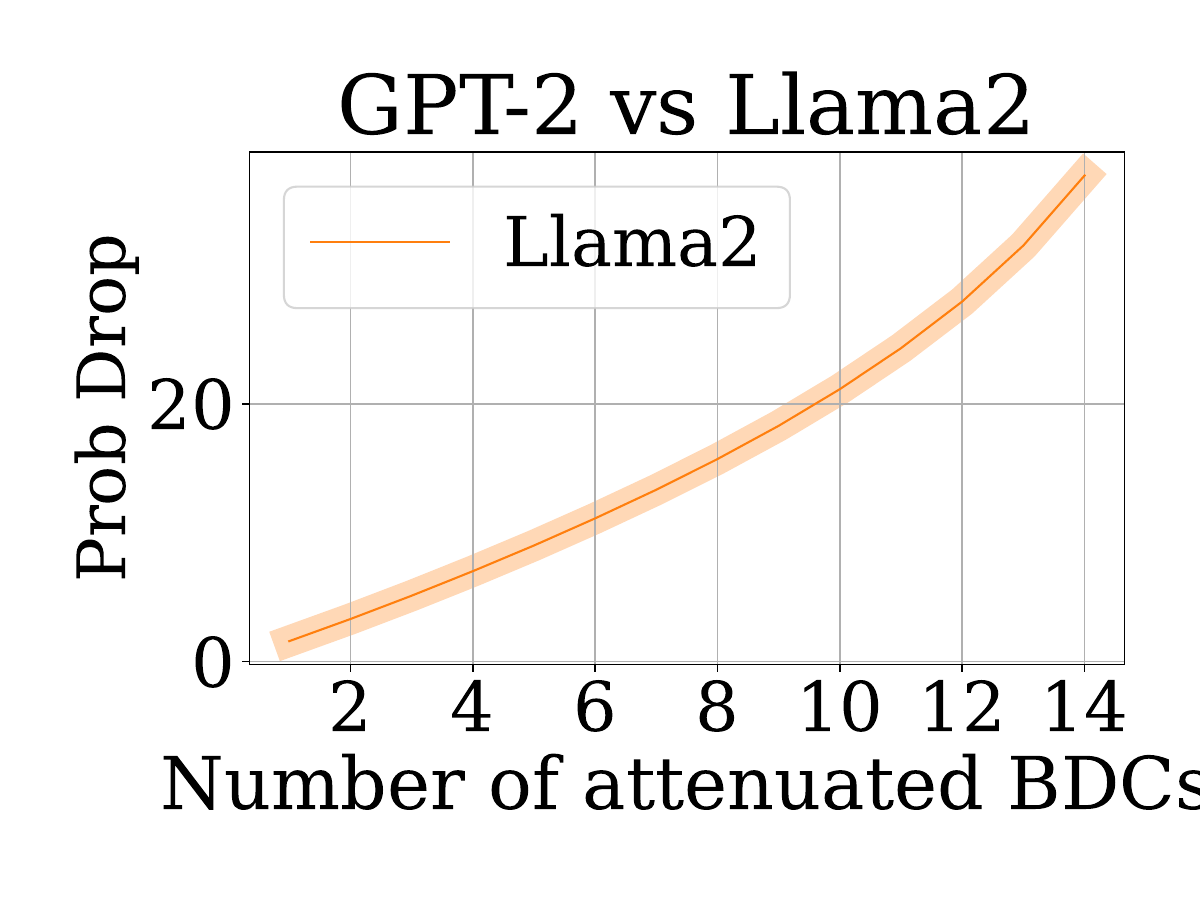}
\end{subfigure}%
\begin{subfigure}{.5\linewidth}
\centering
\includegraphics[width=\linewidth]{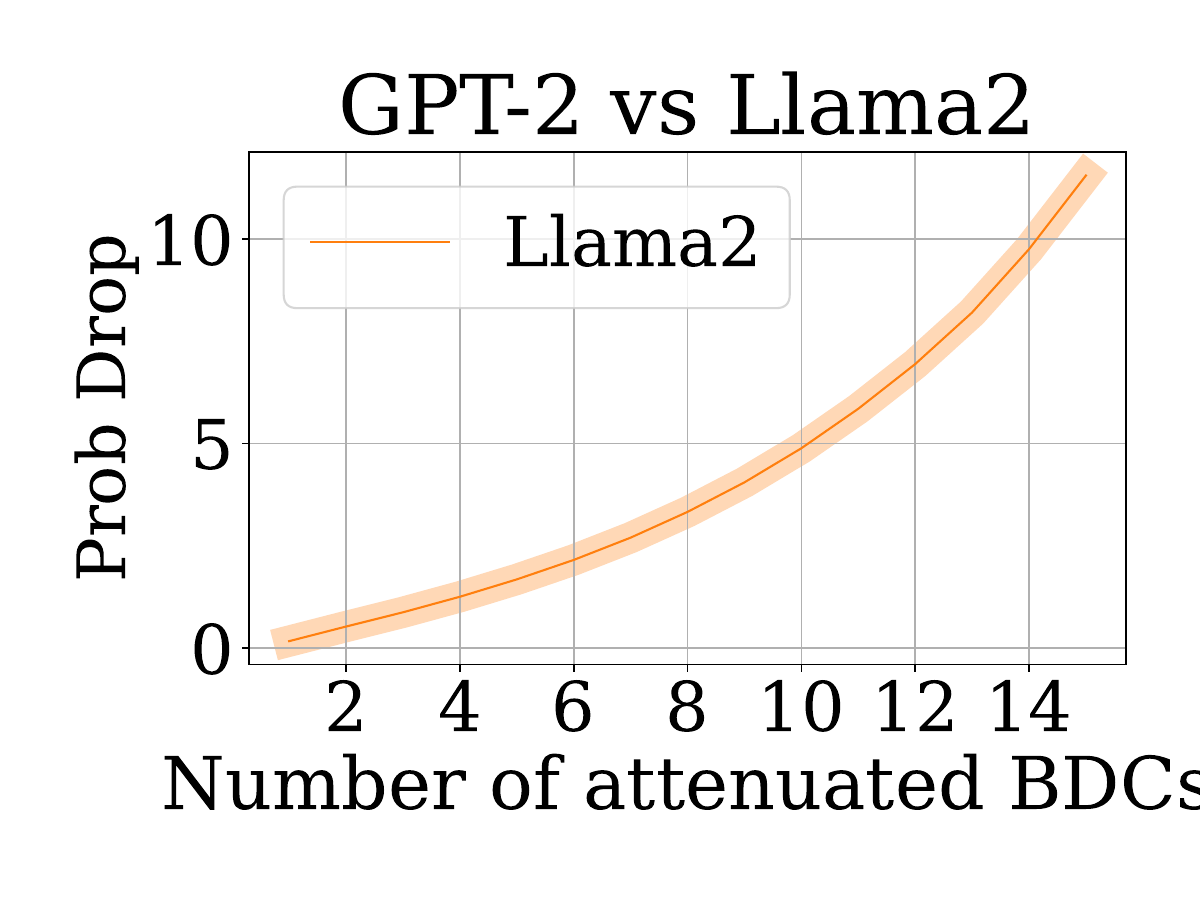}
\end{subfigure}

\begin{subfigure}{.5\linewidth}
\centering
\includegraphics[width=\linewidth]{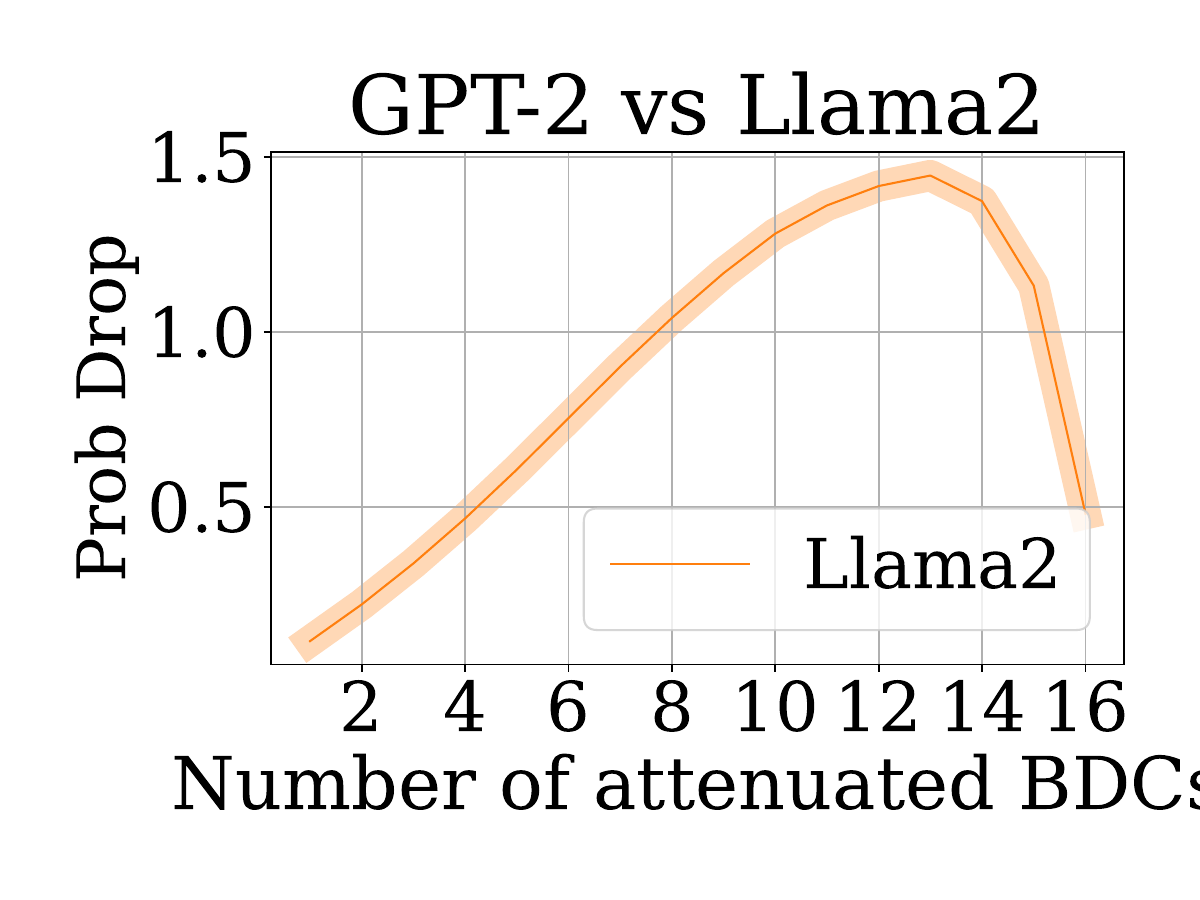}
\end{subfigure}%
\begin{subfigure}{.5\linewidth}
\centering
\includegraphics[width=\linewidth]{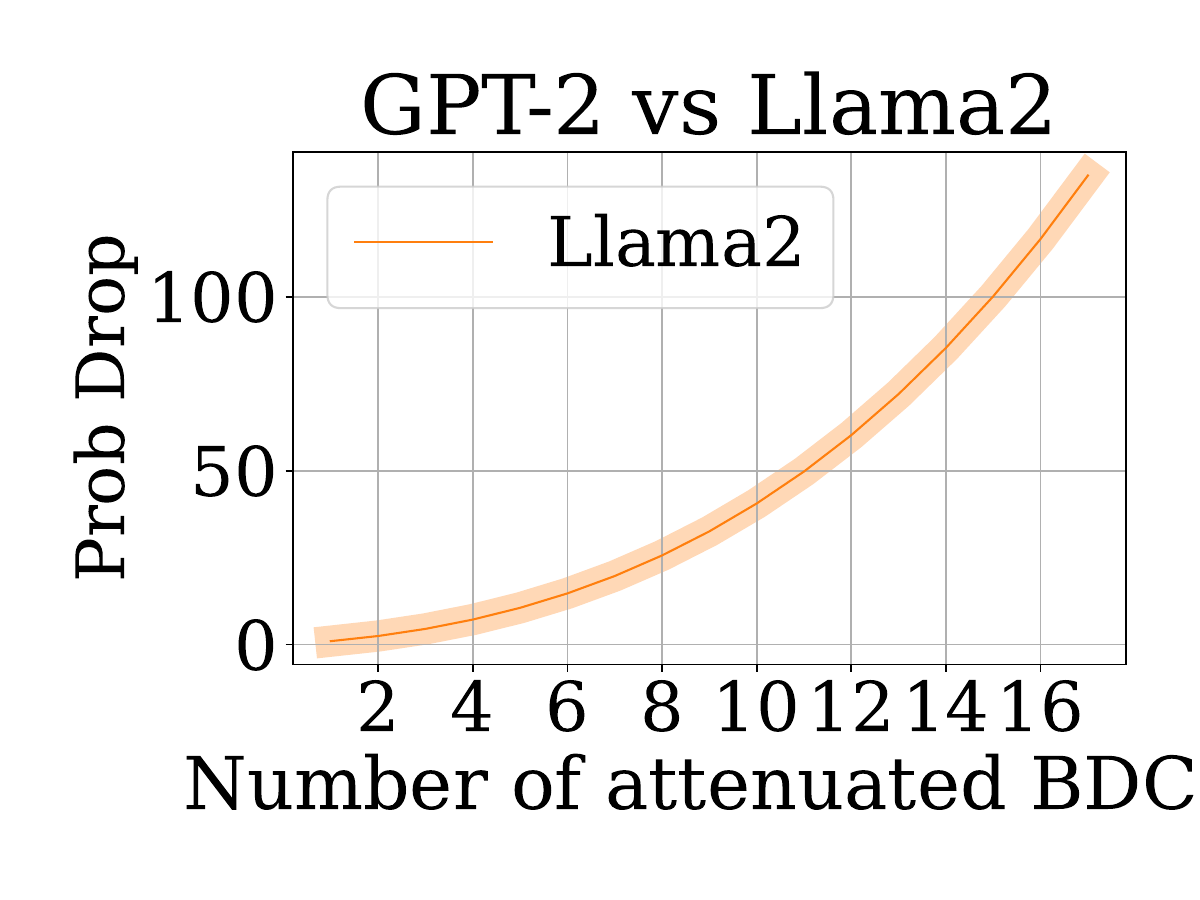}
\end{subfigure}
\caption{The graph of $\Delta Prob$ relative to the number of suppressed BDCs obtained using the NTC method. }
\label{fig-appendix-main}
\end{figure}

\begin{figure}[h]
\centering
\begin{subfigure}{.5\linewidth}
\centering
\includegraphics[width=\linewidth]{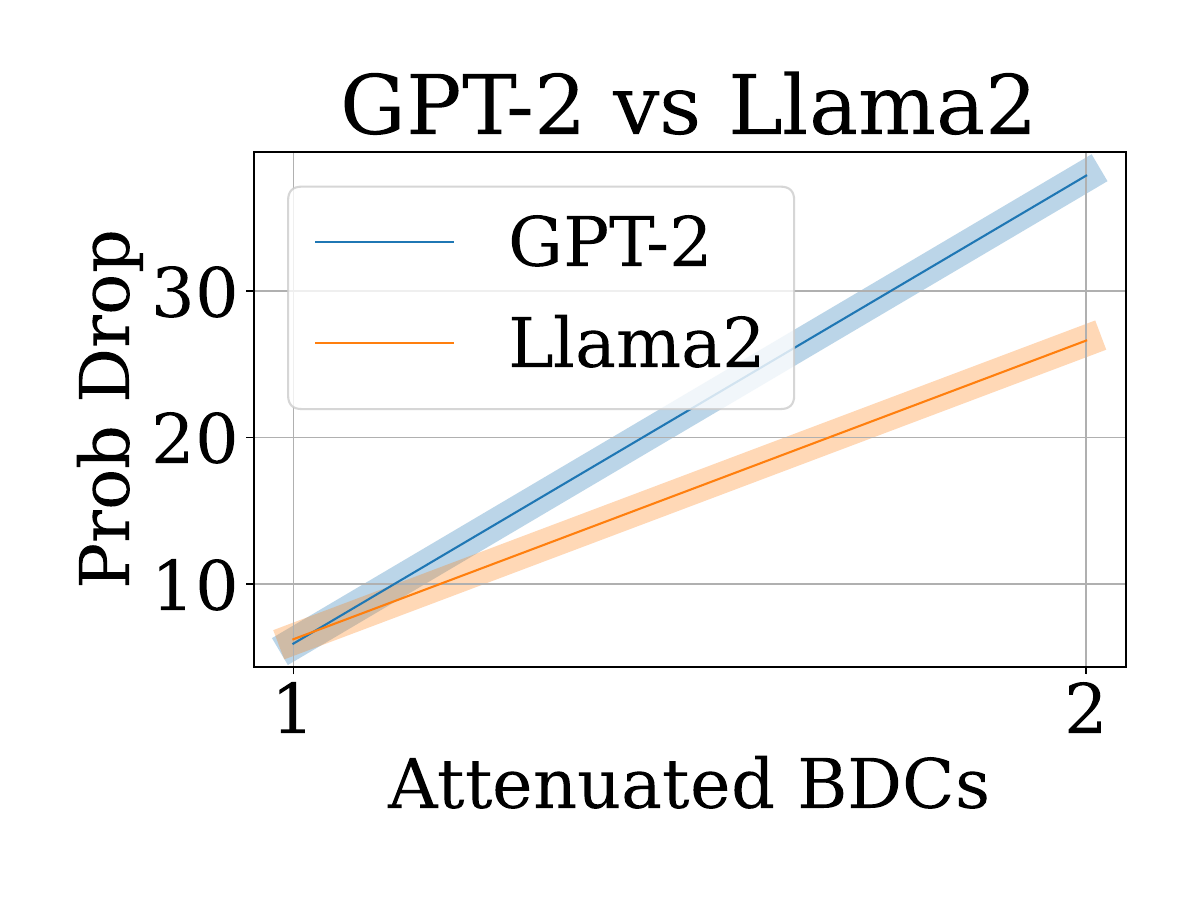}
\end{subfigure}%
\begin{subfigure}{.5\linewidth}
\centering
\includegraphics[width=\linewidth]{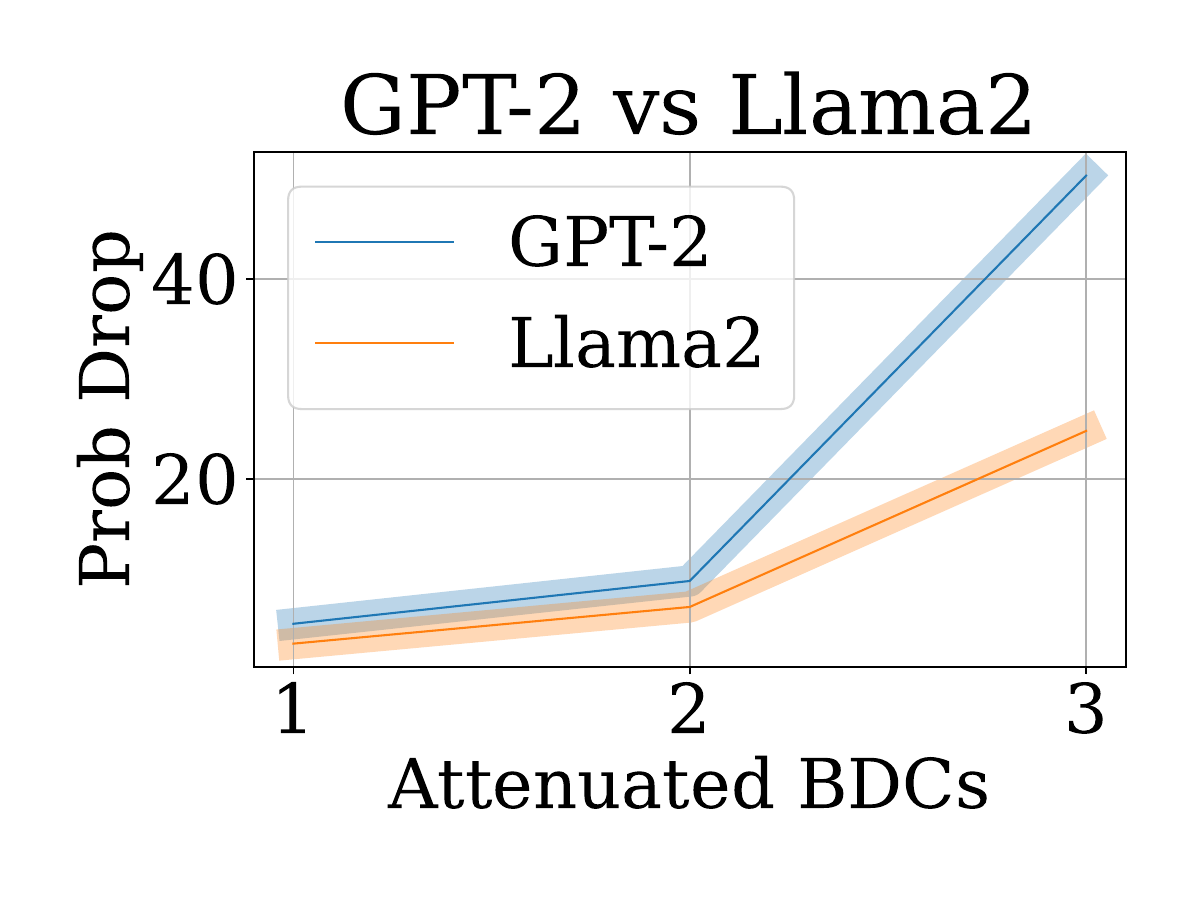}
\end{subfigure}

\begin{subfigure}{.5\linewidth}
\centering
\includegraphics[width=\linewidth]{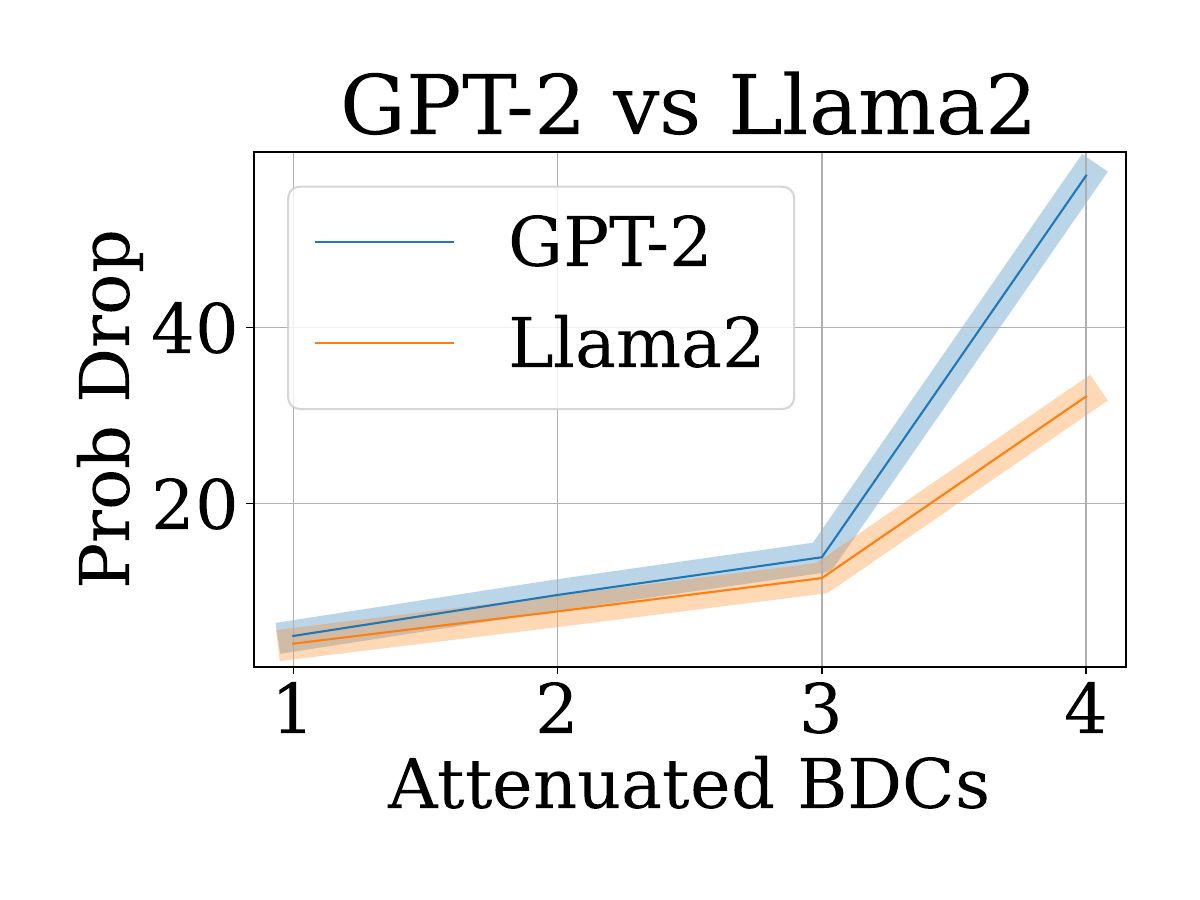}
\end{subfigure}%
\begin{subfigure}{.5\linewidth}
\centering
\includegraphics[width=\linewidth]{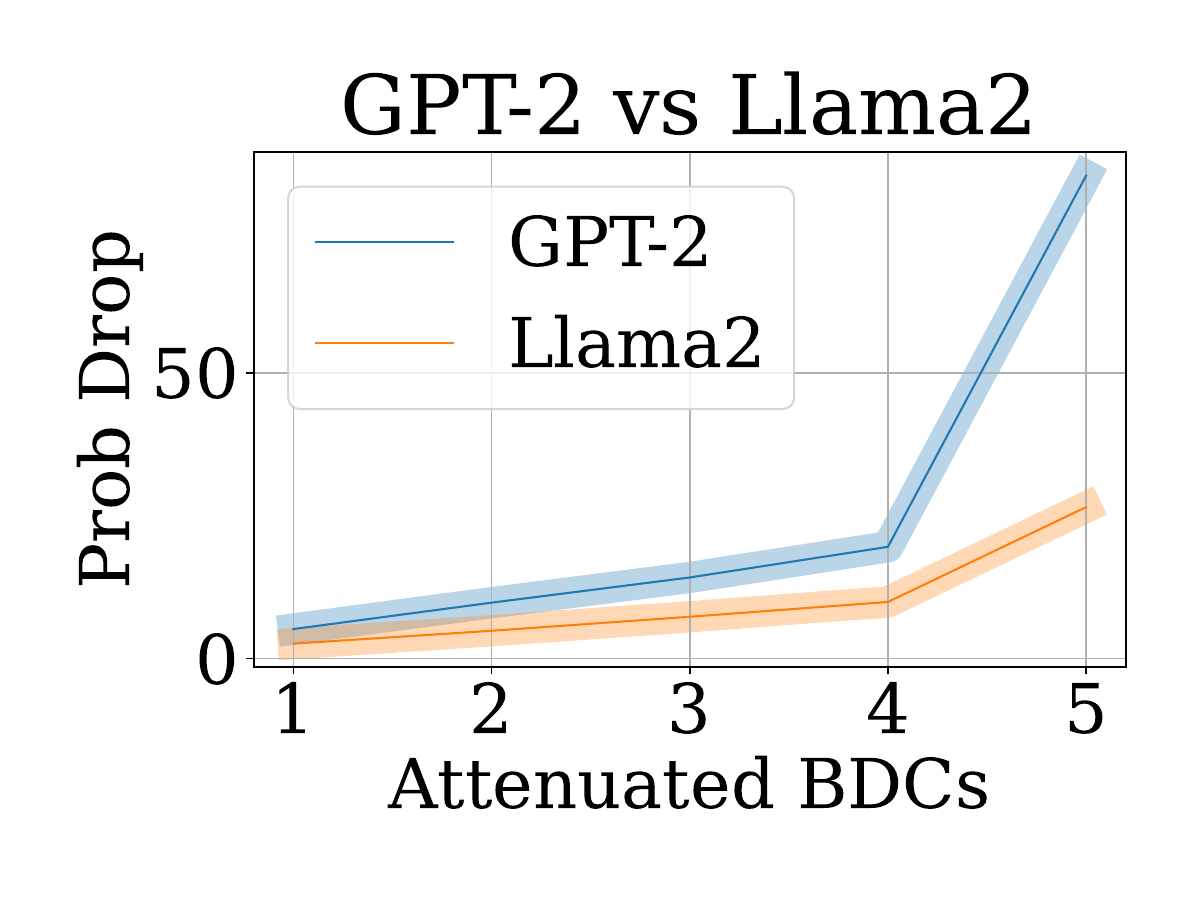}
\end{subfigure}

\begin{subfigure}{.5\linewidth}
\centering
\includegraphics[width=\linewidth]{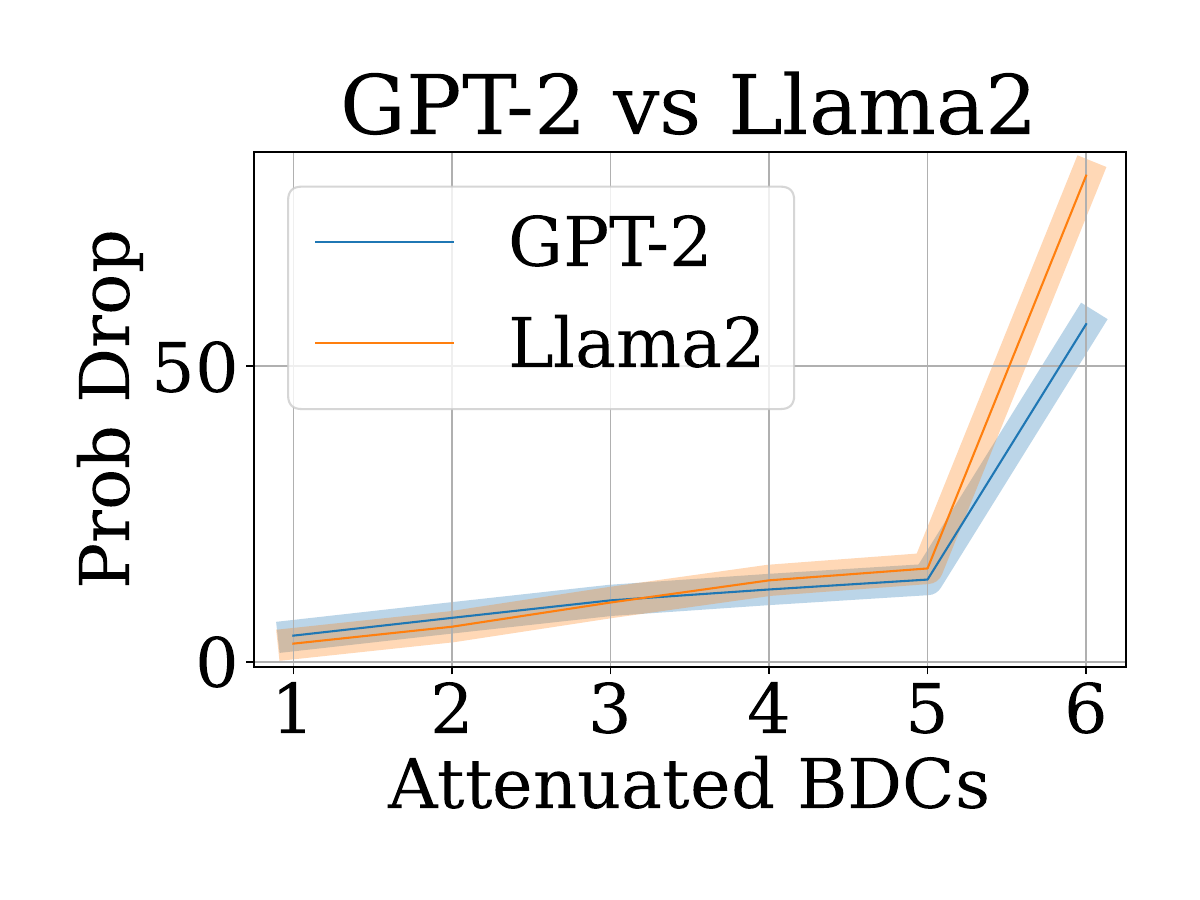}
\end{subfigure}%
\begin{subfigure}{.5\linewidth}
\centering
\includegraphics[width=\linewidth]{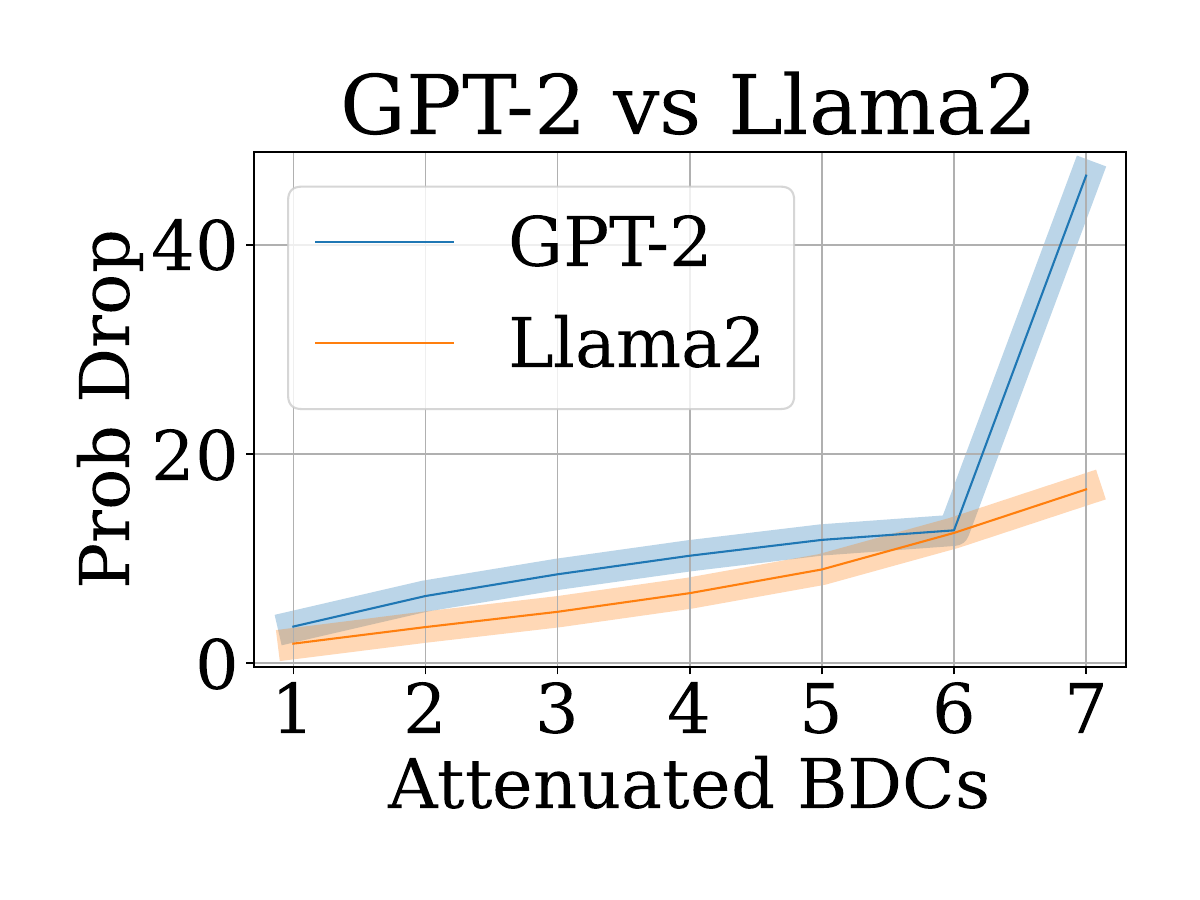}
\end{subfigure}

\begin{subfigure}{.5\linewidth}
\centering
\includegraphics[width=\linewidth]{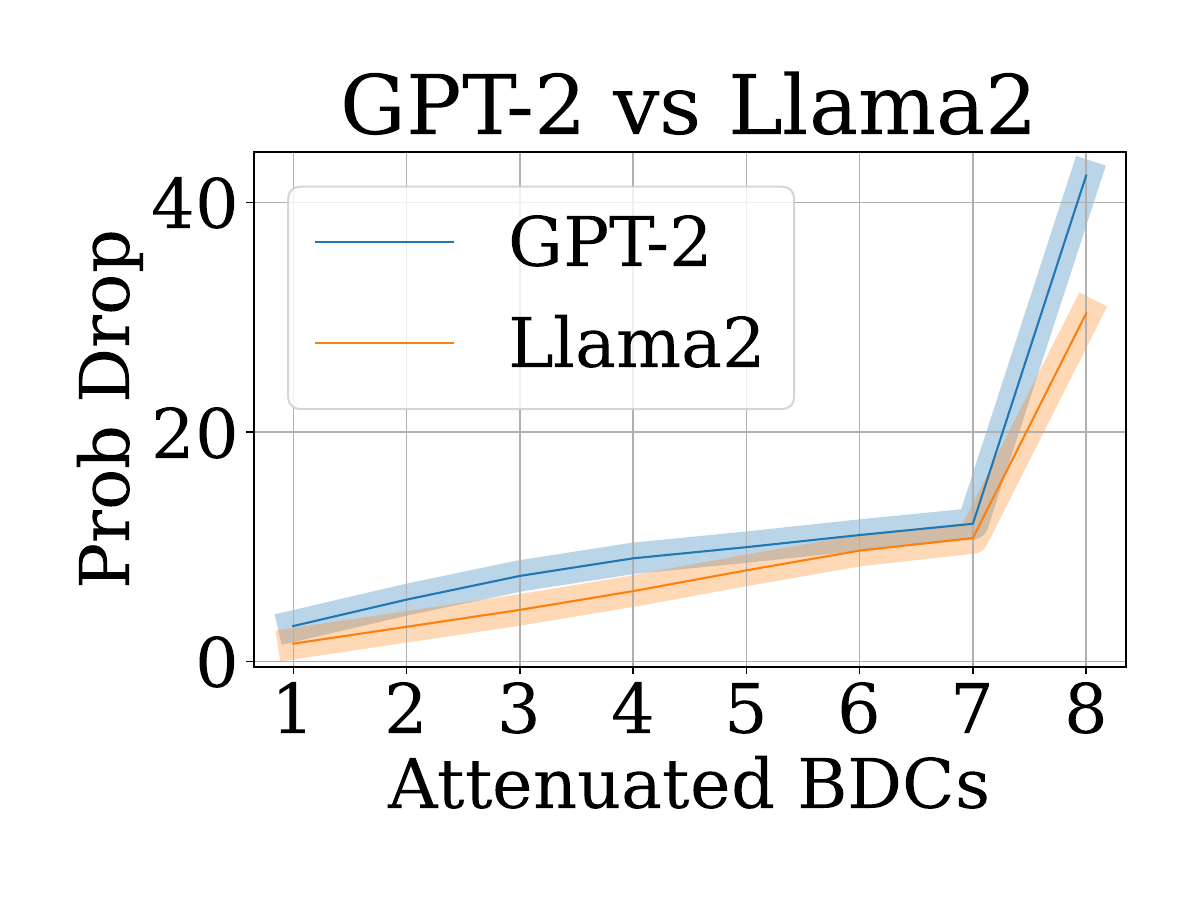}
\end{subfigure}%
\begin{subfigure}{.5\linewidth}
\centering
\includegraphics[width=\linewidth]{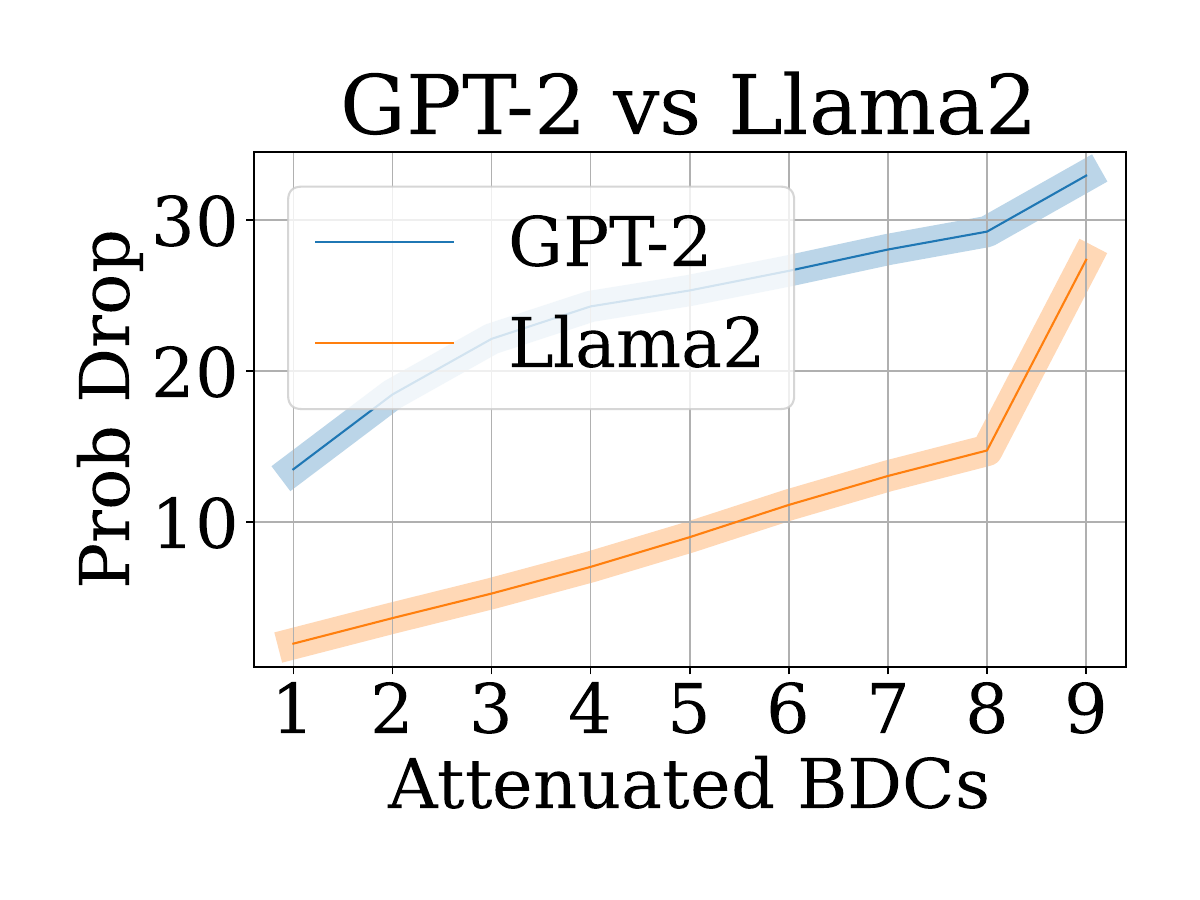}
\end{subfigure}

\begin{subfigure}{.5\linewidth}
\centering
\includegraphics[width=\linewidth]{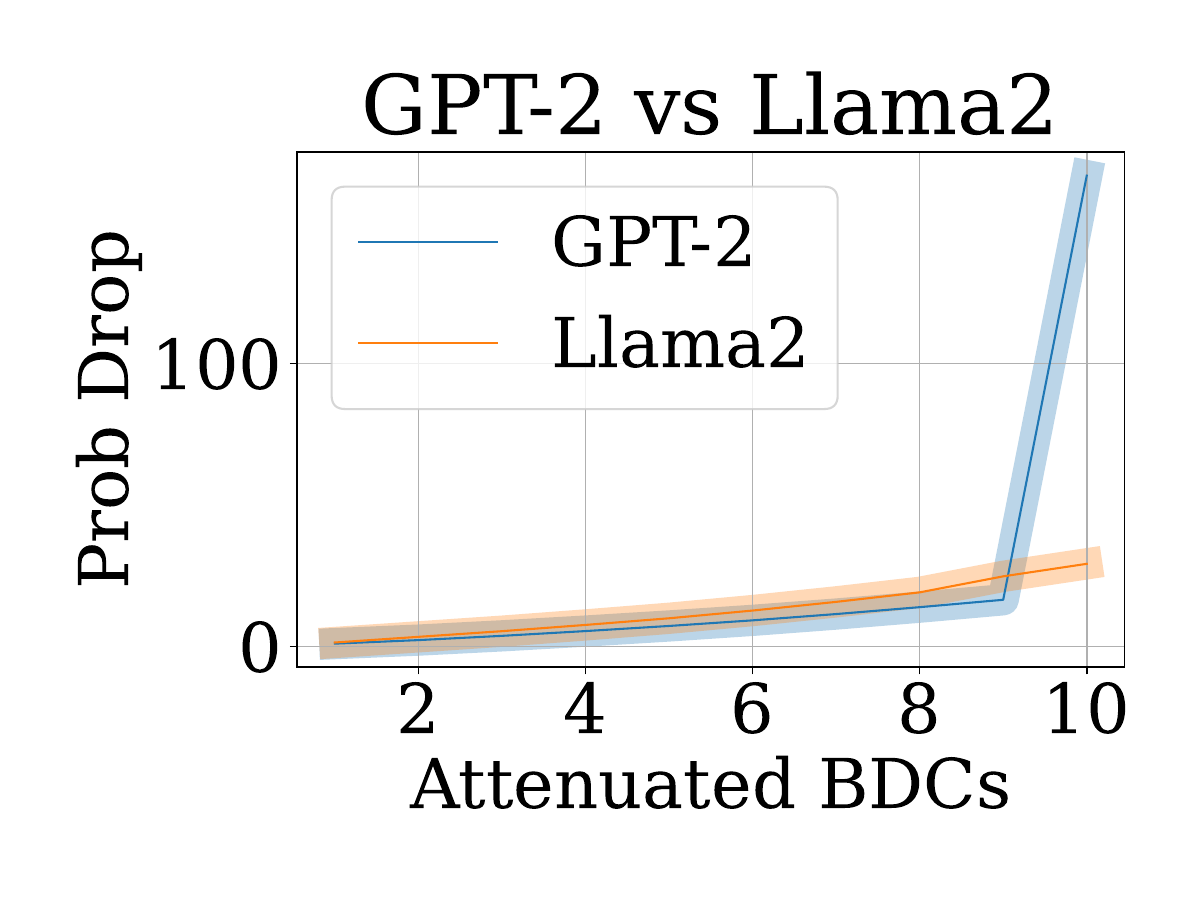}
\end{subfigure}%
\begin{subfigure}{.5\linewidth}
\centering
\includegraphics[width=\linewidth]{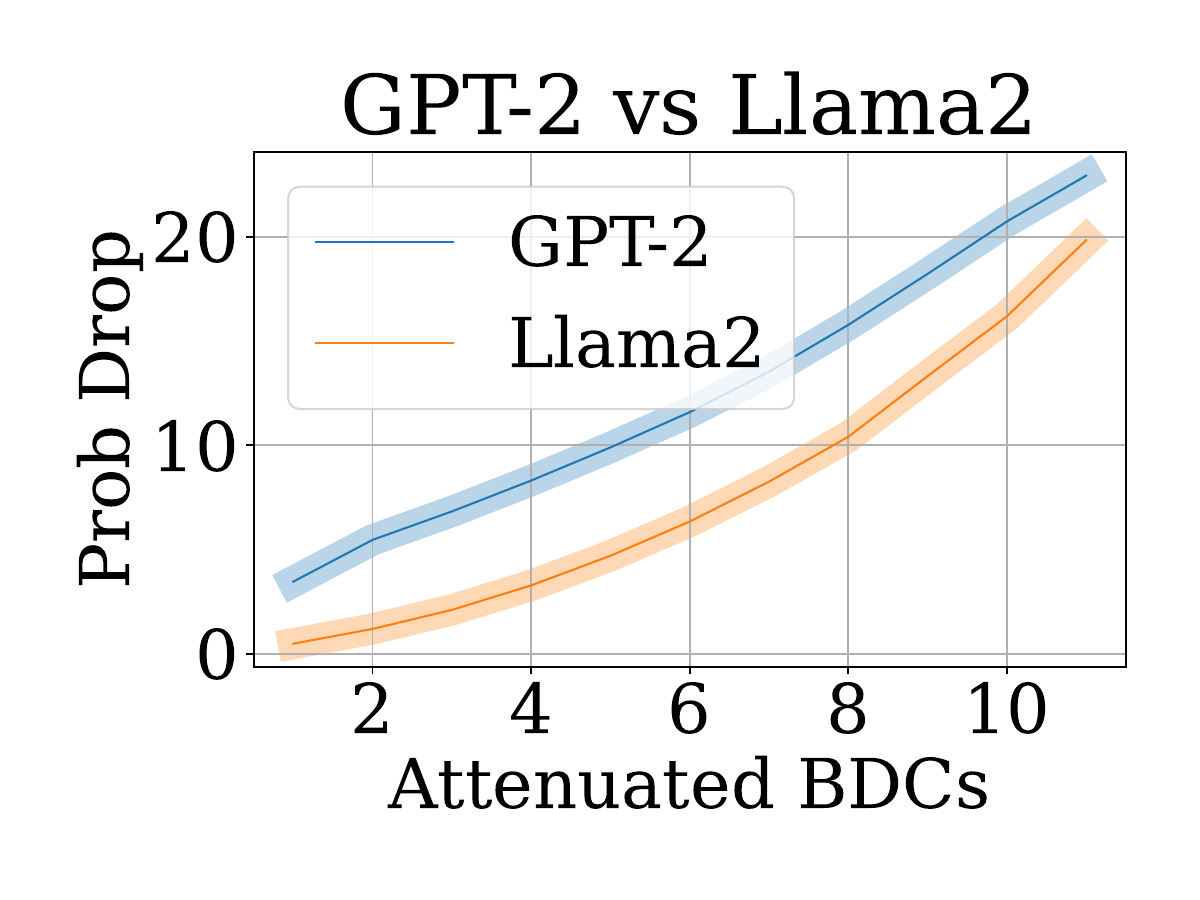}
\end{subfigure}

\begin{subfigure}{.5\linewidth}
\centering
\includegraphics[width=\linewidth]{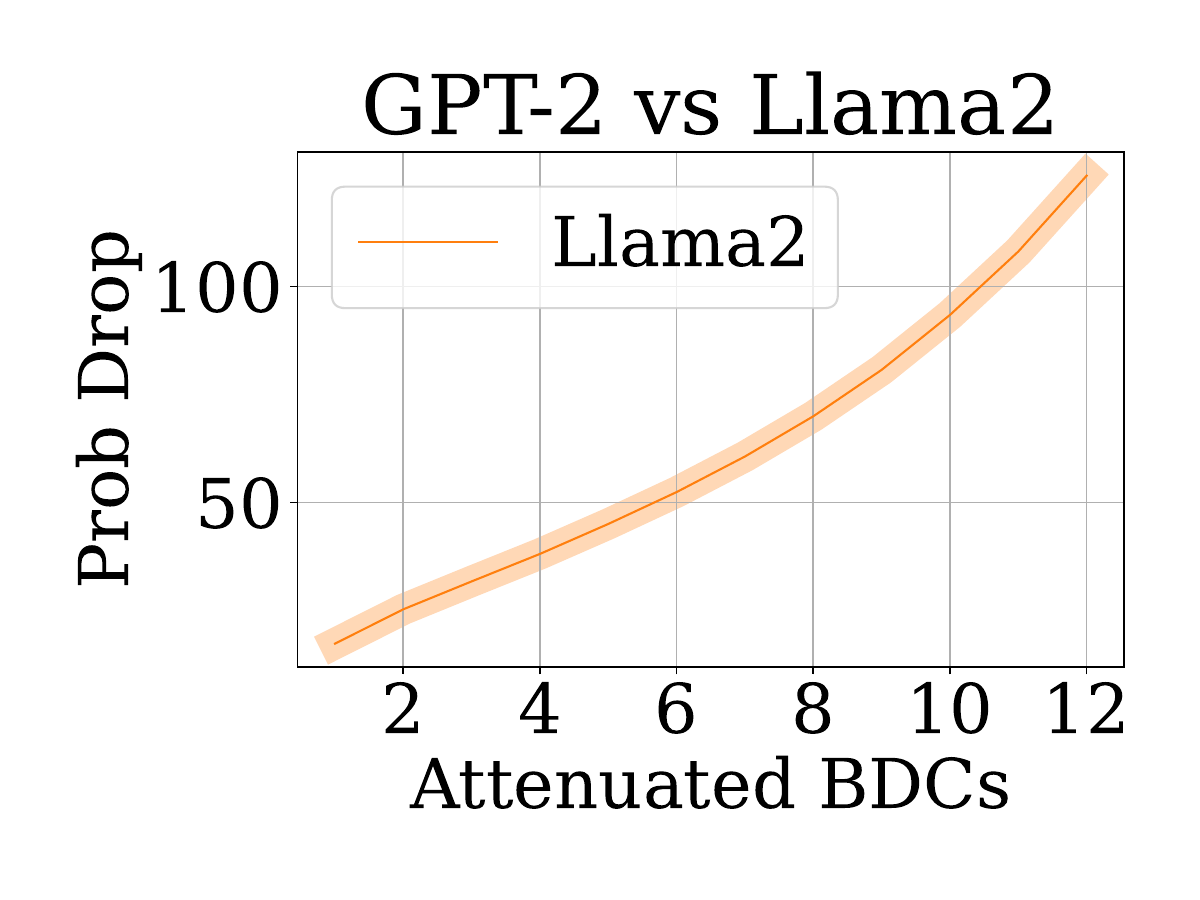}
\end{subfigure}%
\begin{subfigure}{.5\linewidth}
\centering
\includegraphics[width=\linewidth]{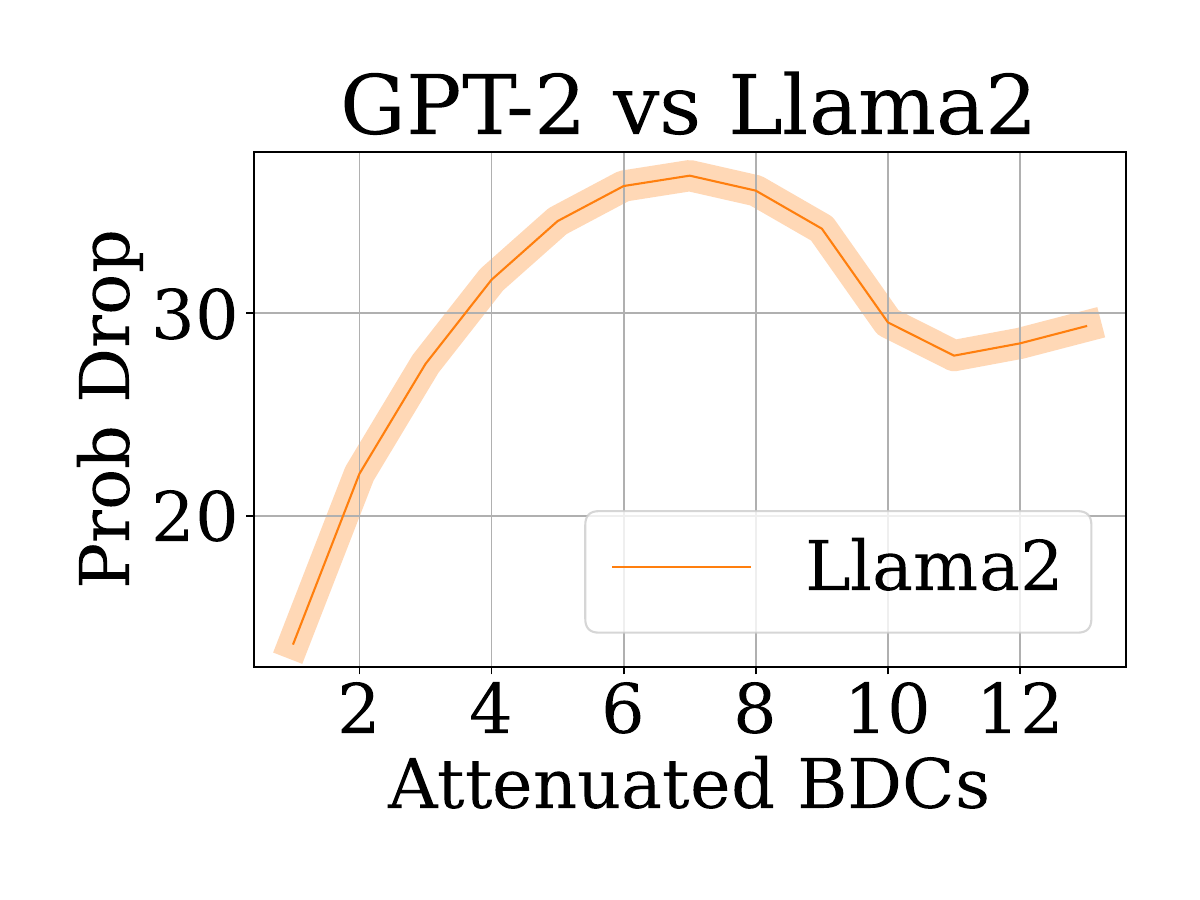}
\end{subfigure}

\begin{subfigure}{.5\linewidth}
\centering
\includegraphics[width=\linewidth]{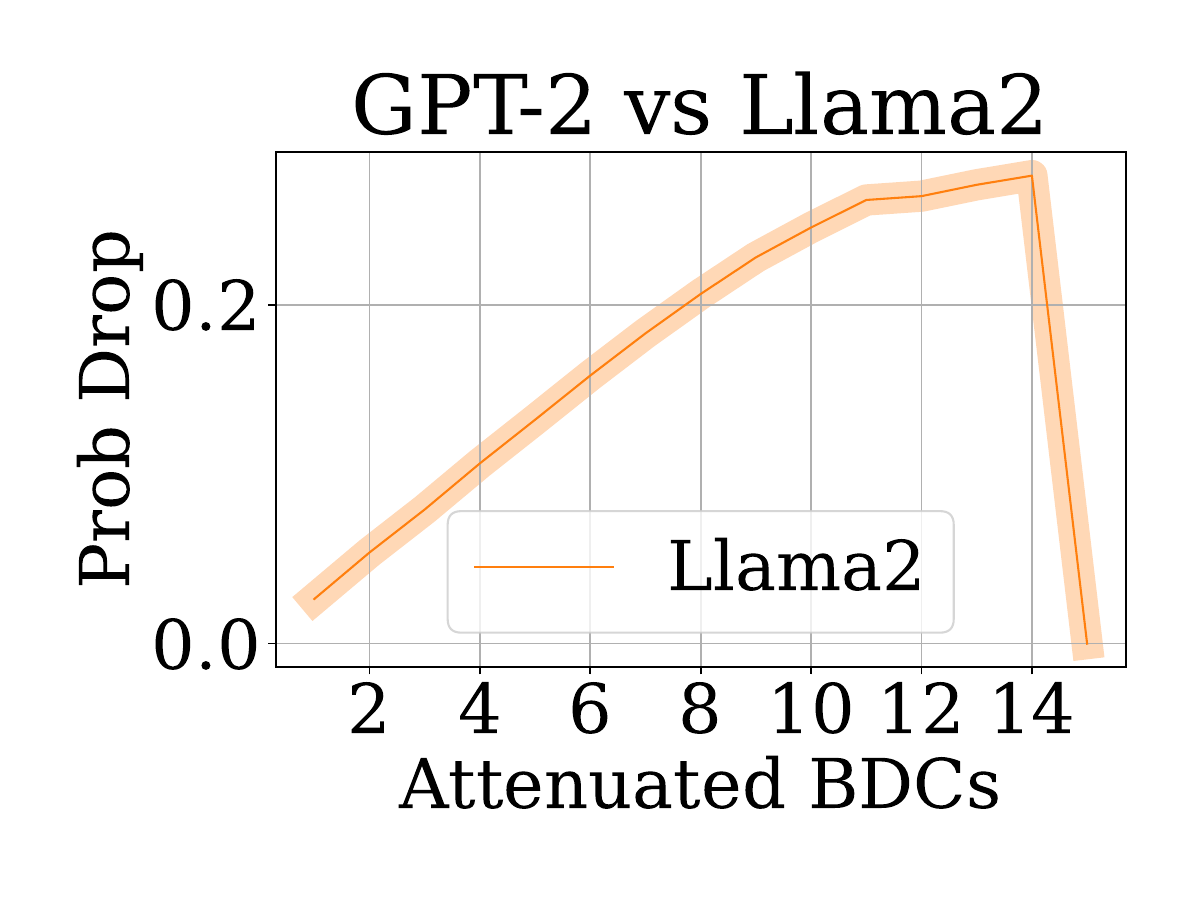}
\end{subfigure}
\caption{The graph of $\Delta Prob$ relative to the number of suppressed BDCs obtained using the DBSCAN method. }
\label{fig-appendix-DBSCAN}
\end{figure}

\begin{figure}[h]
\centering
\begin{subfigure}{.5\linewidth}
\centering
\includegraphics[width=\linewidth]{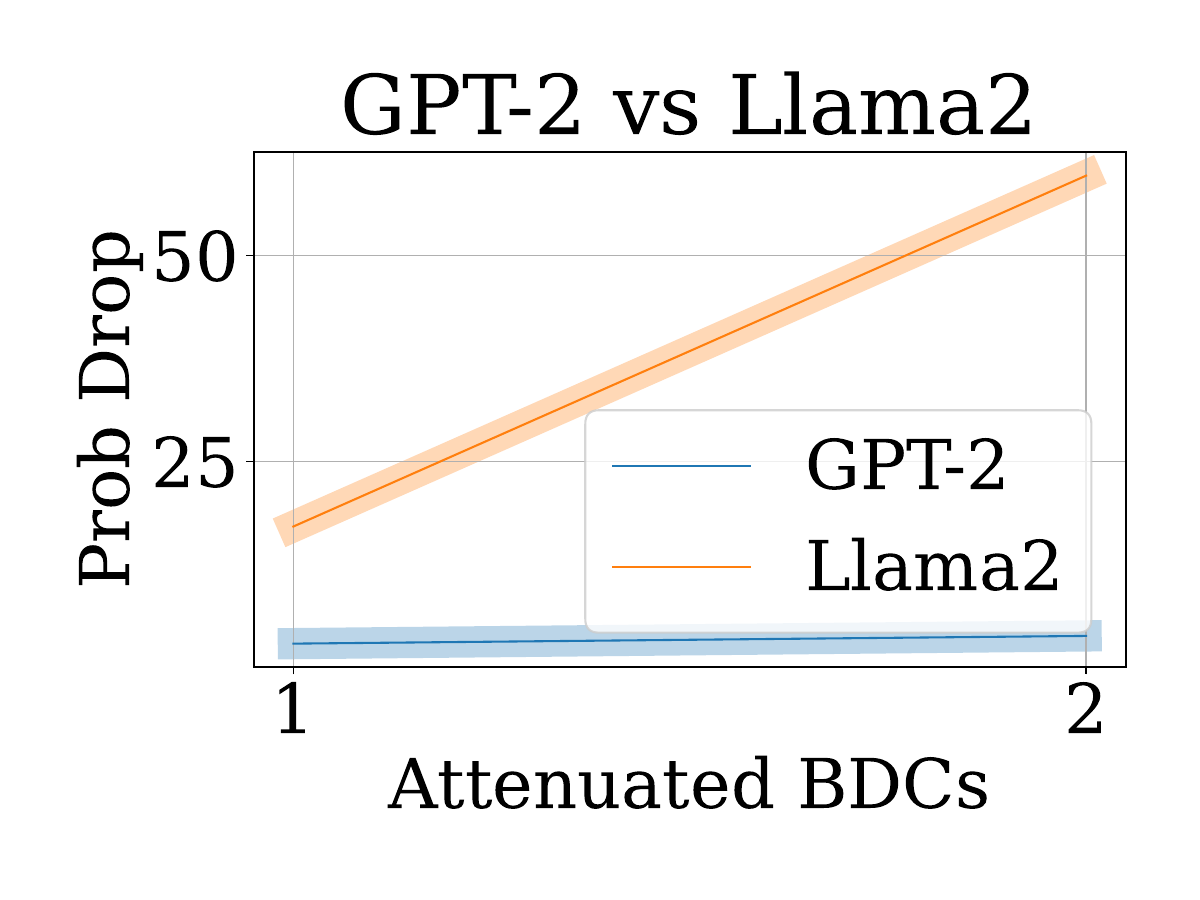}
\end{subfigure}%
\begin{subfigure}{.5\linewidth}
\centering
\includegraphics[width=\linewidth]{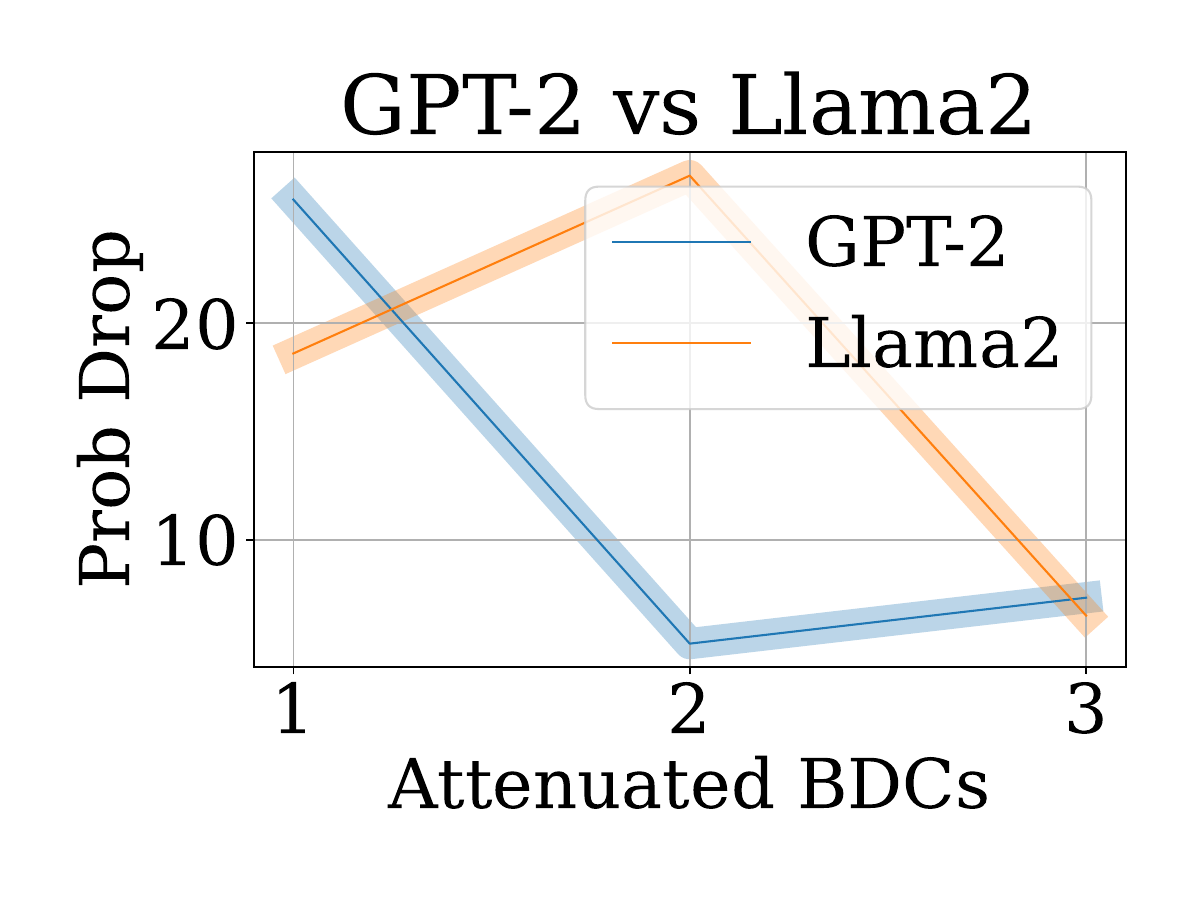}
\end{subfigure}

\begin{subfigure}{.5\linewidth}
\centering
\includegraphics[width=\linewidth]{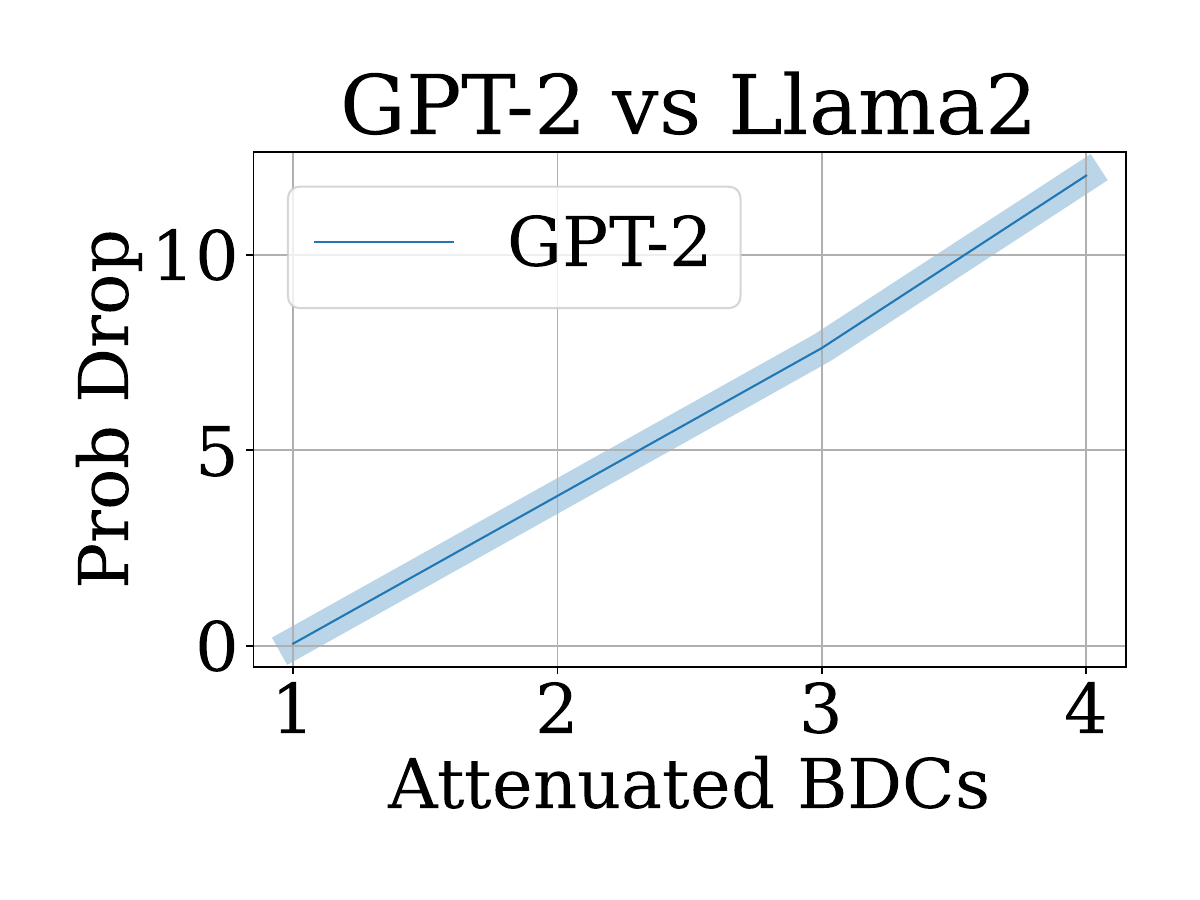}
\end{subfigure}%
\begin{subfigure}{.5\linewidth}
\centering
\includegraphics[width=\linewidth]{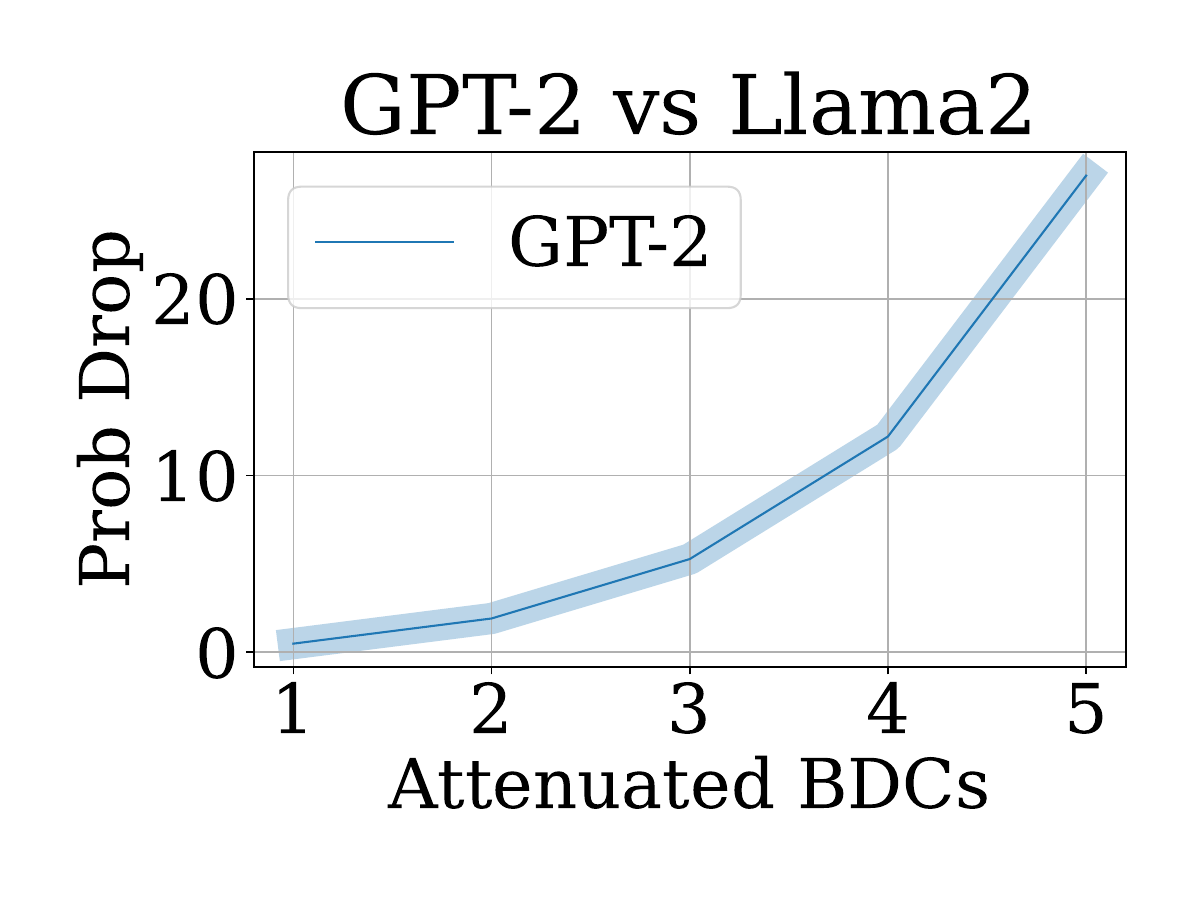}
\end{subfigure}

\begin{subfigure}{.5\linewidth}
\centering
\includegraphics[width=\linewidth]{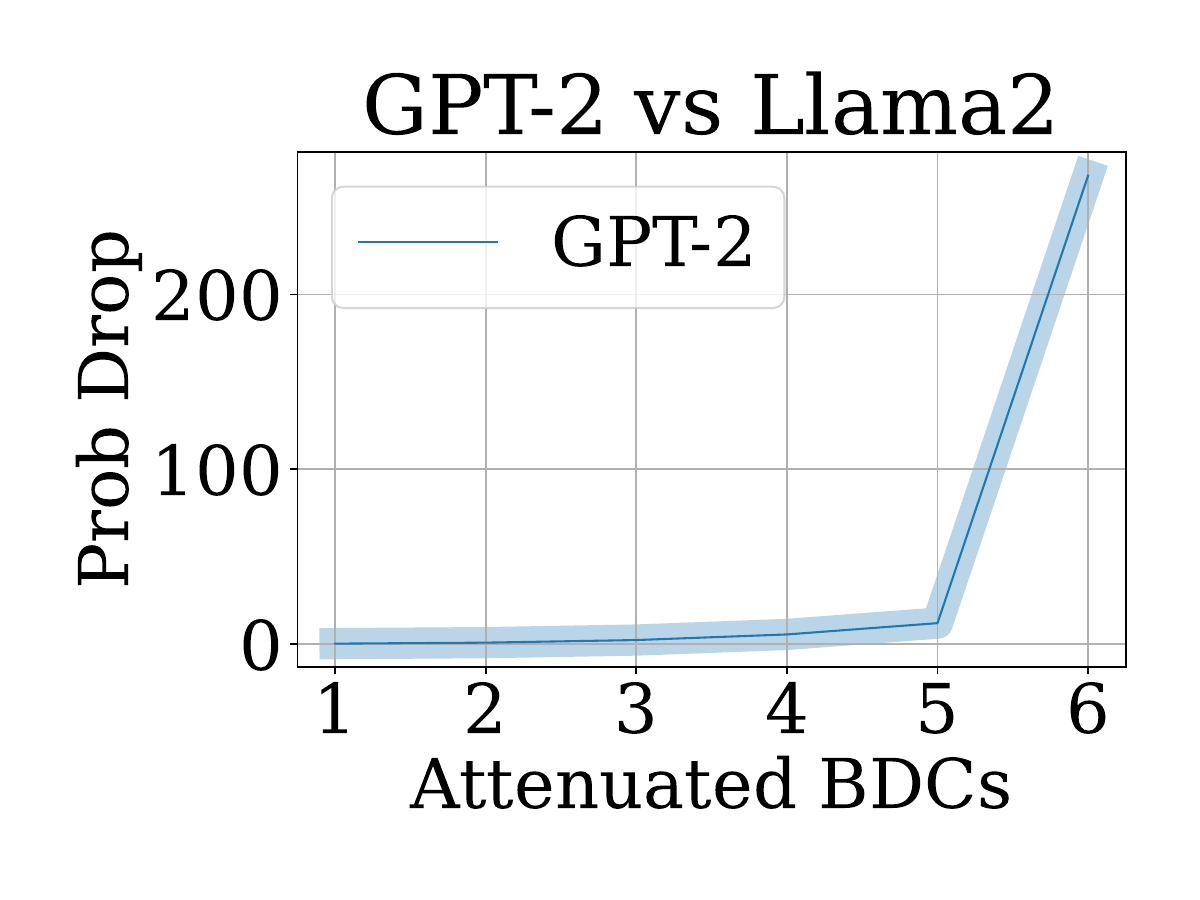}
\end{subfigure}%
\begin{subfigure}{.5\linewidth}
\centering
\includegraphics[width=\linewidth]{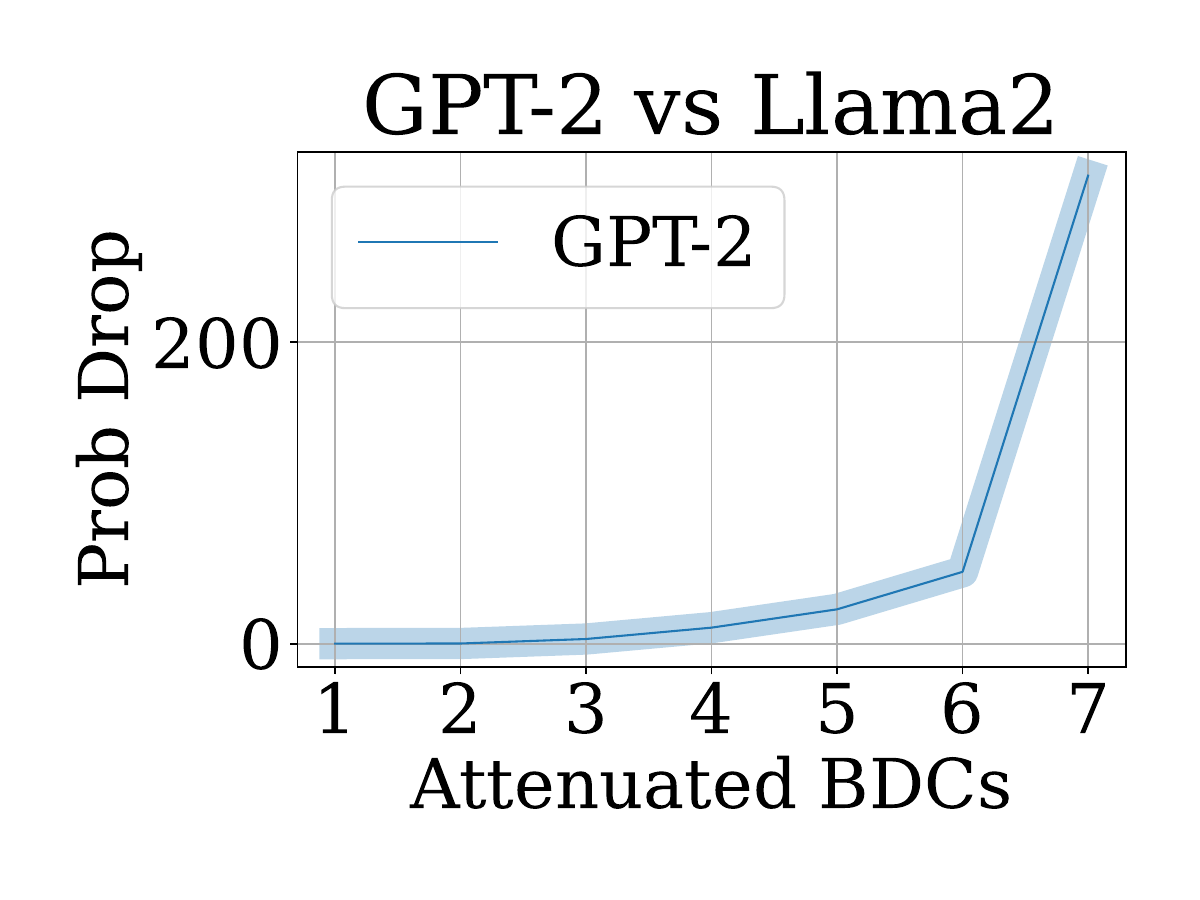}
\end{subfigure}

\begin{subfigure}{.5\linewidth}
\centering
\includegraphics[width=\linewidth]{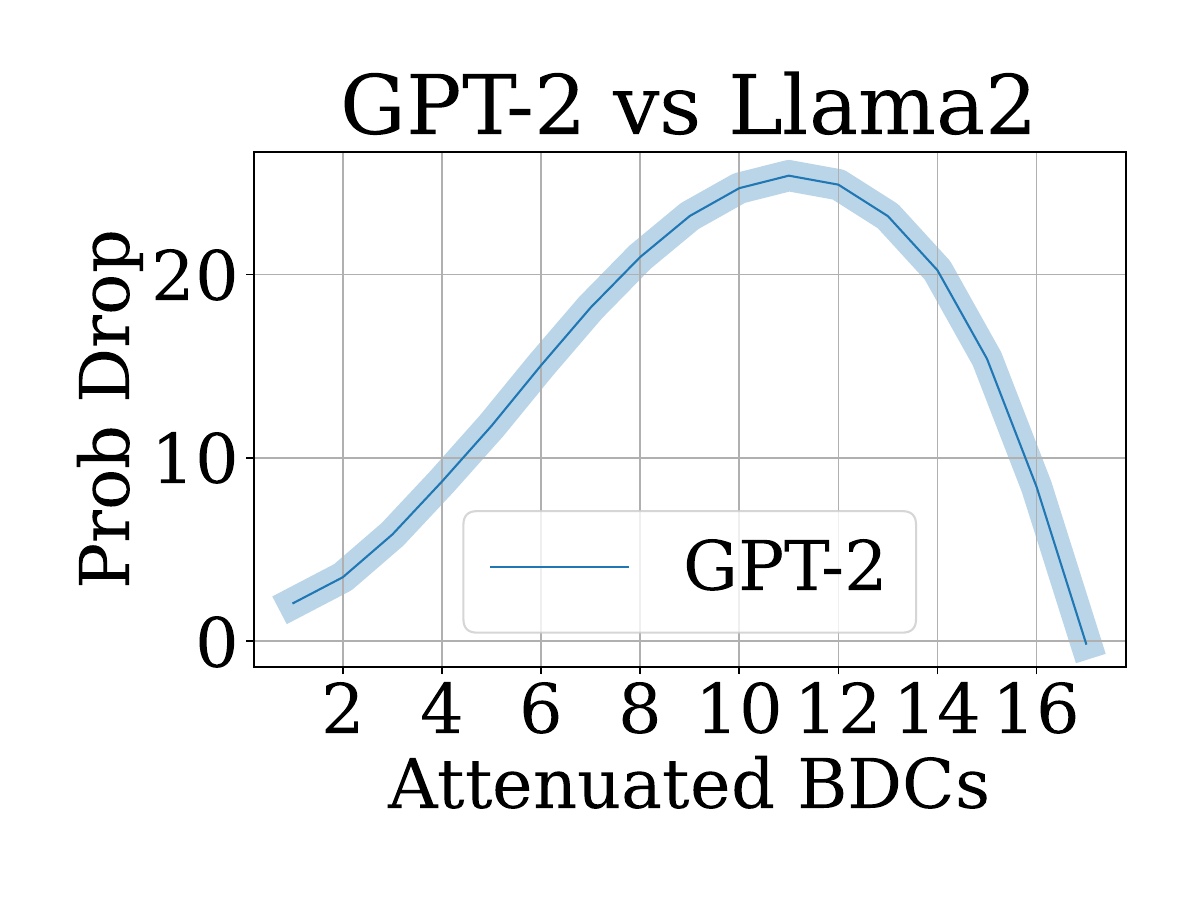}
\end{subfigure}
\caption{The graph of $\Delta Prob$ relative to the number of suppressed BDCs obtained using the Hierarchical method. }
\label{fig-appendix-Hierarchical}
\end{figure}

\begin{figure}[h]
\centering
\begin{subfigure}{.5\linewidth}
\centering
\includegraphics[width=\linewidth
]{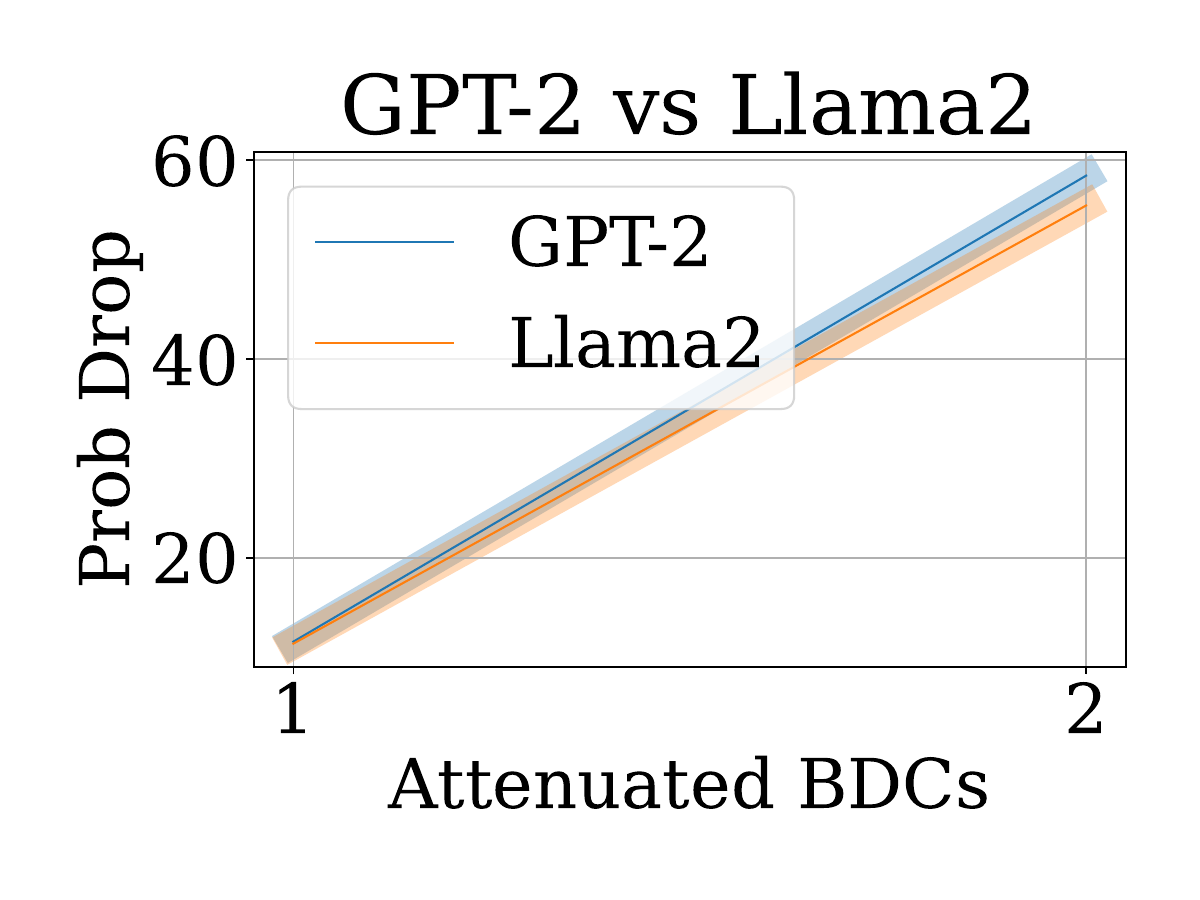}
\end{subfigure}%
\begin{subfigure}{.5\linewidth}
\centering
\includegraphics[width=\linewidth]{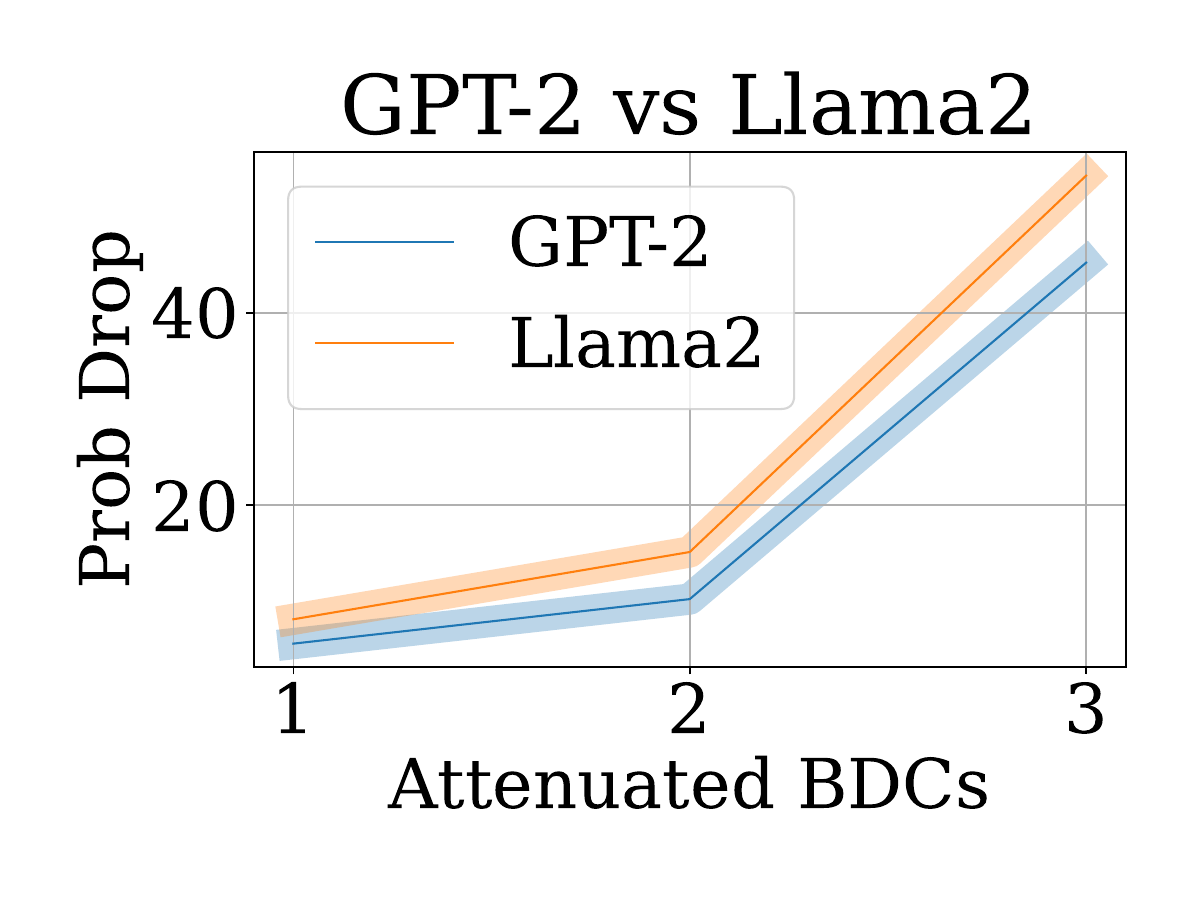}
\end{subfigure}

\begin{subfigure}{.5\linewidth}
\centering
\includegraphics[width=\linewidth]{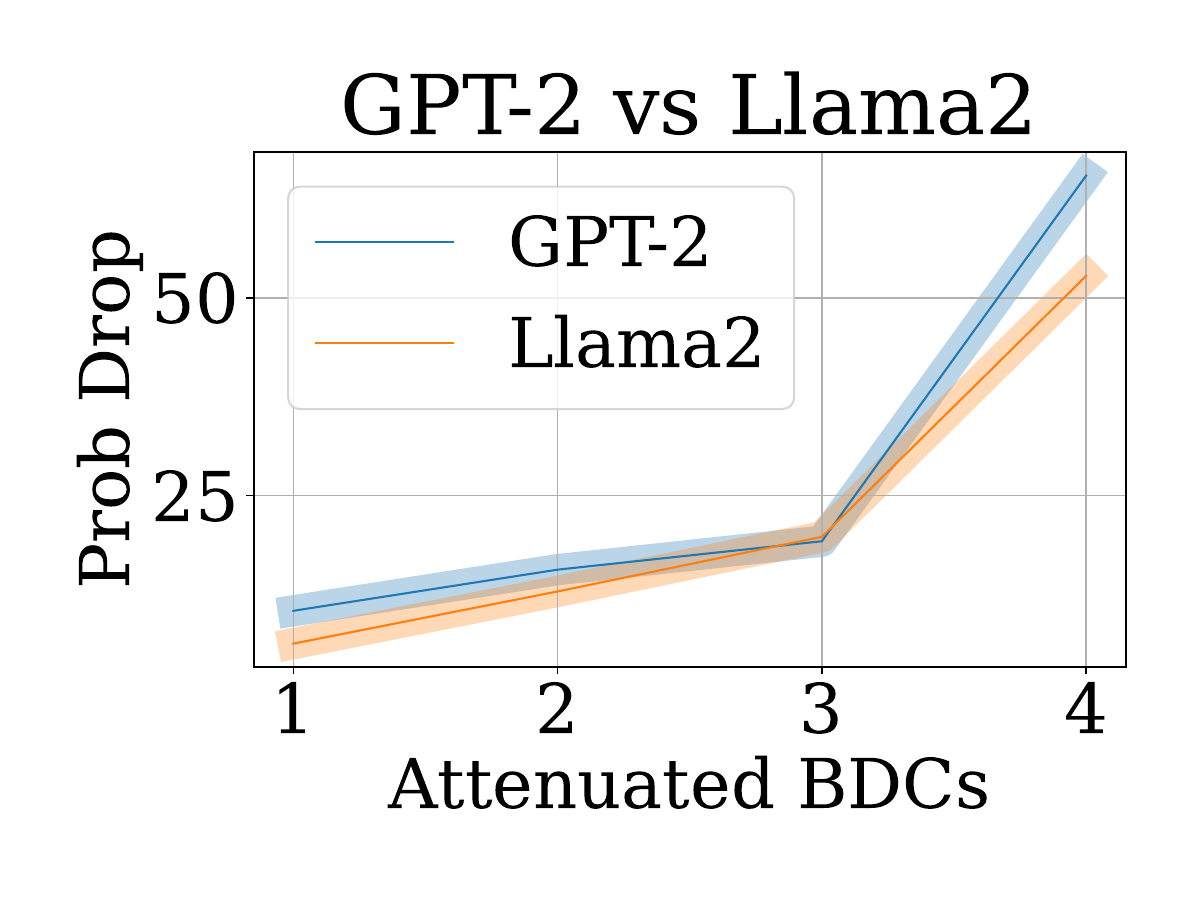}
\end{subfigure}%
\begin{subfigure}{.5\linewidth}
\centering
\includegraphics[width=\linewidth]{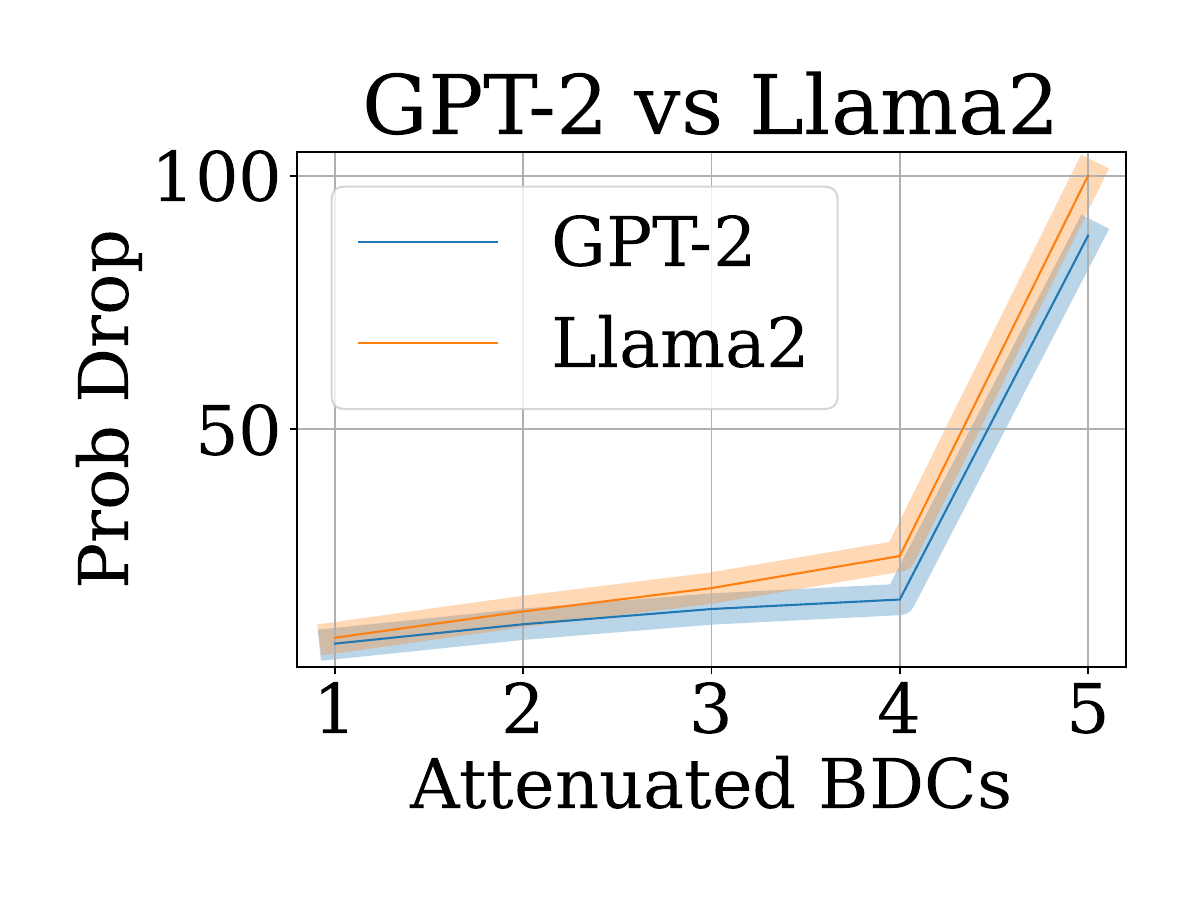}
\end{subfigure}

\begin{subfigure}{.5\linewidth}
\centering
\includegraphics[width=\linewidth]{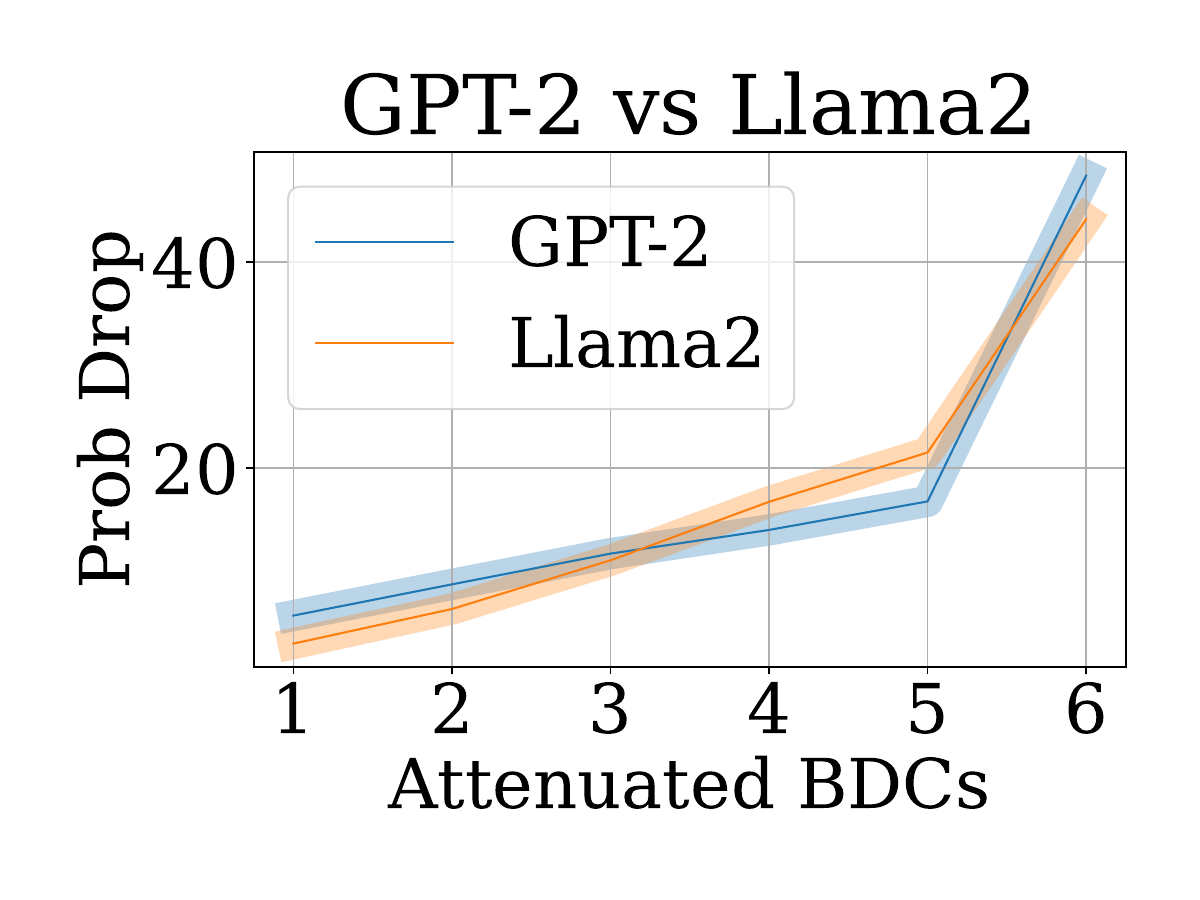}
\end{subfigure}%
\begin{subfigure}{.5\linewidth}
\centering
\includegraphics[width=\linewidth]{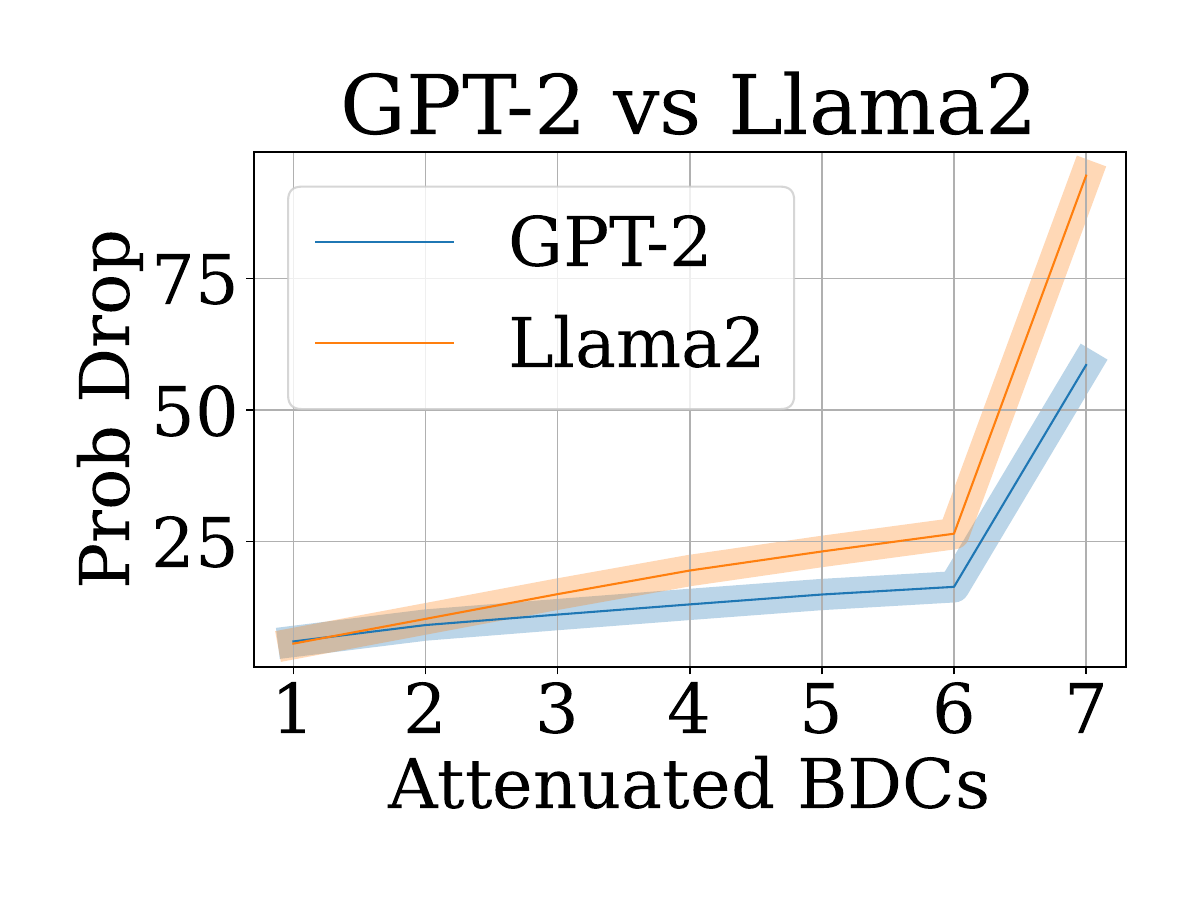}
\end{subfigure}

\begin{subfigure}{.5\linewidth}
\centering
\includegraphics[width=\linewidth]{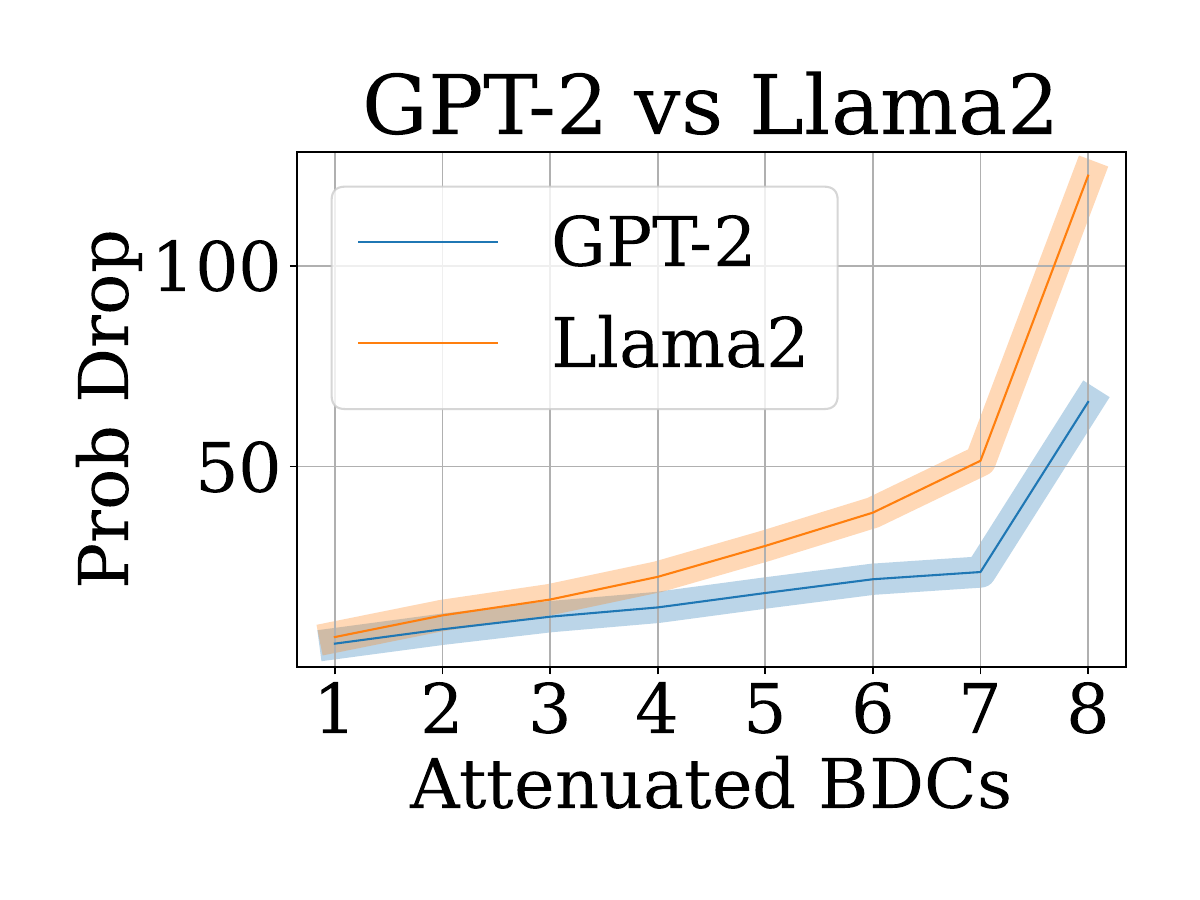}
\end{subfigure}%
\begin{subfigure}{.5\linewidth}
\centering
\includegraphics[width=\linewidth]{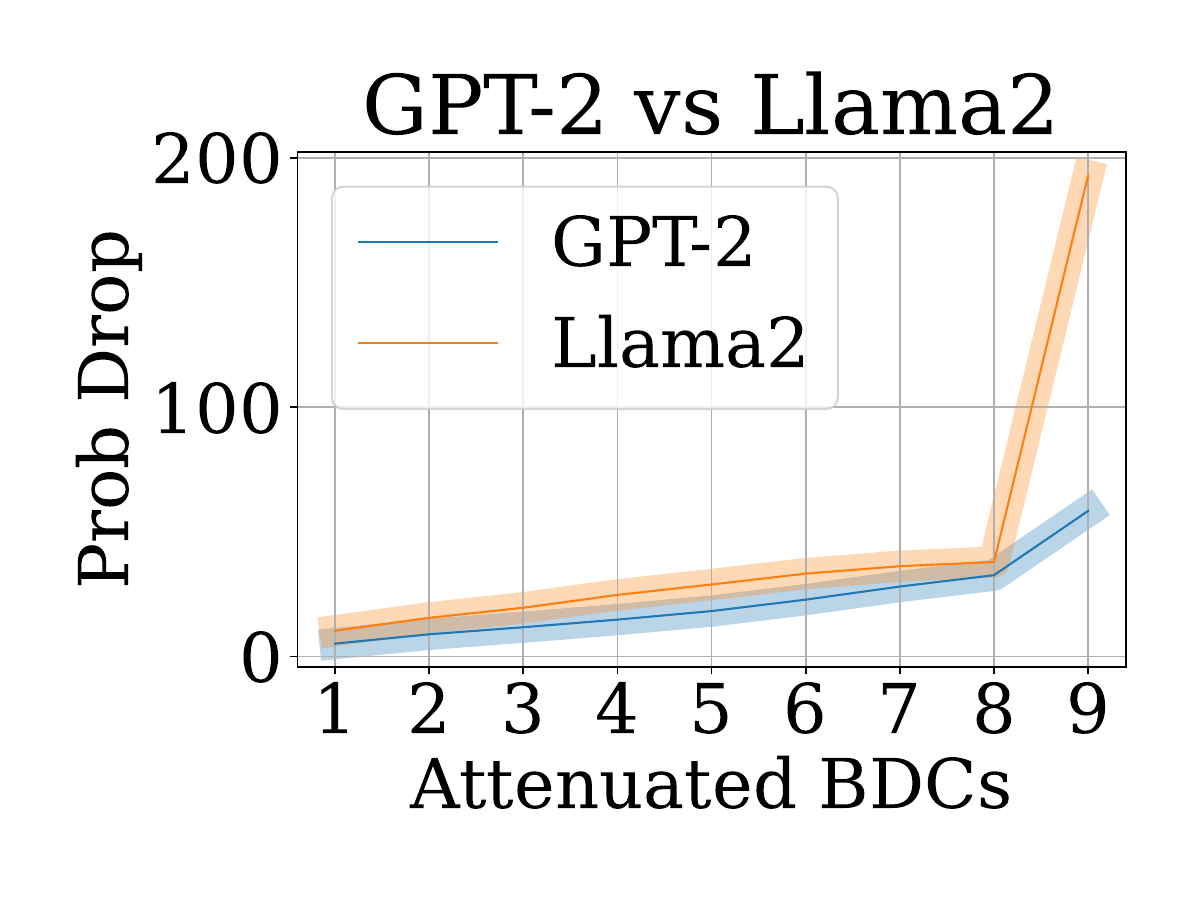}
\end{subfigure}

\begin{subfigure}{.5\linewidth}
\centering
\includegraphics[width=\linewidth]{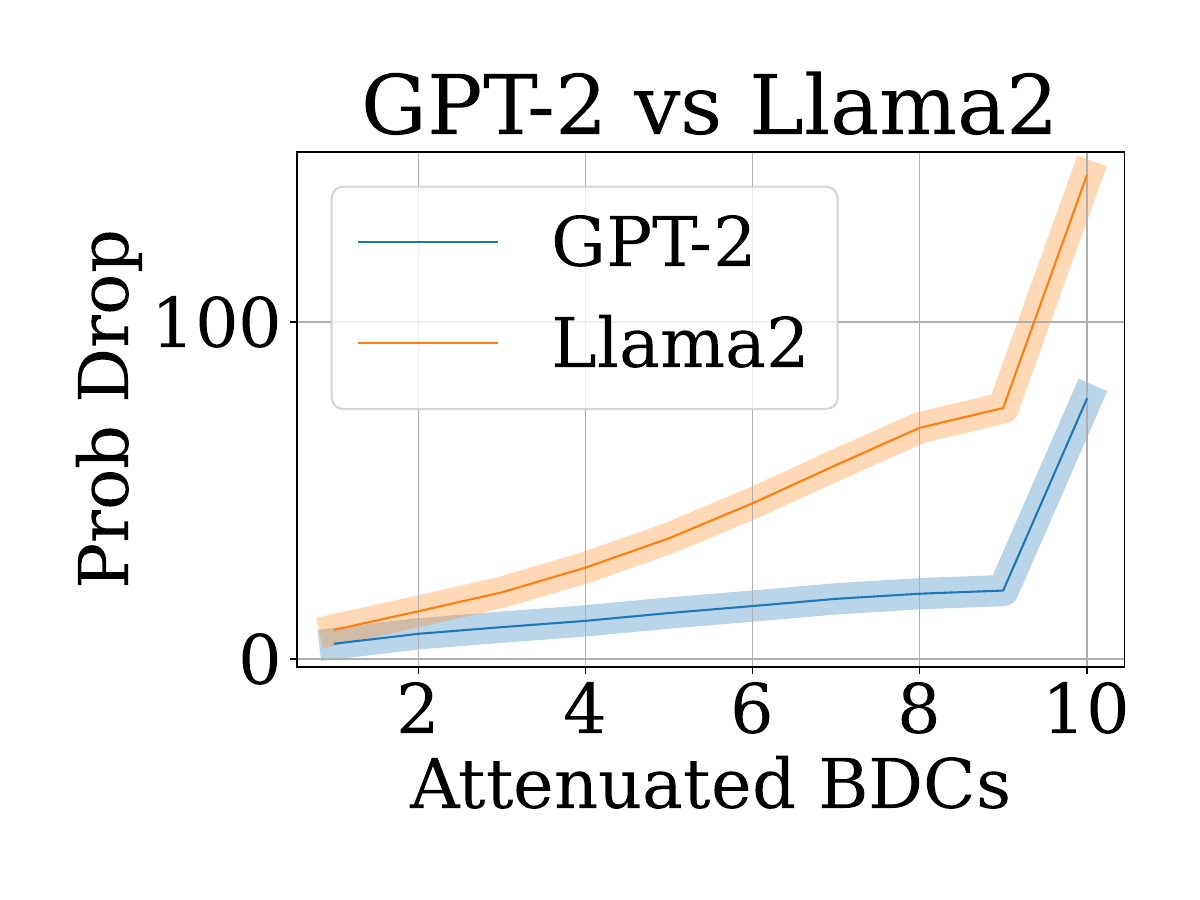}
\end{subfigure}%

\caption{The graph of $\Delta Prob$ relative to the number of suppressed BDCs obtained using the K-Means method (Part 1). }
\label{fig-appendix-K-Means-1}
\end{figure}

\begin{figure}[h]
\centering
\begin{subfigure}{.5\linewidth}
\centering
\includegraphics[width=\linewidth
]{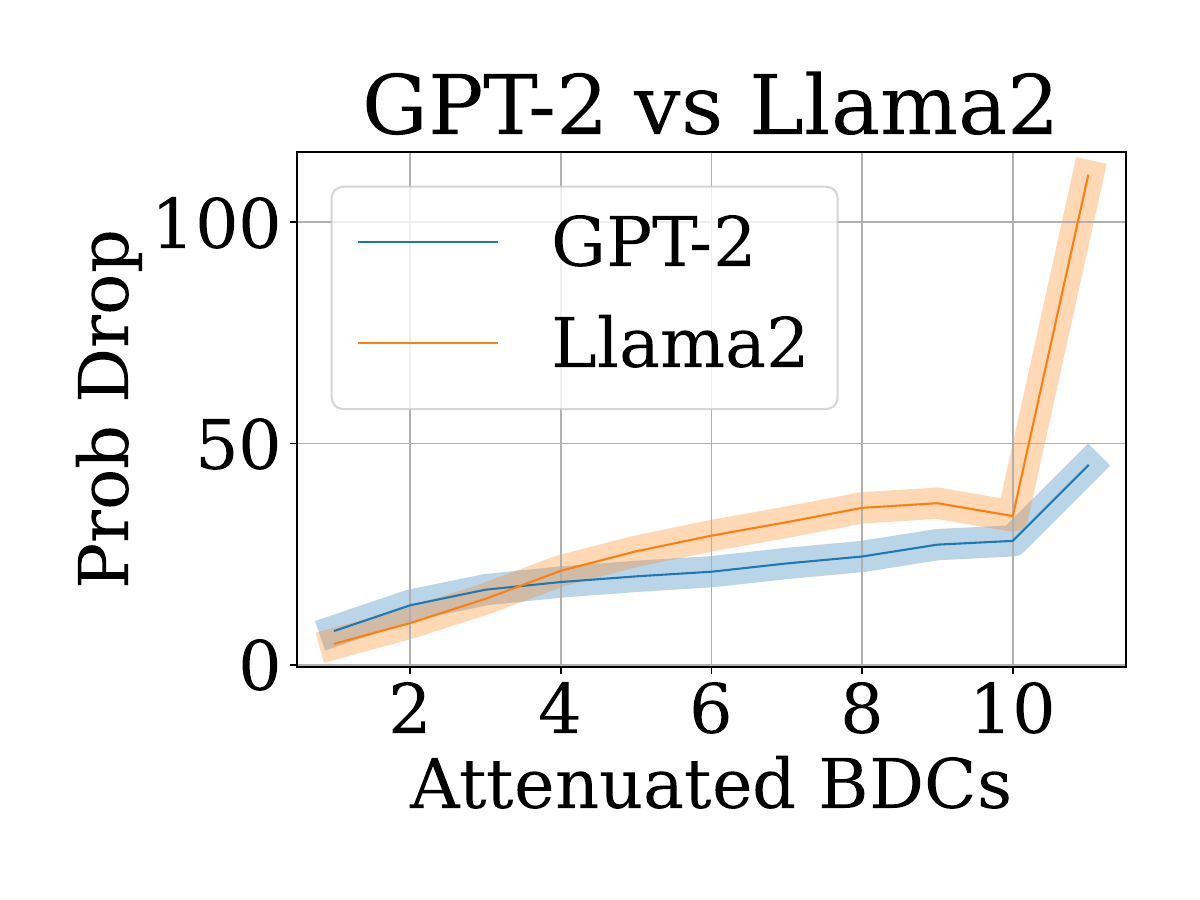}
\end{subfigure}%
\begin{subfigure}{.5\linewidth}
\centering
\includegraphics[width=\linewidth]{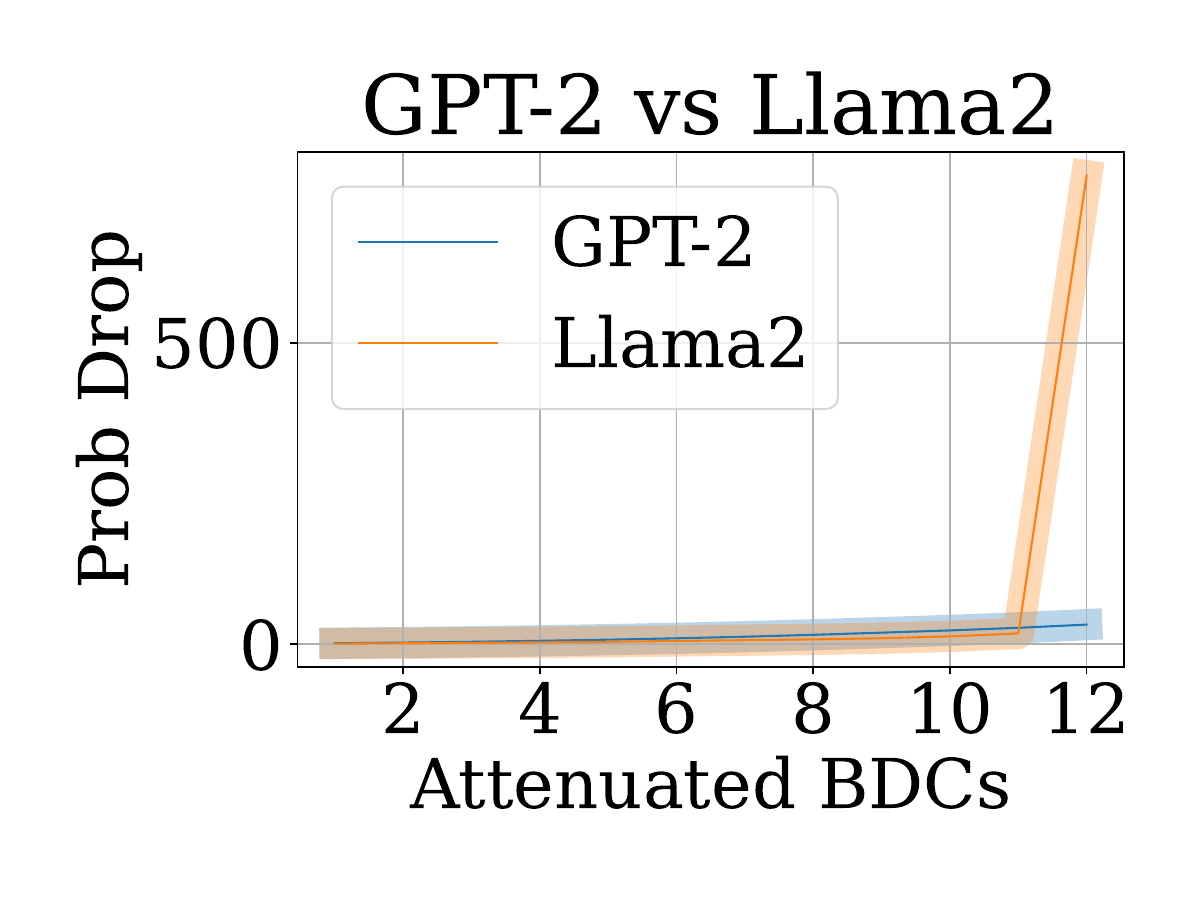}
\end{subfigure}

\begin{subfigure}{.5\linewidth}
\centering
\includegraphics[width=\linewidth]{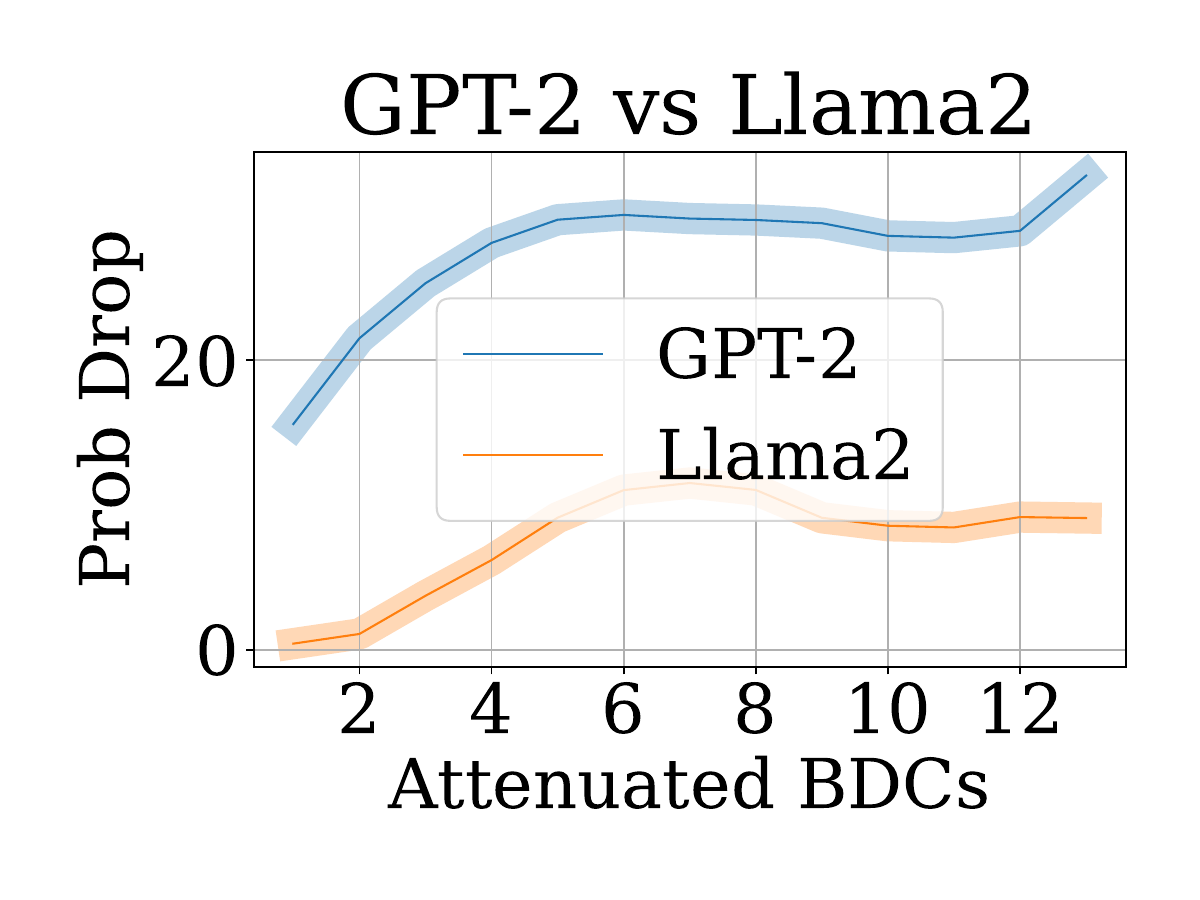}
\end{subfigure}%
\begin{subfigure}{.5\linewidth}
\centering
\includegraphics[width=\linewidth]{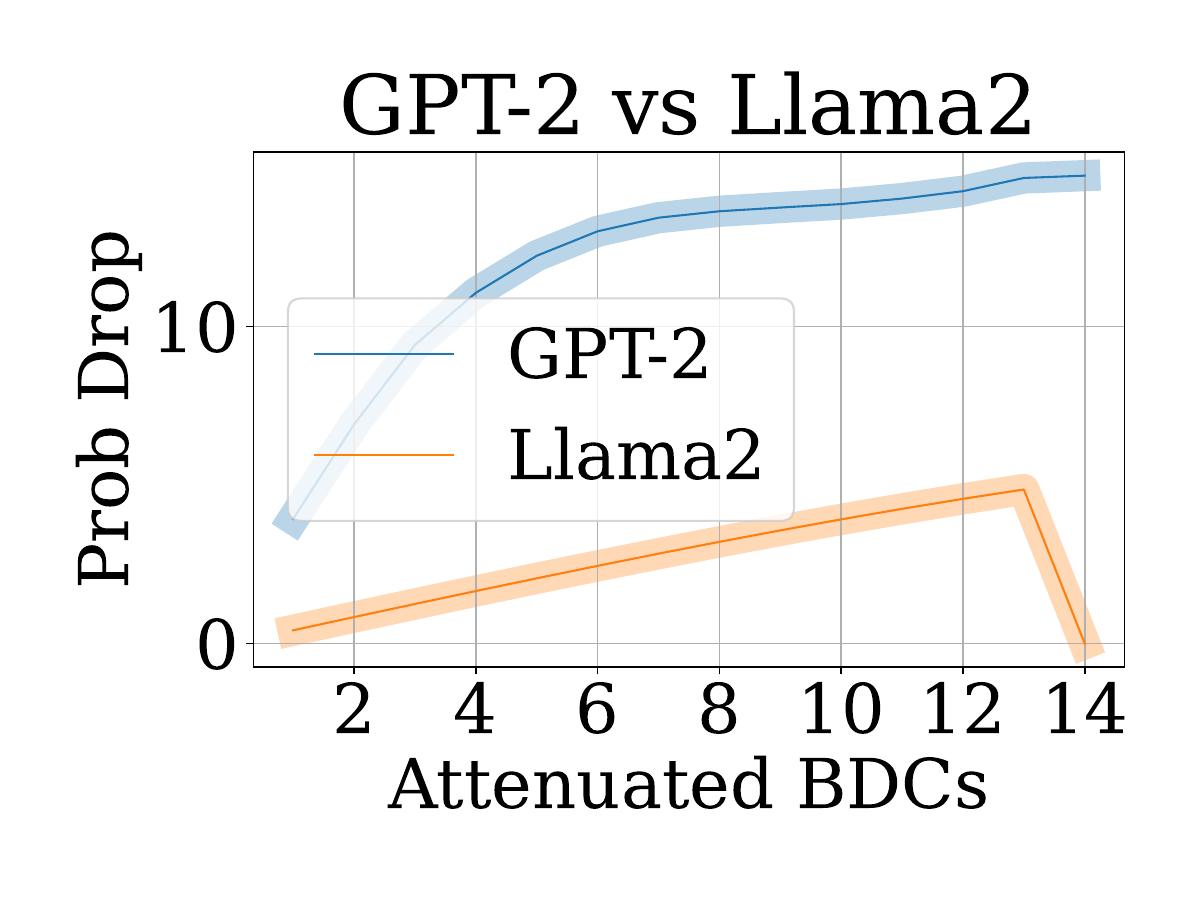}
\end{subfigure}

\begin{subfigure}{.5\linewidth}
\centering
\includegraphics[width=\linewidth]{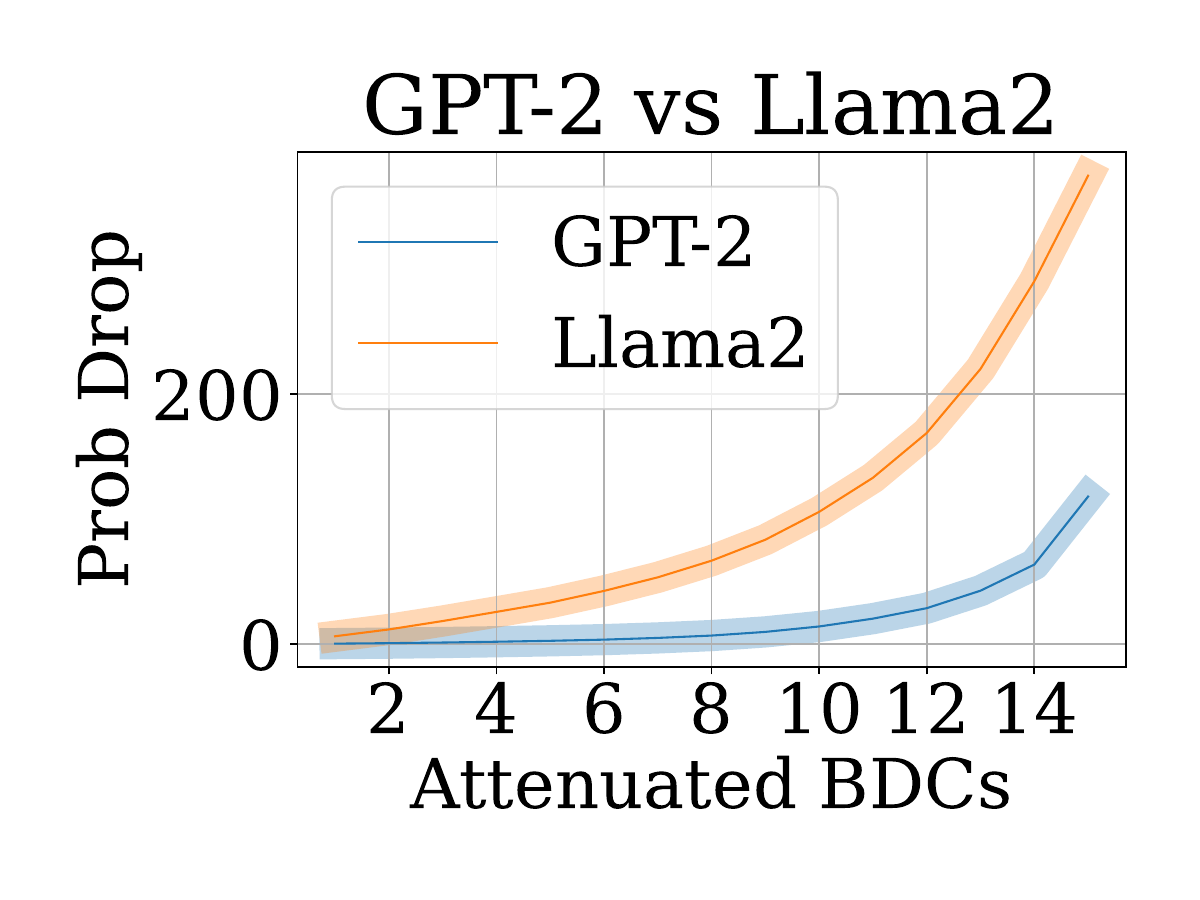}
\end{subfigure}%
\begin{subfigure}{.5\linewidth}
\centering
\includegraphics[width=\linewidth]{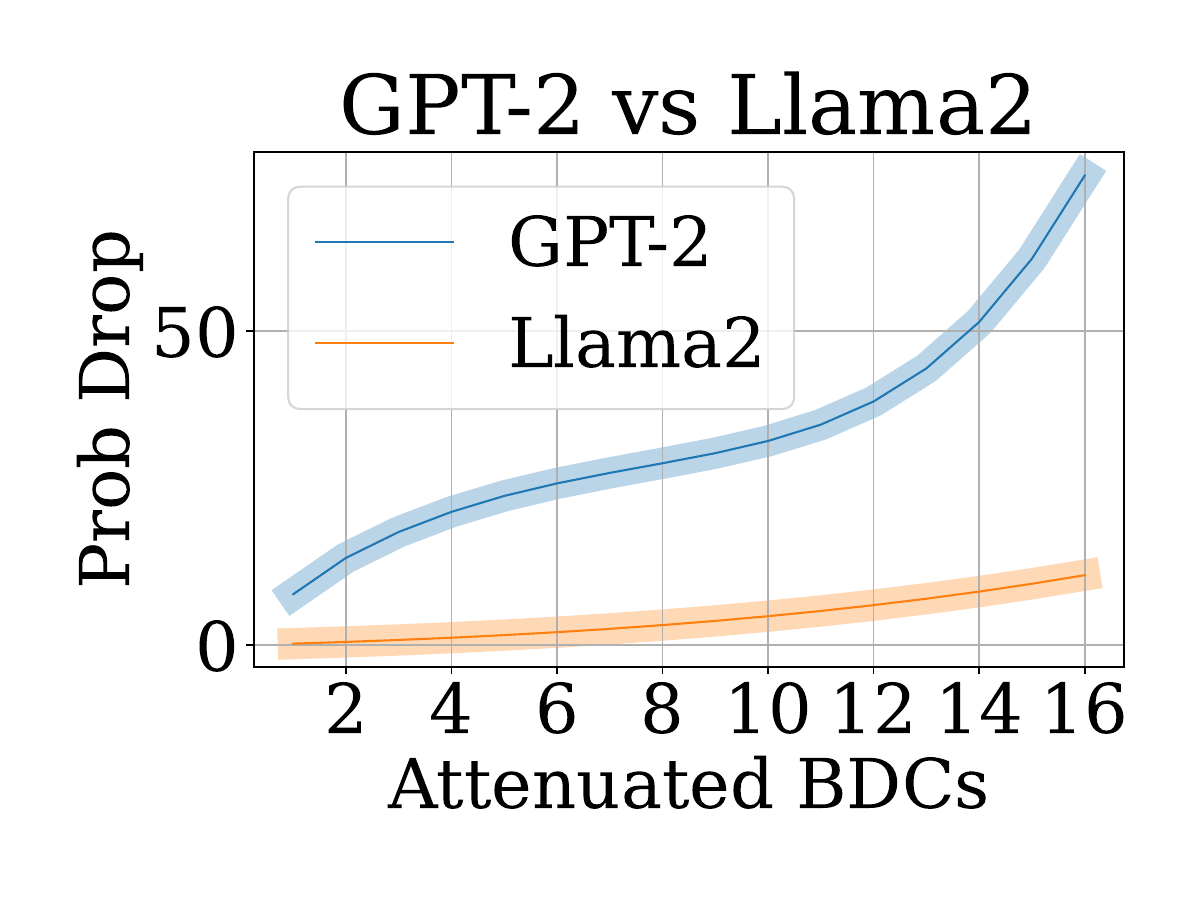}
\end{subfigure}

\begin{subfigure}{.5\linewidth}
\centering
\includegraphics[width=\linewidth]{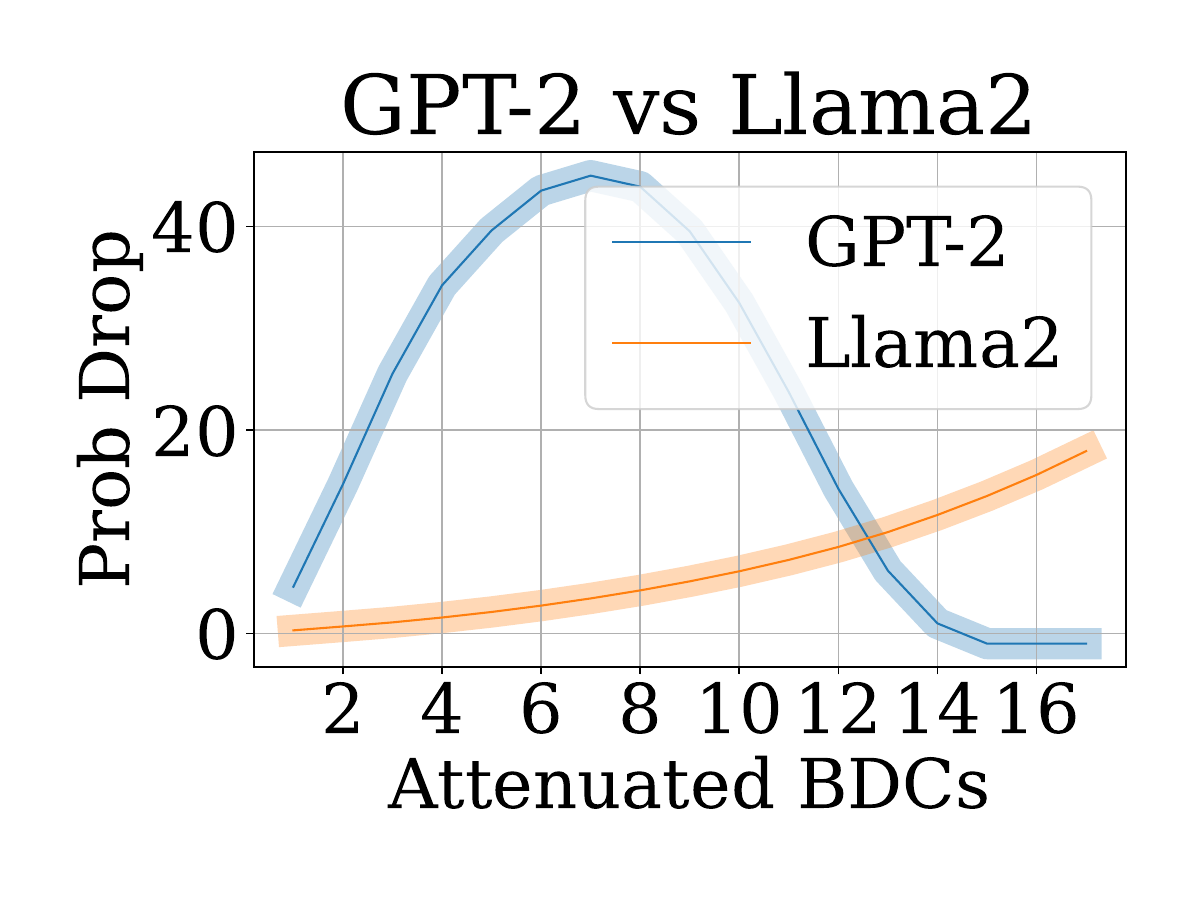}
\end{subfigure}%
\begin{subfigure}{.5\linewidth}
\centering
\includegraphics[width=\linewidth]{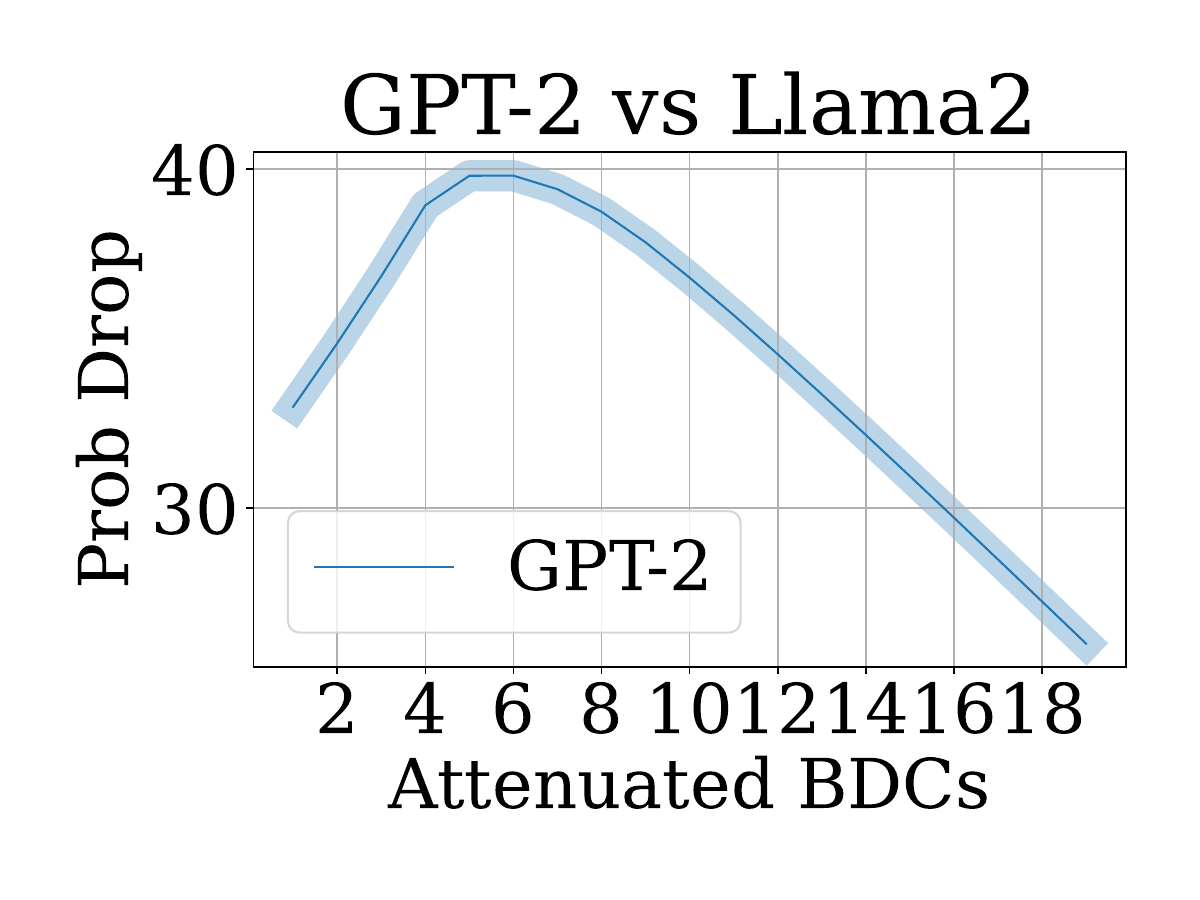}
\end{subfigure}

\caption{The graph of $\Delta Prob$ relative to the number of suppressed BDCs obtained using the K-Means method (Part 2). }
\label{fig-appendix-K-Means-2}
\end{figure}

For our method, we have also created line charts for DKNs of each length, serving as a supplement to Figure \ref{fig-main}. We use the same background color as the figures in the main text to represent images that are completely consistent with the main text, while a white background is used for the line charts corresponding to DKNs of other lengths.
Figure \ref{fig-appendix-main} show the results corresponding to DKN obtained using the NTC method. Additionally, we also present the results corresponding to other methods. Figure \ref{fig-appendix-DBSCAN} display the results corresponding to DKN obtained using the DBSCAN method. Figure \ref{fig-appendix-Hierarchical} show the results corresponding to DKN obtained using the Hierarchical method. Figures \ref{fig-appendix-K-Means-1} and \ref{fig-appendix-K-Means-2} illustrate the results corresponding to DKN obtained using the K-Means method. Since the DKN length for the method corresponding to AMIG \cite{chen2023journey} is limited to 2, there is no need to display the inflection points.

\label{sec:appendix}

\end{document}